\pdfoutput=1
\documentclass{article}
\usepackage[margin=1in]{geometry}
\usepackage{eqparbox}

\usepackage{microtype}
\usepackage{graphicx}
\usepackage{booktabs}

\usepackage{xcolor,colortbl}
\usepackage{microtype}      
\usepackage{multirow}
\usepackage{amsmath,amsfonts,amssymb, amsthm}
\usepackage{algorithm, algorithmic}
\usepackage{tikz}
\usepackage{pgfplots}
\usepackage{mathtools}
\usepackage{fixmath}
\usepackage{arydshln}
\usepackage{wrapfig}
\usepackage{enumitem}
\usepackage{graphicx}
\usepackage{transparent}
\usepackage{tabularx}
\usepackage{natbib}

\usepackage{float}

\usepackage{epstopdf}
\usepackage{caption}
\usepackage{subcaption}

\usepackage{caption}
\usepackage{calc}
\usepackage{relsize}
\usepackage{pifont}

\usepackage{tikz}

\usepackage{import}
\usepackage{pdfpages}

\usepackage[pdf]{graphviz}

\definecolor{MidnightBlue}{RGB}{25, 25, 112}
\definecolor{Red}{rgb}{0.6,0,0}
\definecolor{lightgreen}{RGB}{171, 225, 175}

\usepackage[backref=page]{hyperref}
\hypersetup{colorlinks=true}
\hypersetup{linktoc=all}
\hypersetup{citecolor=MidnightBlue}
\hypersetup{linkcolor=Red}
\hypersetup{urlcolor=MidnightBlue}
\usepackage[all]{hypcap}
\usepackage[noabbrev, nameinlink, capitalize]{cleveref}

\definecolor{radarblue}{HTML}{186d9c}
\definecolor{radarorange}{HTML}{ba611b}

\definecolor{radargreen}{HTML}{158a6a}
\definecolor{radaryellow}{HTML}{c38723}

\definecolor{radarpurple}{HTML}{8b008b}
\definecolor{radartomato}{HTML}{ff6348}

\title{Multi-modal Data Spectrum: \\ Multi-modal Datasets are Multi-dimensional}

\author{
    Divyam Madaan\thanks{New York University. Correspondence to \texttt{divyam.madaan@nyu.edu}}\\
    \and
     Varshan Muhunthan\footnotemark[1]\\
  \and 
    Kyunghyun Cho\footnotemark[1] \textsuperscript{,}\thanks{Genentech} \textsuperscript{,}\thanks{CIFAR} \\
    \and 
  Sumit Chopra\footnotemark[1] \textsuperscript{,}\thanks{New York University Grossman School of Medicine}
  \and 
}

\date{}
\pgfplotsset{compat=1.18}

\begin{document}

\maketitle

\begin{abstract}
Understanding the interplay between intra-modality dependencies (the contribution of an individual modality to a target task) and inter-modality dependencies (the relationships between modalities and the target task) is fundamental to advancing multi-modal learning. However, the nature of and interaction between these dependencies within current benchmark evaluations remains poorly characterized. In this work, we present a large-scale empirical study to quantify these dependencies across 23 visual question-answering benchmarks using multi-modal large language models (MLLMs) covering domains such as general and expert knowledge reasoning, optical character recognition, and document understanding. Our findings show that the reliance on vision, question (text), and their interaction varies significantly, both across and within benchmarks. We discover that numerous benchmarks intended to mitigate text-only biases have inadvertently amplified image-only dependencies. This characterization persists across model sizes and types, with models often obtaining high performance by using each modality independently and showing limited dependence on their interaction. We provide a quantitative characterization of multi-modal datasets, enabling a principled approach to multi-modal benchmark design and evaluation. 

\end{abstract}

\vspace{-0.1in}
\section{Introduction}

Rapid advancement of MLLMs has been accompanied by a significant increase in the number of evaluation benchmarks. A recent survey~\citep{li2024survey} identified over 200 multi-modal benchmarks. However, this growth has not been accompanied by a systematic investigation of what these datasets measure. This means the relationships, redundancies, and unique contributions across and within the benchmarks are not well understood. It is unclear whether a new dataset improves multi-modal evaluation or is redundant with existing benchmarks. This ambiguity makes benchmark selection for model evaluation a significant challenge.

For example, datasets such as AI2D~\citep{kembhavi2016diagram}, ChartQA~\citep{masry2022chartqa}, BLINK~\citep{fu2024blink}, RealworldQA~\citep{grokv2024}, $V^*$ Bench~\citep{wu2024v}, TextVQA~\citep{singh2019towards} were included in the Gemini 1.5 evaluation~\citep{team2024gemini}, but were omitted from Gemini 2.5~\citep{comanici2025gemini} with little justification for the changes. Such inconsistencies in evaluation protocols are common~\citep{grokv2024, commandA2025}, making it difficult to determine whether the reported gains in performance represent true advances in capability or simply adaptation to a different set of benchmark artifacts.

This lack of understanding has led to an inefficient cycle of benchmark development. New datasets are created to address specific uni-modal dependencies~\citep{agrawal2018don}, which in turn are found to have new and unforeseen artifacts~\citep{ dancette2021beyond,si2022language}. This process hinders consistent, long-term model comparison and undermines scientific rigor.

Prior work has analyzed the dependence on individual modalities and their interaction in multi-modal models using representation similarity~\citep{kornblith2019similarity}, information-theoretic measures~\citep{tjandrasuwita2025understanding, lu2023theory, madaan2024jointly}, and score-based methods~\citep{gat2021perceptual, parcalabescu2022mm, hu2022shape, wenderoth2025measuring}. While providing valuable insights, these studies were limited in scope, focusing on synthetic data, smaller-scale benchmarks such as VQA~\citep{agrawal2018don, goyal2017making}, or earlier generations of models. 

\begin{figure}[!t] 
    \centering
    \includegraphics[width=\linewidth]{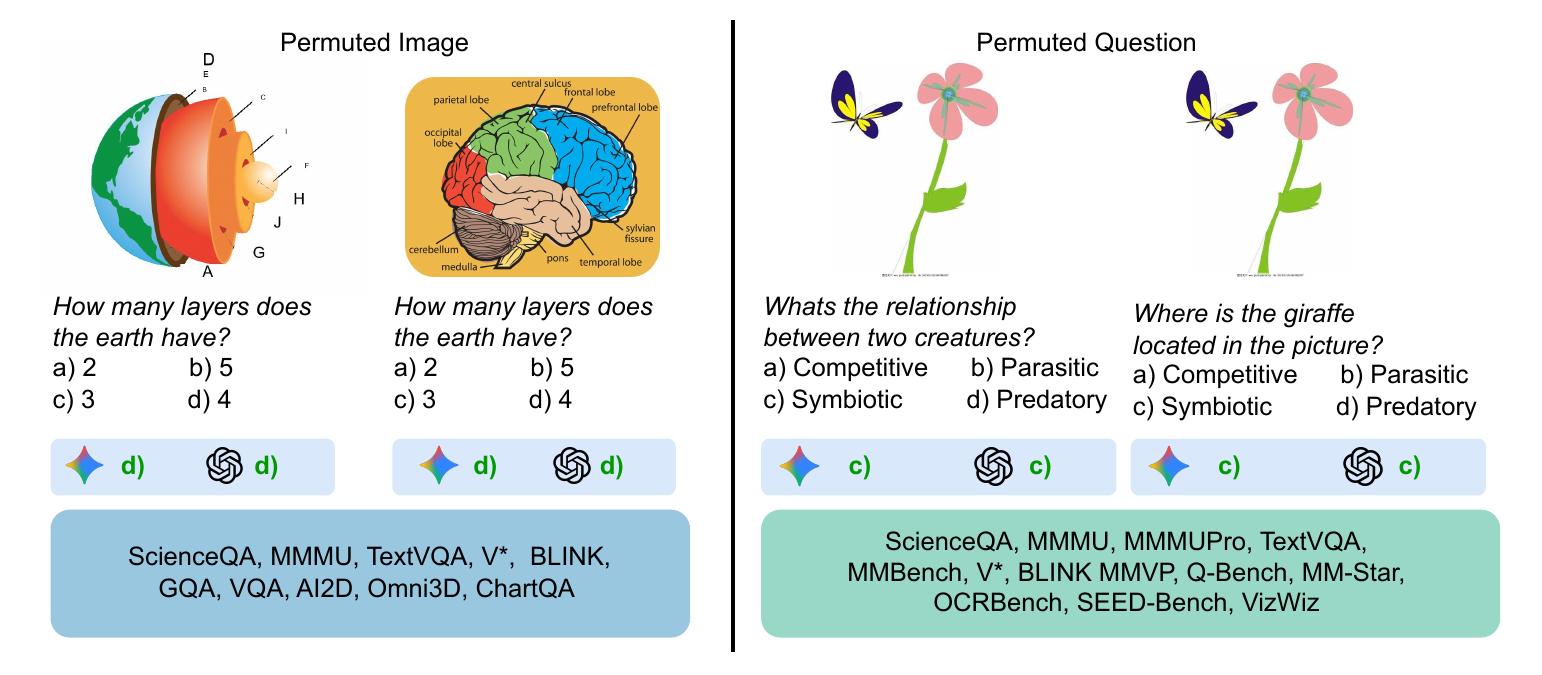}
    \vspace{-0.3in}
    \caption{{\bf Demonstration of intra-modality dependencies in multi-modal models using input permutation.}
(Left) The models answers about layers of Earth even when the image is replaced by an unrelated diagram of a brain.
(Right) The model identifies a symbiotic relationship from the image even when the question is unrelated. These examples highlight a failure of multi-modal reasoning, where models exploit uni-modal priors with the options to obtain an associated answer.}
    \label{fig:concept_figure}
\end{figure}

To address this gap, we conduct a large-scale empirical study to characterize widely-used multi-modal benchmarks. We hypothesize that these benchmarks evaluate distinct combinations of underlying capabilities. To quantify the multi-dimensional nature of these dependencies, we use intra-modality dependencies (reliance on a single modality for the target task) and inter-modality dependencies (reliance on the interaction between modalities for the target task) based on prior studies~\citep{liang2023quantifying, madaan2024jointly}. As shown in \Cref{fig:concept_figure}, MLLMs often exploit intra-modality dependencies, answering questions correctly even when a relevant input modality is replaced with corrupted or random data. To quantify modality reliance, we adapt Perceptual Score~\citep{gat2021perceptual}. We permute one modality across samples while keeping the other aligned with the labels and measure the resulting performance drop.

Our evaluation spans 23 multiple-choice visual question answering (MCVQA) benchmarks, spanning applications such as general visual question answering, knowledge-based reasoning, real-world spatial understanding, optical character recognition (OCR), and document and chart understanding. We evaluate MLLMs at varying scales, including 8B, 13B, and 34B models~\citep{tong2024cambrian, liu2023visual, bai2025qwen2, bai_qwen3-vl_2025}. Our findings confirm our hypothesis; the strength of intra- and inter-modality dependencies vary substantially across and within these benchmarks.

We show that models depend on one input modality while underutilizing the other, rather than using inter-modality dependencies (see \Cref{fig:concept_figure}). We find that many benchmarks designed to mitigate text-only dependencies~\citep{singh2019towards, li2023seed, tong2024eyes, fu2024blink} have inadvertently introduced strong image-only biases,  trading one uni-modal shortcut for another rather than evaluating multi-modal reasoning. This issue is not resolved by increasing model scale. On the contrary, larger models often become more adept at using uni-modal dependencies. These results underscore the fundamental limitations of evaluating models with a single aggregate score and highlight the need for a characterization of our evaluation benchmarks based on their strengths of inter- and intra-modality dependencies. 

{\bf Contributions.} We conduct the first large-scale empirical analysis of multi-modal dependencies across 23 popular VQA benchmarks. Our analysis shows that these datasets are multi-dimensional regarding their reliance on vision, text, and their interaction, and consequently measure different aspects of multi-modal learning. We find that these differences vary not only across datasets but also within individual benchmarks. To perform this analysis, we use a systematic method to characterize these dependencies. Our results provide a quantitative basis for the design and selection of future multi-modal benchmarks.

\section{The Multi-modal Spectrum}\label{sec:multimodal_spectrum}

This section defines inter- and intra- modality dependencies (\Cref{subsec:problem_setup}) for multi-modal learning. We argue that the failure to systematically measure these dependencies has led to an iterative cycle of benchmark design and circumvention (\Cref{sec:game_design}). Existing quantification methods (\Cref{subsec:relatedwork}) do not scale to recent datasets and MLLMs, which motivates our work.

\subsection{Problem Setup}\label{subsec:problem_setup}

In supervised multi-modal learning, given a dataset $\mathcal{D} = \{(\mathbf{x}_1^{(i)}, \mathbf{x}_2^{(i)}, \mathbf{y}^{(i)})\}_{i=1}^N$, the goal is to learn a mapping to predict the target label $\mathbf{y}$ from two distinct modalities, $\mathbf{x}_1$ and $\mathbf{x}_2$.
The target label $\mathbf{y}$ can be predicted from two distinct dependencies~\citep{liang2023quantifying, madaan2024jointly}: intra-modality dependency or uniqueness, where $\mathbf{y}$ is dependent on an individual modality, and inter-modality dependency or synergy, where modalities provide joint information not present in isolation. For example, in video-based sentiment analysis, a positive sentiment might be uniquely determined from strong lexical cues within a text transcript alone. In contrast, detecting sarcasm requires interpreting the conflict between the literal semantics of the text and audio or visual expressions of the video.

Following prior work~\citep{liang2023quantifying, madaan2024jointly}, we model this distinction with a selection variable $\mathbf{v}$ in the multi-modal data generating process, where $\mathbf{v}=1$ is a mechanism to model the dependencies between the modalities and the target task:
\begin{equation}\label{eq:joint_dist}
p(\mathbf{y}, \mathbf{x}_1, \mathbf{x}_2, \mathbf{v}=1) = p(\mathbf{y}) p(\mathbf{x}_1|\mathbf{y}) p(\mathbf{x}_2|\mathbf{y})p(\mathbf{v}=1|\mathbf{x}_1,\mathbf{x}_2,\mathbf{y}).
\end{equation}
Although this framework provides a way to separate the effects of individual modalities from their joint combinations, the actual strength of uniqueness and synergy within popular benchmarks and MLLMs remains largely unquantified.

\subsection{Cat-and-Mouse Game of Benchmark Design}
\label{sec:game_design}

The lack of a principled characterization of these dependencies has resulted in a cat-and-mouse game of benchmark development and subsequent circumvention. This iterative cycle spans a spectrum of multi-modal datasets, from those solvable with a single modality to those that require inter-modality dependencies. To evaluate the multi-modal capabilities of a model, new benchmarks deliberately weaken intra-modality dependencies to necessitate inter-modality dependencies~\citep{goyal2017making, agrawal2018don,dancette2021beyond, si2022language, fu2024blink, tong2024eyes, wu2024v}. Despite these design constraints, models frequently achieve high performance by using intra-modality dependencies. This reliance on intra-modality dependencies is subsequently framed as an exploitation of uni-modal artifacts~\citep{wang2020makes,liang2023quantifying, zhang2024understanding}, a behavior that has been assigned labels such as model laziness~\citep{zhang2024multimodal}, modality competition~\citep{huang2022modality}, or modality greediness~\citep{wu2022characterizing}, which prompts further cycles of benchmark revision.

The history of VQA exemplifies this cycle. The original VQA dataset \citep{antol2015vqa} contained strong language priors, allowing models to achieve high accuracy by guessing common answers based on the type of questions. To counter this, VQAv2~\citep{goyal2017making} was introduced, which balanced the dataset by ensuring each question had two images leading to different answers. The subsequent VQA-CP benchmark \citep{agrawal2018don} further enforced this constraint by changing the answer distribution between the training and test sets to penalize models that relied only on question-based priors. Similarly, the VQA-CE~\citep{dancette2021beyond} and VQA-VS~\citep{si2022language} datasets were introduced to highlight the prevalence of multi-modal shortcuts in prior VQA benchmarks. This iterative pattern of creation and attack continues with recent benchmarks, such as the progression from MMMU~\citep{yue2024mmmu} to MMMU-Pro~\citep{yue2025mmmupro}.

Without a systematic way to quantify these dependencies, it is difficult to determine whether the performance of a multi-modal model  stems from multi-modal capabilities or from simply exploiting dominant intra-modality dependencies. This ambiguity hinders progress, as we continue to develop complex architectures and algorithms~\citep{li2021align, wu2022characterizing, zheng2023judging, liu2023visual, young2024yi} without a clear understanding of the spectrum of inter- and intra-modality dependencies in current models and datasets.

\subsection{Quantifying the Strength of Dependencies}
\label{subsec:relatedwork}

Several quantitative metrics have been developed to measure the dependence of a model on individual modalities. A straightforward approach is to measure performance degradation after shuffling a modality's input at test time, where the resulting performance drop is attributed to that modality's contribution~\citep{gat2021perceptual}. More sophisticated methods, such as MM-Shap~\citep{parcalabescu2022mm}, SHAPE~\citep{hu2022shape}, and InterShap~\citep{wenderoth2025measuring}, use Shapley values to assign importance scores to individual image regions and text tokens, yielding a fine-grained analysis independent of task accuracy.

Despite these advances, no work has systematically positioned recent MLLM evaluation datasets along a continuous multi-modal spectrum defined by their inter- and intra-modality dependencies. In the next section, we adapt a practical methodology based on the perceptual score~\citep{gat2021perceptual} to measure these dependency strengths. We select this method for its simplicity in the two-modality case and its ability to directly compute each modality's marginal contribution.
By characterizing datasets along the spectrum of multi-modal dependencies, we can design more targeted benchmarks. Further, we gain deeper insights into model capabilities, paving the way for more robust and generalizable multi-modal systems.

\section{Recipe for Future Datasets and Models}
\label{sec:recipe}

Given a multi-modal dataset $\mathcal{D}$ consisting of instances $(\mathbf{x_1}, \mathbf{x_2}, \mathbf{y})$, where $\mathbf{x_1}$ is an image, $\mathbf{x_2}$ is a text, and $\mathbf{y}$ is the ground truth label, we detail a principled evaluation framework inspired by \citet{gat2021perceptual}. This requires a baseline multi-modal model $f_\theta$ to evaluate performance, measured by a metric $\mathcal{M}$, under four different input conditions. The chosen baseline model should ideally be a state-of-the-art multi-modal model that has not been trained on the dataset under analysis, thus preventing data leakage. 

The four evaluation conditions are:

\begin{enumerate}
    \item \textbf{Paired modalities (Normal):} The model's performance is measured on original, paired data points, $\mathcal{M}(f_\theta(\mathbf{x_1}, \mathbf{x_2}), \mathbf{y})$.

    \item \textbf{Unimodal (Image only):} The paired text $\mathbf{x_2}$ is replaced with a text instance $\mathbf{x'_2}$ randomly sampled from another data point. Performance on\ $\mathcal{M}(f_\theta(\mathbf{x_1}, \mathbf{x'_2}), \mathbf{y})$ isolates the informational contribution of the image modality $\mathbf{x_1}$.

    \item \textbf{Unimodal (Text only):} Symmetrically, the image $\mathbf{x_1}$ is replaced with a random image $\mathbf{x_1}'$. Performance on $\mathcal{M}(f_\theta(\mathbf{x'_1}, \mathbf{x_2}), \mathbf{y})$ isolates the contribution of the text modality $\mathbf{x_2}$.

    \item \textbf{Both modalities shuffled (Random):} Both modalities are replaced with randomly sampled, uncorrelated instances $(\mathbf{x'_1}, \mathbf{x'_2})$. The model's performance on $\mathcal{M}(f_\theta(\mathbf{x'_1}, \mathbf{x'_2}), \mathbf{y})$ establishes a random baseline.

\end{enumerate}

A dataset that appears balanced at the global level can still contain strong intra-modality dependencies within specific subsets of its data. It is therefore essential that this procedure be supplemented with a more granular analysis of data subgroups. This involves applying the same diagnostic to data subsets categorized by relevant features, such as question type or object categories.

{\bf Rationale for modality shuffling.} We adopt modality shuffling over the option of zeroing out (e.g., using a blank image or an empty string)~\citep{, gu_illusion_2025} or input perturbation as in prior studies~\citep{hu2022shape,tong2024cambrian}. Zeroing out or adding perturbation creates unnatural, out-of-distribution inputs that elicit unpredictable model behavior, confounding the measurement of inter- and intra-modality dependencies. In contrast, shuffling preserves the marginal distribution of each modality. The model still receives valid inputs, but the inter-modality dependency is broken. 
The performance metrics derived from this shuffling procedure, visualized in \Cref{sec:experiments}, enable a direct quantification of inter- and intra-modality dependencies.

{\bf Model-based analysis.}
Multi-modal dependencies are a function of both the data and the model interpreting it. Thus, an analysis based on a single model may be confounded by specific inductive biases of that model. To obtain a robust estimate of intrinsic data dependencies, the effect of any single model must be marginalized out. We achieve this using a majority-vote ensemble~\citep{dietterich2000ensemble} of diverse models to reduce the influence of idiosyncratic model biases.

\section{Experiments}\label{sec:experiments}

In this section, we describe the evaluation datasets and models used in \Cref{sec:datasets_and_models}.~\Cref{sec:results} shows the overall performance metrics and \Cref{sec:category_analysis} shows the results in various subcategories across multiple datasets.

\subsection{Datasets and Models}\label{sec:datasets_and_models}

To assess the capabilities of MLLMs, we use a comprehensive suite of benchmark datasets. Based on the core evaluation skills, we categorize the benchmarks chronologically to show the progression in each category.

\begin{itemize}

\item {\bf General visual question answering.}
For general VQA, we focus on benchmarks that test real-world and compositional reasoning. We include VizWiz~\citep{gurari2018vizwiz}, which poses questions from visually impaired users about everyday, uncurated scenes. Following this, we use GQA~\citep{hudson2019gqa} to evaluate visual reasoning and compositional reasoning. To evaluate a wider range of abilities, we incorporate MME~\citep{fu2023mme}, which covers 14 perception tasks. SEED-Bench~\citep{li2023seed} expands on these with a large-scale multiple choice question format. MMBench~\citep{liu2024mmbench} further evaluates 20 abilities, including object localization and social reasoning. 

\item {\bf Expert visual question answering.}
To measure performance on tasks requiring specialized knowledge, we evaluate with multiple benchmarks. This includes ScienceQA~\citep{lu2022learn}, which contains questions from the natural sciences, language and social sciences. We also use MathVista~\citep{lu2023mathvista}, which tests mathematical reasoning (logical, arithmetic, and statistical) in diverse visual formats such as word problems, geometric shapes, and plots. For expert-level evaluation, we incorporate MMMU~\citep{yue2024mmmu} and MMMU-Pro~\citep{yue2025mmmupro}, which consist of college-level problems from exams and textbooks in six core disciplines, probing multi-modal understanding and reasoning.

\item {\bf Real-world spatial understanding.}
We use Microsoft COCO dataset~\citep{lin2014microsoft} for object recognition and ADE~\citep{zhou2019semantic} for scene understanding. To measure object-level hallucinations, we use the POPE benchmark~\citep{li2023evaluating} and measure spatial understanding using RealWorldQA~\citep{grokv2024}. To address the growing importance of temporal reasoning, we include MMVP~\citep{tong2024eyes}, which tests comprehension and reasoning about long-form video content. Omni3D~\citep{brazil2023omni3d, tong2024cambrian} contains the task of determining the depth order and relative distance of 3D objects. Q-Bench~\citep{wu2024qbench} and BLINK~\citep{fu2024blink} evaluate low-level visual perception and general understanding.  $V^*$ Bench~\citep{wu2024v} focuses on visual grounding in high-resolution images. MM-Star~\citep{chen2024we} is another vision-centric benchmark with human-validated samples to test six fundamental multi-modal capabilities.

\item {\bf Optical character recognition (OCR) and document, chart understanding.}
We start by evaluating using TextVQA~\citep{singh2019towards}, which requires models not only to read, but also to reason about text embedded in images. We expand the scope of evaluation with OCRBench~\citep{liu2024ocrbench}, which provides a multifaceted assessment that includes text recognition, scene text-centric VQA, document-oriented VQA, key information extraction, and handwritten mathematical expression recognition. 

For document and chart understanding, we evaluate the model's ability to comprehend complex layouts and the relationships between visual elements. We start with AI2D~\citep{kembhavi2016diagram} for understanding schematic diagrams followed by a ChartQA~\citep{masry2022chartqa}, a challenging dataset of human-generated question-answer pairs on various charts and plots.

\end{itemize}

\begin{figure}[t!] 
    \centering
    \begin{subfigure}{0.49\linewidth}
        \centering
        \includegraphics[width=\linewidth]{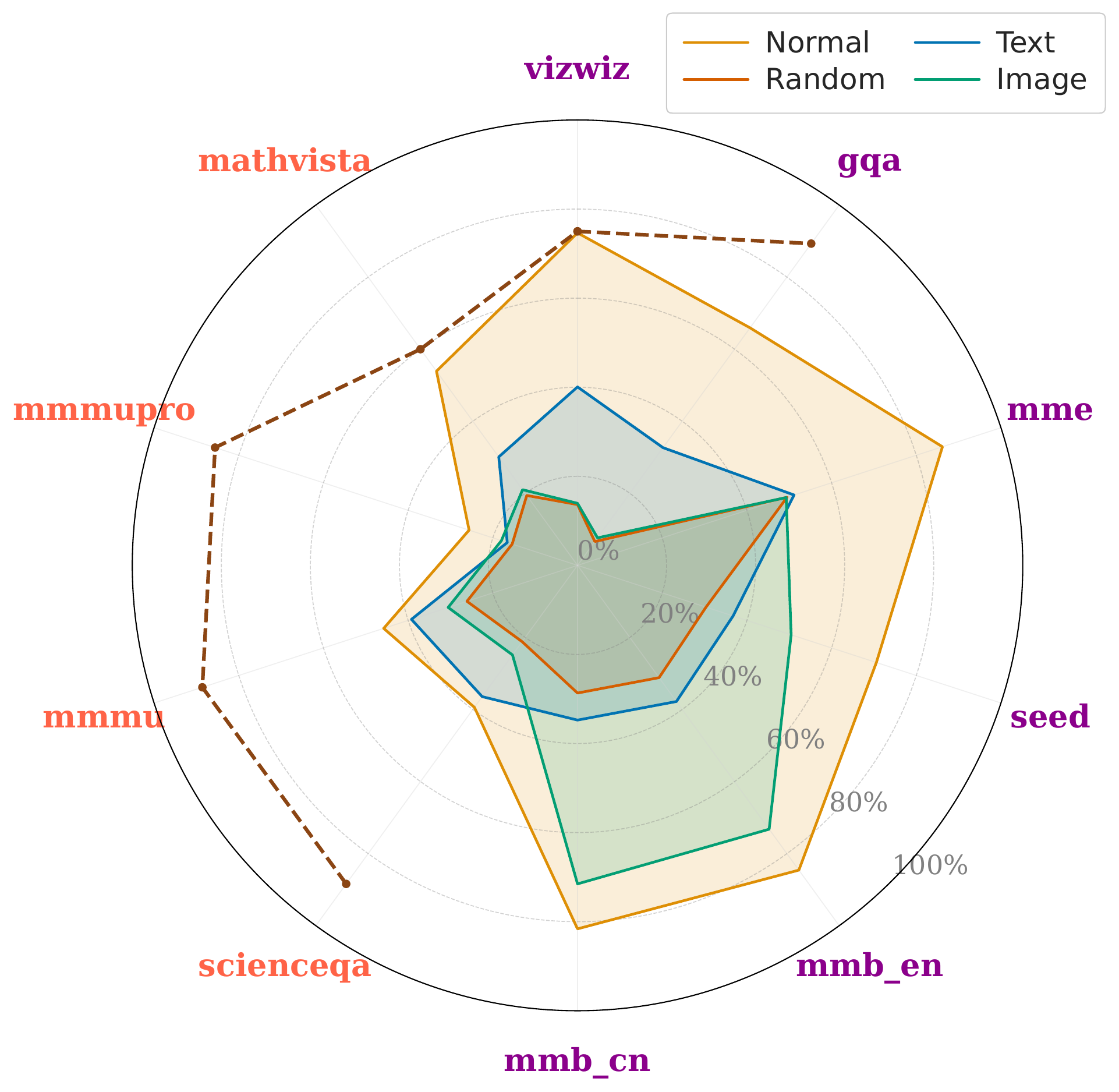}
        \caption{Datasets evaluating visual question answering with \textcolor{radarpurple}{\bf general}  and \textcolor{radartomato}{\bf expert}  questions.\label{subfig:general_expert_vqa}}
    \end{subfigure}
    \hfill
    \begin{subfigure}{0.49\linewidth}
        \centering
        \includegraphics[width=\linewidth]{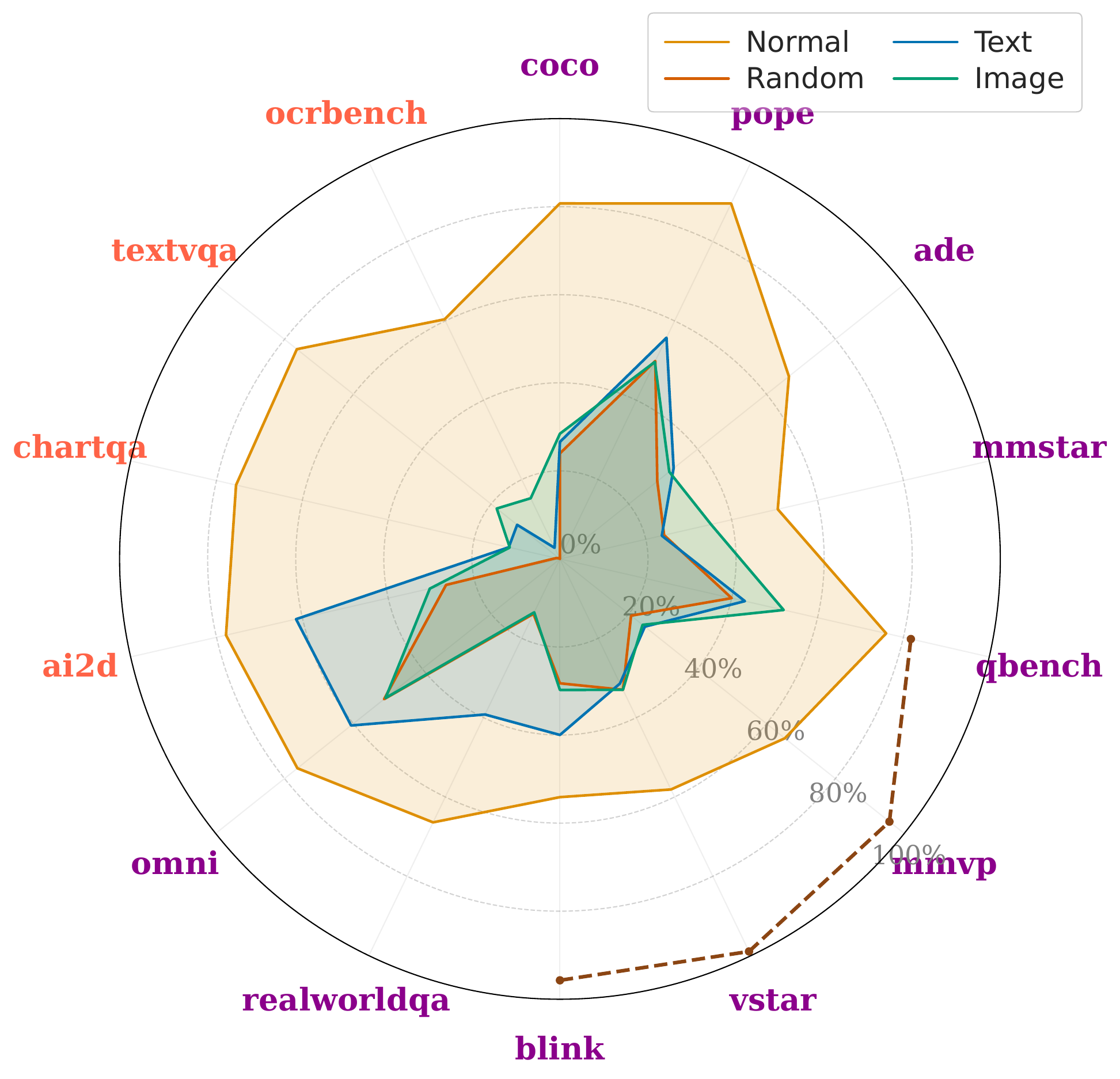}
        \caption{Datasets evaluating \textcolor{radarpurple}{\bf spatial understanding} and \textcolor{radartomato}{\bf OCR, data and chart understanding}  .\label{subfig:ocr_spatial_vqa}}
    \end{subfigure}
    \caption{Radar plot showing the comparison of an ensemble of \textcolor{radaryellow}{standard} MLLMs with \textcolor{radargreen}{image only}, \textcolor{radarblue}{text only} and \textcolor{radarorange}{random} performance using the recipe from \Cref{sec:recipe}. The dashed line indicates human performance, which is shown partially due to a lack of data for other benchmarks. \label{fig:multimodal_results}}
   
\end{figure}

We use the openly available 8B, 13B, and 34B models from Cambrian-1 ~\citep{tong2024cambrian}. These models are built upon Llama-3 8B~\citep{liu2023visual}, Vicuna-1.5 13B~\citep{vicuna2023}, and Nous-Yi 34B~\citep{young2024yi} for language processing. For vision, they incorporate a combination of architectures including ViT from SigLIP~\citep{zhai2023sigmoid, radford2021learning}, DINOv2~\citep{oquabdinov2}, and ConvNeXt-XXL~\citep{liu2022convnet}. Our main results are generated by taking a majority vote of the answers among these three models. The code is available online at \url{https://github.com/divyam3897/multimodal-spectrum}.

\subsection{Overall results}
\label{sec:results}

Our evaluation in \Cref{fig:multimodal_results} across 23 multi-modal datasets shows that most benchmarks contain both intra- and inter-modality dependencies, allowing models to answer questions without jointly using both the image and the question text.
We categorize datasets based on the modality dependencies they contain: (1) inter-modal only, and two non-exclusive categories for datasets that additionally contain (2) text intra-modality dependencies and (3) image intra-modality dependencies.

\paragraph{Datasets with inter-modality dependency only.} 

We show that multi-modal datasets with inter-modality dependency only are surprisingly rare. Across all evaluated benchmarks, only four datasets exhibit this characteristic.

\emph{For general and expert question answering, MME~\citep{fu2023mme} is the only dataset that demonstrates that permuting one modality makes the task challenging for the model.}  For spatial understanding, POPE~\citep{li2023evaluating}, COCO~\citep{lin2014microsoft,tong2024cambrian}, and V$^*$~\citep{wu2024v} datasets contain predominantly inter-modality dependencies. \emph{No datasets in the OCR and chart understanding categories exhibit inter-modality dependencies only.}

The simplest way to curate vision-language inter-modality datasets is to ensure that the answer changes with the change in one modality. This approach has been used in a few binary classification inter-modality datasets~\citep{suhr2019nlvr2, fu2023mme, li2023evaluating}. For instance, POPE and MME contains questions with yes and no answers for the same set of images. This ensures that a model relying on only one modality might correctly answer one question but will fail to correctly answer the corresponding inverse question. This leads to random performance when the inter-modality dependencies are ignored when a modality is shuffled.

\paragraph{Datasets with text intra-modality dependency.} 
\emph{Models when evaluated on general and expert knowledge show a reliance on text across all datasets.} 
For example, models with only the correct input question achieve scores well above random chance on GQA~\citep{hudson2019gqa}~$(+26\%)$, ScienceQA~\citep{lu2022learn}~$(+17.5\%)$, and MMMU~\citep{yue2024mmmu}~$(+11.35\%)$, demonstrating that visual input is often not considered necessary by the model for these datasets. This even extends to datasets designed specifically to emphasize visual grounding, such as Blink~\citep{fu2024blink}, RealWorldQA~\citep{grokv2024}, and Omni3D~\citep{brazil2023omni3d, tong2024cambrian}. The same pattern holds for OCR, document and chart understanding datasets such as AI2D~\citep{kembhavi2016diagram}, ChartQA~\citep{masry2022chartqa} and TextVQA~\citep{singh2019towards}, where using question only surpass random performance by $34.94$, $11.69$ and $12.19$ absolute points, respectively. 

These results underscore the challenge of designing benchmarks that do not contain any examples without text-only dependencies. While it is challenging to dissect the precise cause for these biases, many studies have conjectured several issues in data curation. These issues include shortcuts between language and corresponding answers~\citep{goyal2017making}, IID train-test splits~\citep{agrawal2018don}, shifted prior distributions~\citep{gat2021perceptual}, limited human-level perception abilities~\citep{fu2024blink} and failures to identify visual patterns in the image~\citep{tong2024eyes}.

\begin{figure*}[t]
    \centering
    \begin{subfigure}[b]{0.49\linewidth}
        \includegraphics[width=\linewidth]{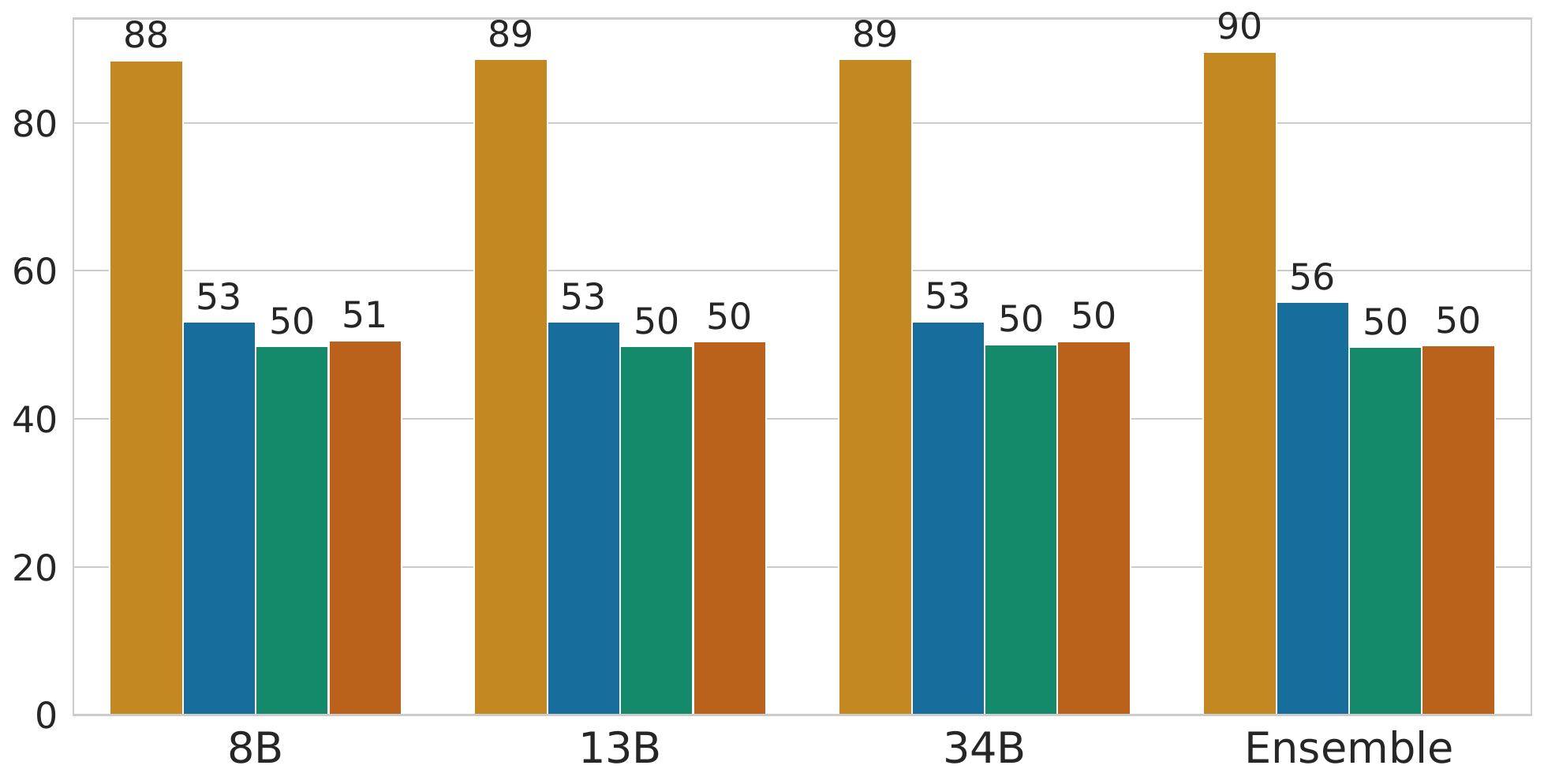}
        \caption{POPE}
    \end{subfigure}
    \hfill
    \begin{subfigure}[b]{0.49\linewidth}
        \includegraphics[width=\linewidth]{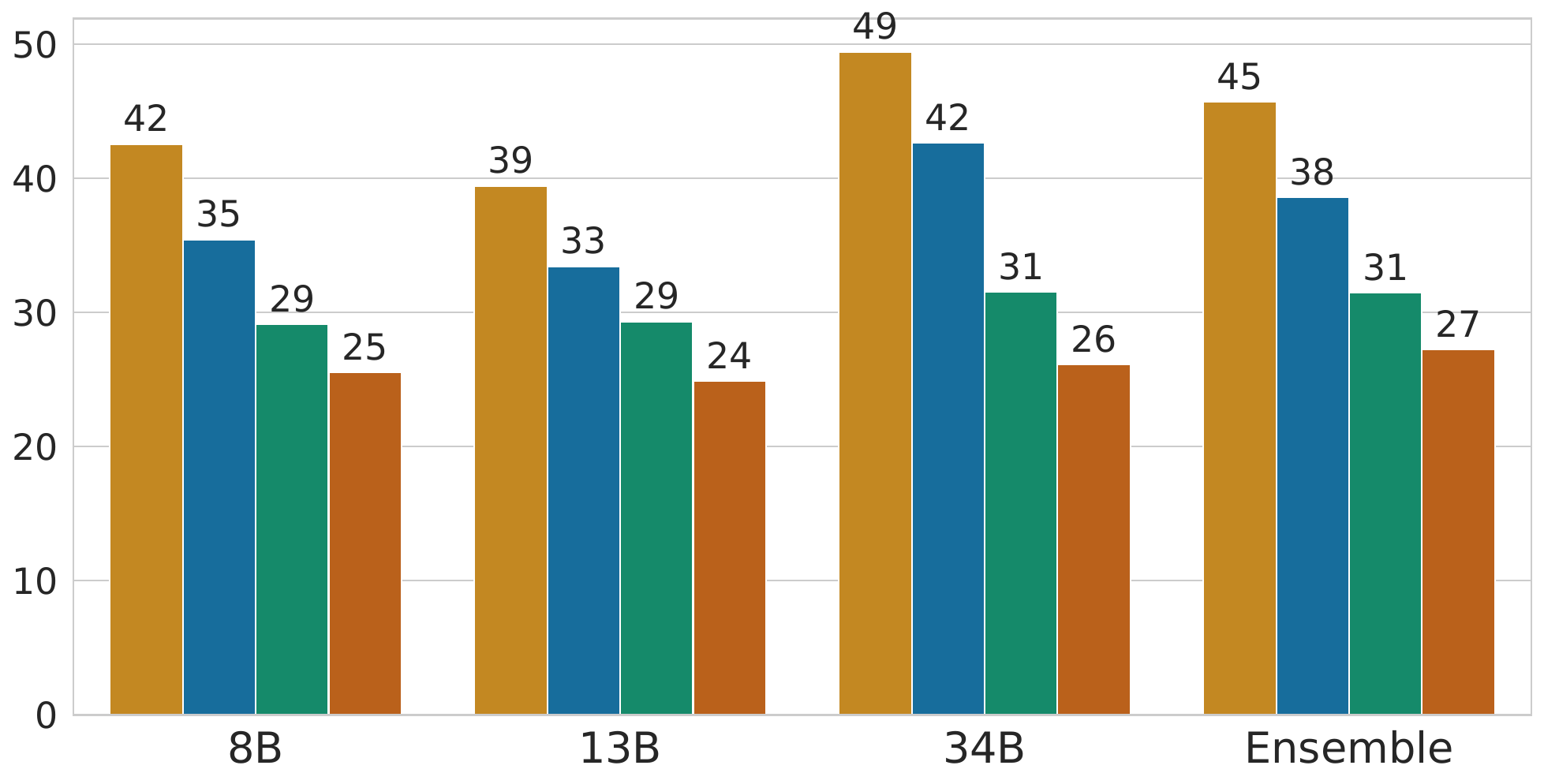}
        \caption{MMMU}
    \end{subfigure}
    \hfill
    \begin{subfigure}[b]{0.49\linewidth}
        \includegraphics[width=\linewidth]{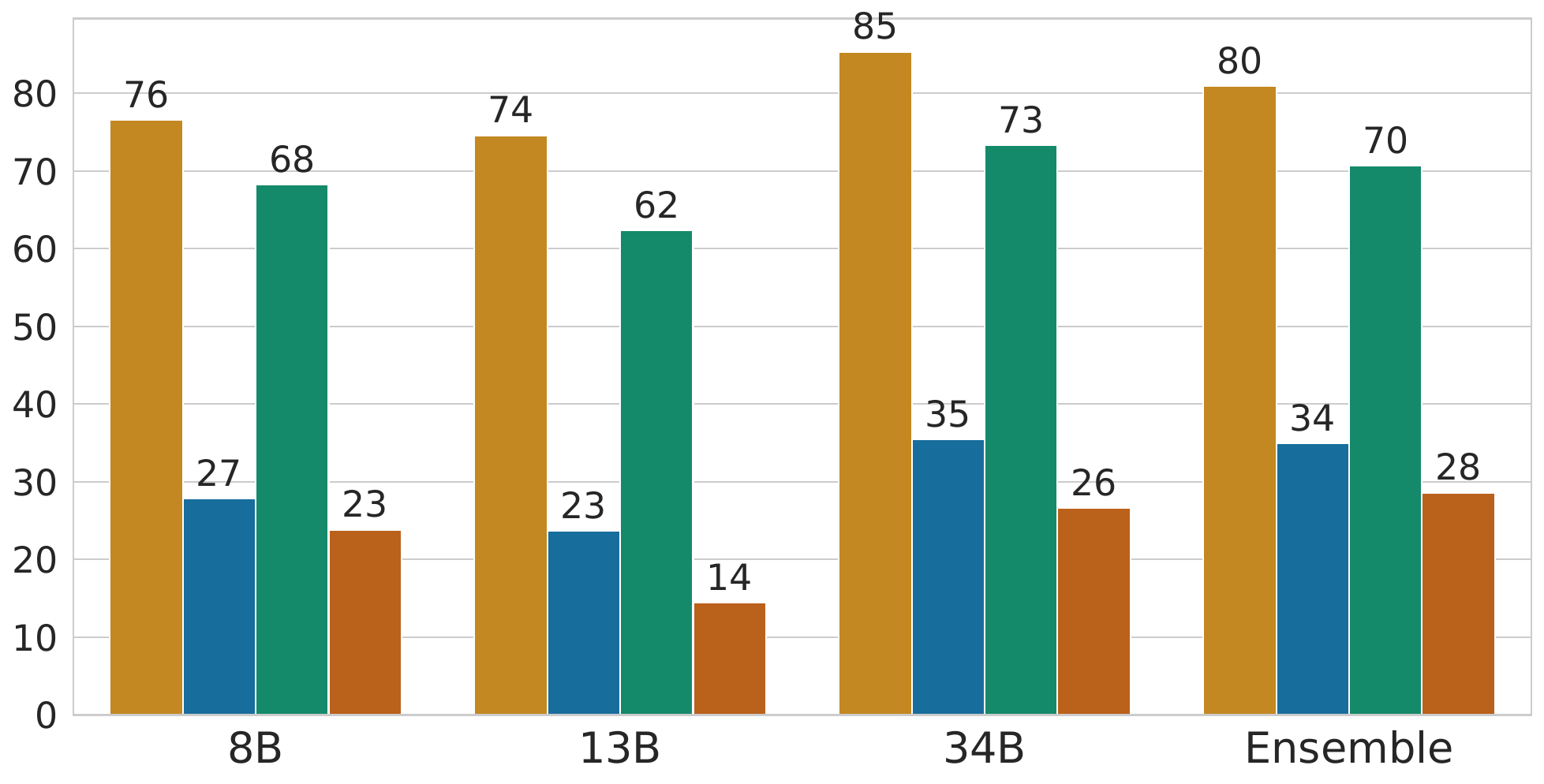}
        \caption{MMBench (cn)}
    \end{subfigure}
    \hfill
    \begin{subfigure}[b]{0.49\linewidth}
        \includegraphics[width=\linewidth]{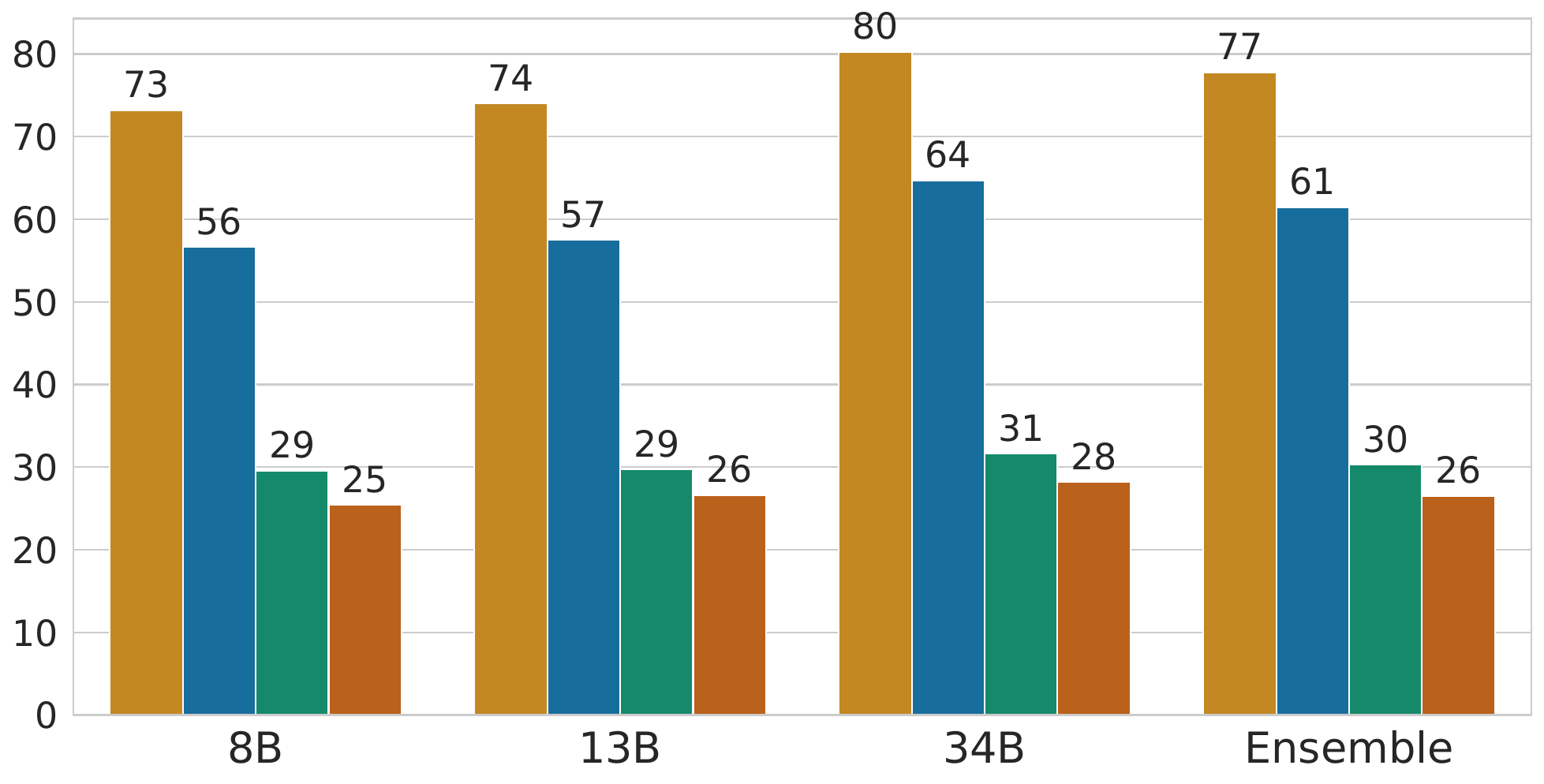}
        \caption{AI2D}
    \end{subfigure}
    \caption{{\bf Effect of Model Scaling on Modality Contribution.} Performance of various models (8B, 13B, 34B, and a majority-vote ensemble) on four datasets selected for their specific dependencies: GQA (text), SEED (image), and POPE (inter-modality). The bars represent \textcolor{radaryellow}{standard accuracy} and attributed contributions from \textcolor{radarblue}{text}, \textcolor{radargreen}{image}, and \textcolor{radarorange}{random} (bars are in the same order).\label{fig:model_size}}
\end{figure*}
\paragraph{Datasets with image intra-modality dependency.}
\emph{Efforts to eliminate textual biases from benchmarks have led to an unintended consequence of introduction of strong visual intra-modality dependencies.} We find that these newer datasets often allow models to succeed by relying solely on the image, effectively ignoring the question. This is most illustrated in MMBench~\citep{liu2024mmbench}, where an image-only model outperforms a random baseline by $41\%$. This issue persists even in benchmarks designed to focus on multi-modal reasoning, including TextVQA~\citep{singh2019towards}, ChartQA~\citep{masry2022chartqa}, SEED-Bench~\citep{li2023seed}, MMMU-Pro~\citep{yue2024mmmu}, MMVP~\citep{tong2024eyes}, Q-Bench~\citep{wu2024qbench}, and MM-Star~\citep{chen2024we}.

Instead of requiring multi-modal understanding, many of these evaluation benchmarks swapped a textual dependency with a visual one to obtain the correct answer. This is because their curation primarily focused on mitigating text-only intra-modality dependencies. We recommend that the central goal of a benchmark design should be to measure the intended task using both modalities for question answering, not to emphasize intra-modality dependencies.

\paragraph{Effect of model scaling.} Since our analysis is based on a model-dependent accuracy metric, we investigate how modal dependencies change across models of varying scales and architectures. We selected four datasets with distinct dependencies in \Cref{fig:model_size}: POPE~\citep{li2023evaluating}, which contains predominantly inter-modality dependencies; MMMU~\citep{yue2024mmmu}, which is reliant on both image and text intra-modality dependencies; MMBench~\citep{liu2024mmbench}, dependent on the image modality; and AI2D~\citep{kembhavi2016diagram}, reliant on the text intra-modality dependencies.

We find that uni-modal biases are not consistently mitigated by model scale and can even be exacerbated. For instance, on MMMU~\citep{yue2024mmmu}, scaling to a 34B parameter model increased the overall performance and the reliance on both image and text-only dependencies. Similarly, on MMBench~\citep{liu2024mmbench}, larger models exhibit an improved performance with a greater dependency on image-only dependencies (\Cref{fig:model_size}). AI2D~\citep{kembhavi2016diagram} shows a similar trend. The text-only intra-modality dependency increases with scale and persists in the majority-ensemble model. Conversely, performance on POPE~\citep{li2023evaluating}, a benchmark requiring only inter-modal dependencies, shows no change in performance with increase in model size. We further show results on additional datasets that depend on neither text nor image intra-modality dependencies, as well as datasets that depend on both in \Cref{fig:appendix_model_size}. We also include datasets that predominantly depend on image and text intra-modality dependencies in \Cref{fig:appendix_image_model_size} and \Cref{fig:appendix_text_model_size} respectively.

\begin{figure*}[t]
    \centering
    \begin{subfigure}[b]{0.49\linewidth}
        \includegraphics[width=\linewidth]{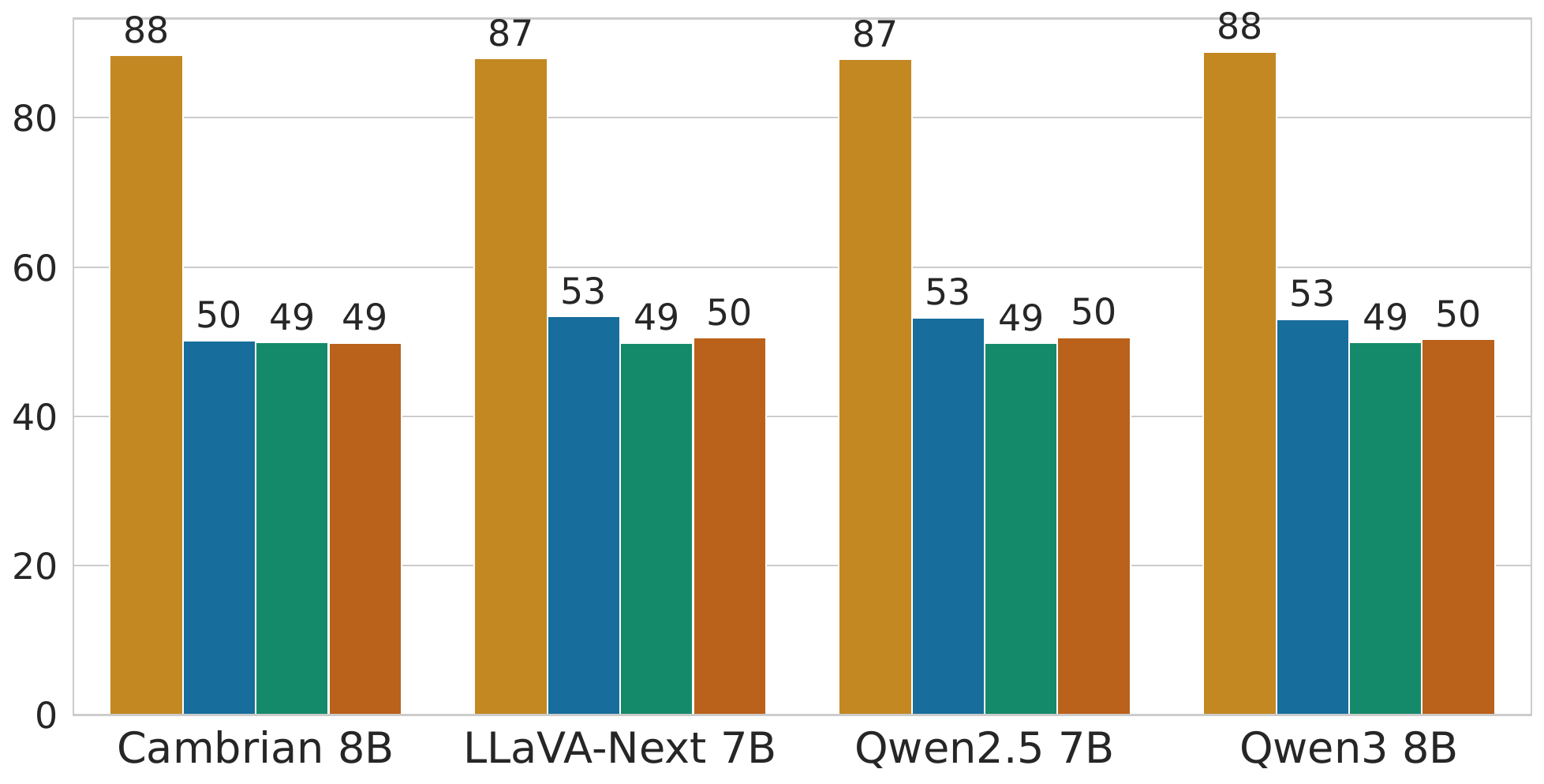}
        \caption{POPE}
    \end{subfigure}
    \hfill
    \begin{subfigure}[b]{0.49\linewidth}
        \includegraphics[width=\linewidth]{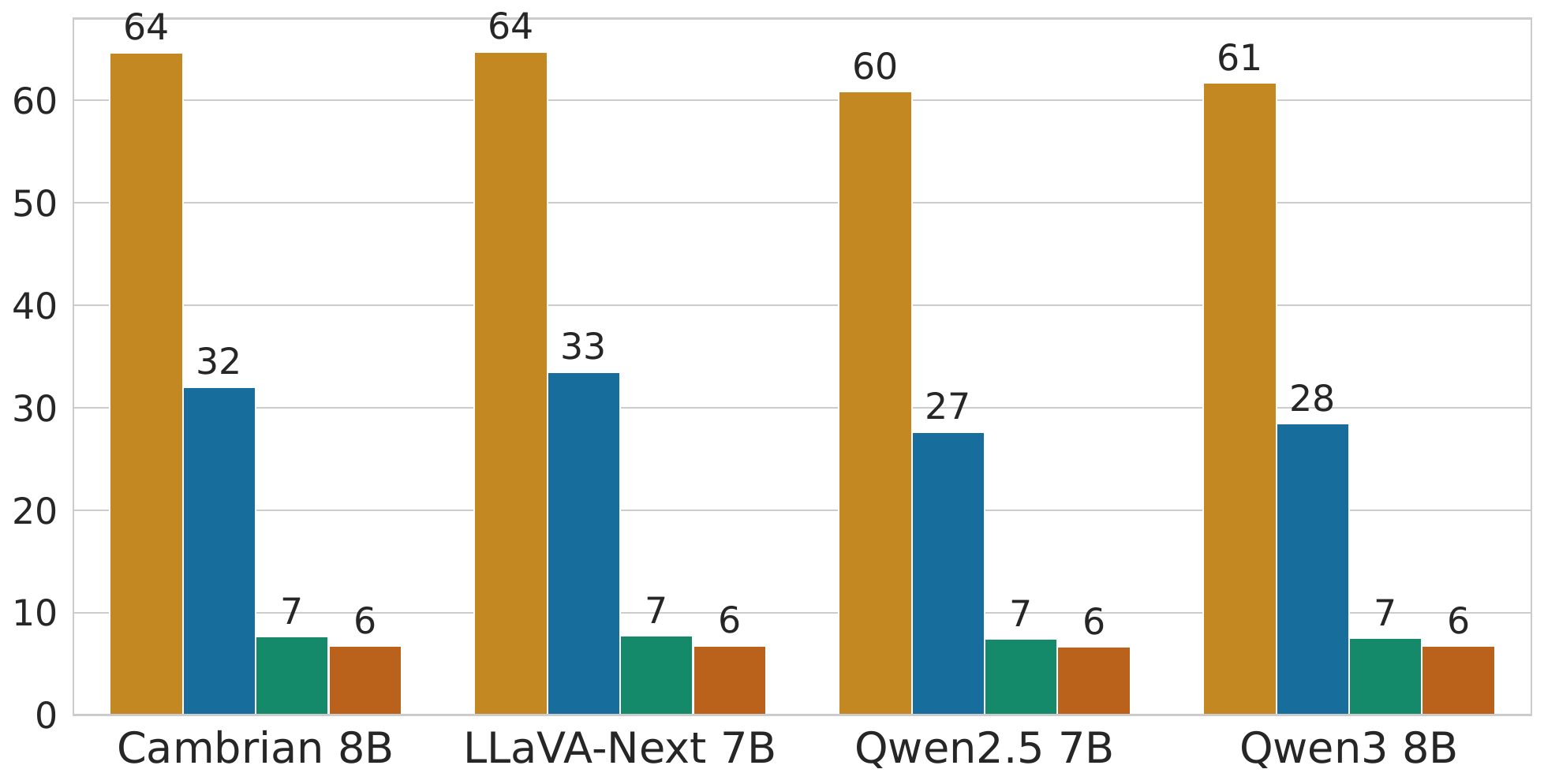}
        \caption{GQA}
    \end{subfigure}
    \hfill
    \begin{subfigure}[b]{0.49\linewidth}
        \includegraphics[width=\linewidth]{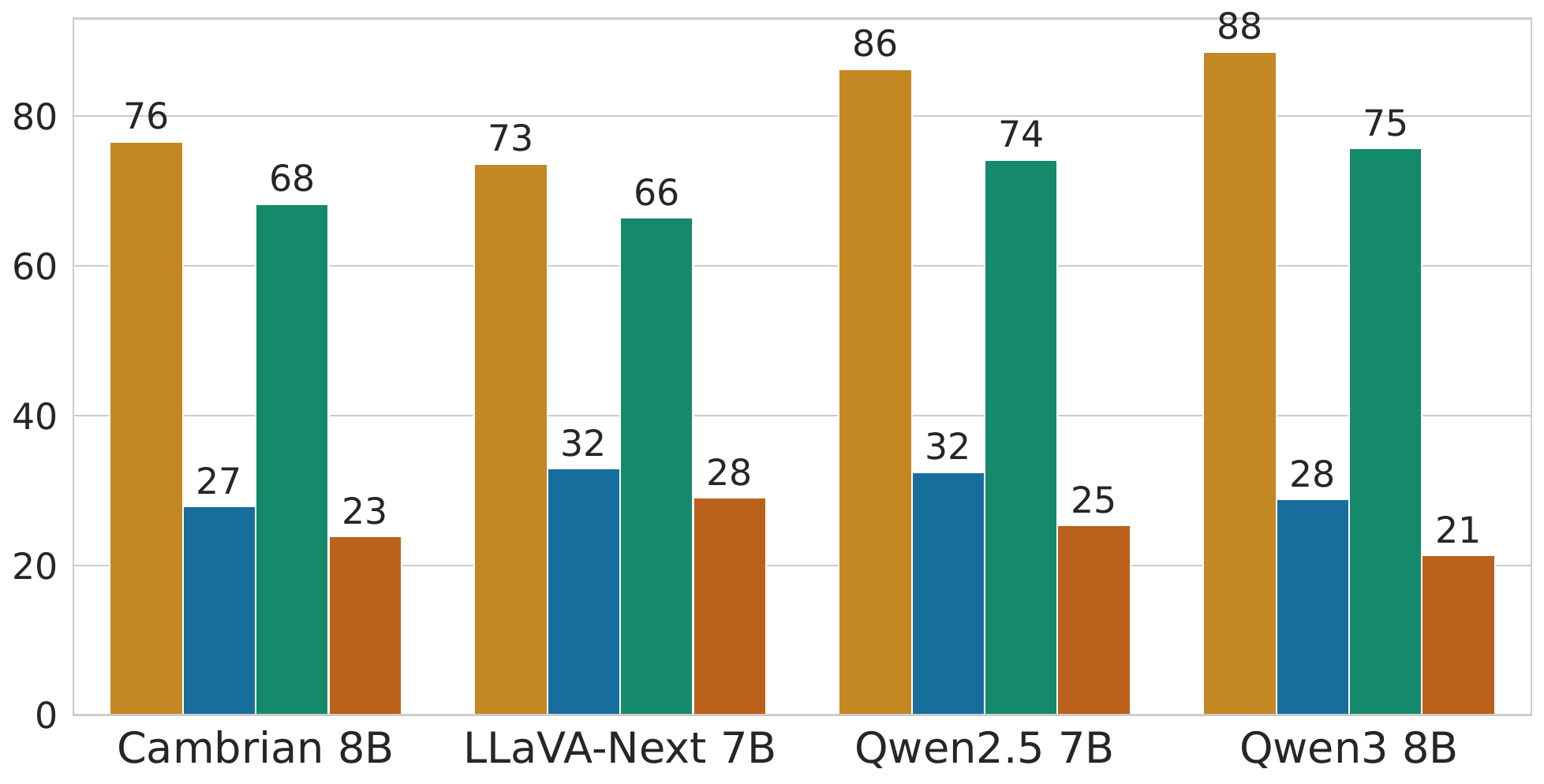}
        \caption{MMBench (cn)}
    \end{subfigure}
    \hfill
    \begin{subfigure}[b]{0.49\linewidth}
        \includegraphics[width=\linewidth]{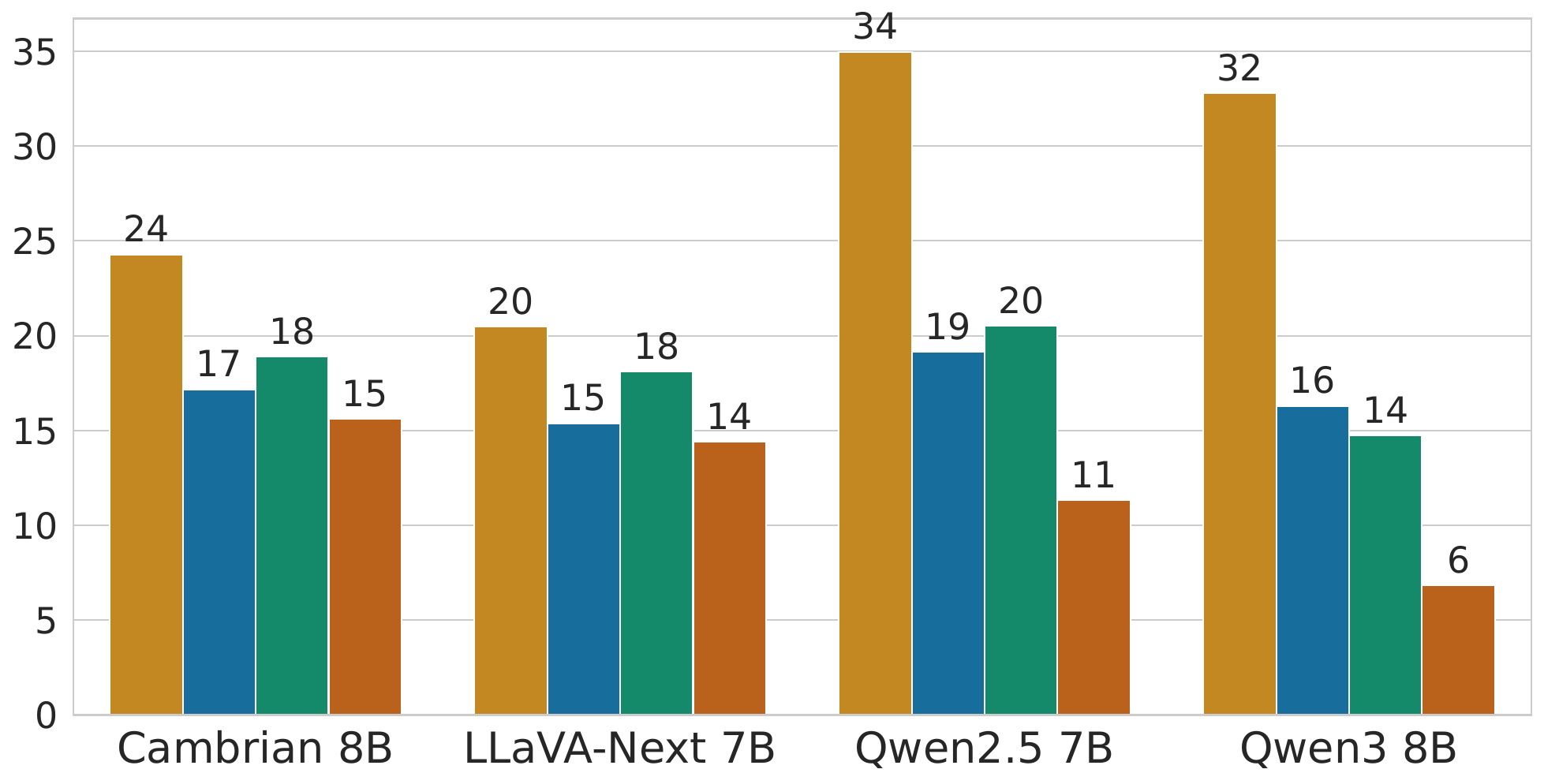}
        \caption{MMMU-Pro}
    \end{subfigure}
    \caption{{\bf Effect of model type on modality contribution.} Performance comparison between LLava-Next (May 2024), Cambrian-1 8b (June 2024), Qwen2.5-VL (April 2025) and Qwen3-VL (October 2025) on four datasets selected for their specific dependencies: GQA (text), MMBench (image), POPE (inter-modality) and MMMU-Pro (both image and text). The bars represent \textcolor{radaryellow}{standard accuracy} and attributed contributions from \textcolor{radarblue}{text}, \textcolor{radargreen}{image}, and \textcolor{radarorange}{random} (bars are in the same order).\label{fig:model_type}}
\end{figure*}
\paragraph{Effect of model types.} We compare four different instruction-tuned models in \Cref{fig:model_type}. Particularly, we compare Cambrian-8b released in June 2024 with LLama-3 8B base model~\citep{tong2024cambrian}, LLaVA-Next released in May 2024 with Mistral 7B model~\citep{liu2024llavanext}, Qwen2.5-VL from April 2025 with Qwen2.5 7B language model~\citep{bai2025qwen2} and Qwen3-VL 8B released recently in November 2025 with Qwen3 language model~\citep{bai_qwen3-vl_2025}.

Despite substantial differences in the evaluated models and their release dates, we consistently observe intra-modality biases (when present) across all of them. For instance, Qwen models improve the performance on MMBench by around ten percent compared to Cambrian-1 while also increasing the image-only performance significantly. For POPE, as expected, all models exhibit  near-random performance when using only image or only text inputs. For GQA and MMMU-Pro, we see comparable levels of biases across different models types. Similar results with additional datasets are shown in \Cref{fig:appendix_model_types}.

This findings raise the question of whether improvements in benchmark performance really reflect progress in multi-modal learning or whether models are simply becoming better at using intra-modality dependencies. We hope that our analysis will encourage reporting not only overall benchmark performance but also image-only, text-only, and random baselines, to enable a more holistic evaluation of multi-modal models.

\subsection{Category analysis}\label{sec:category_analysis}

\begin{figure*}[t]
    \centering
    \begin{subfigure}[b]{0.32\linewidth}
        \includegraphics[width=\linewidth]{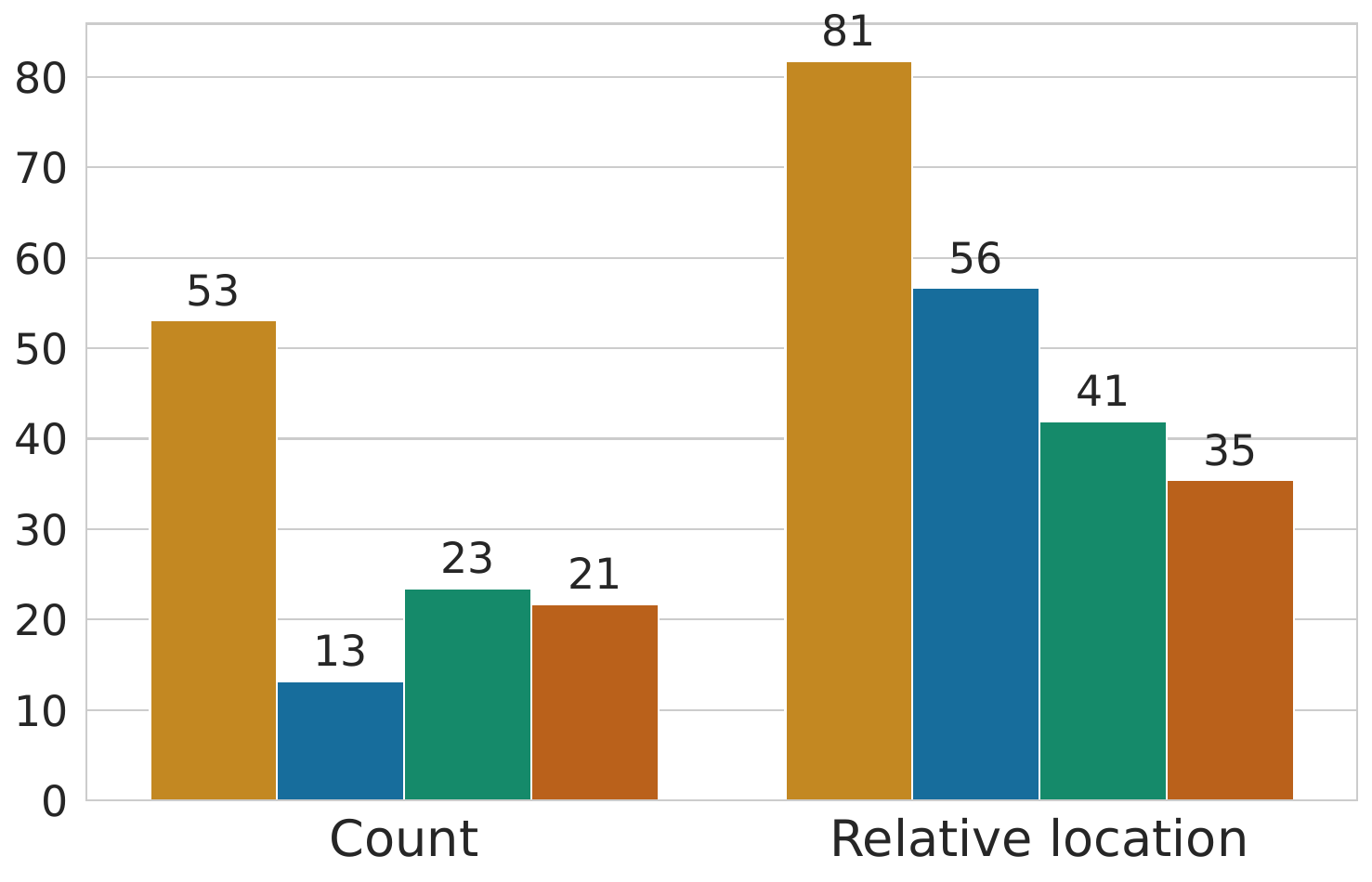}
        \caption{ADE\label{fig:ade}}
    \end{subfigure}
    \hfill
    \begin{subfigure}[b]{0.32\linewidth}
        \includegraphics[width=\linewidth]{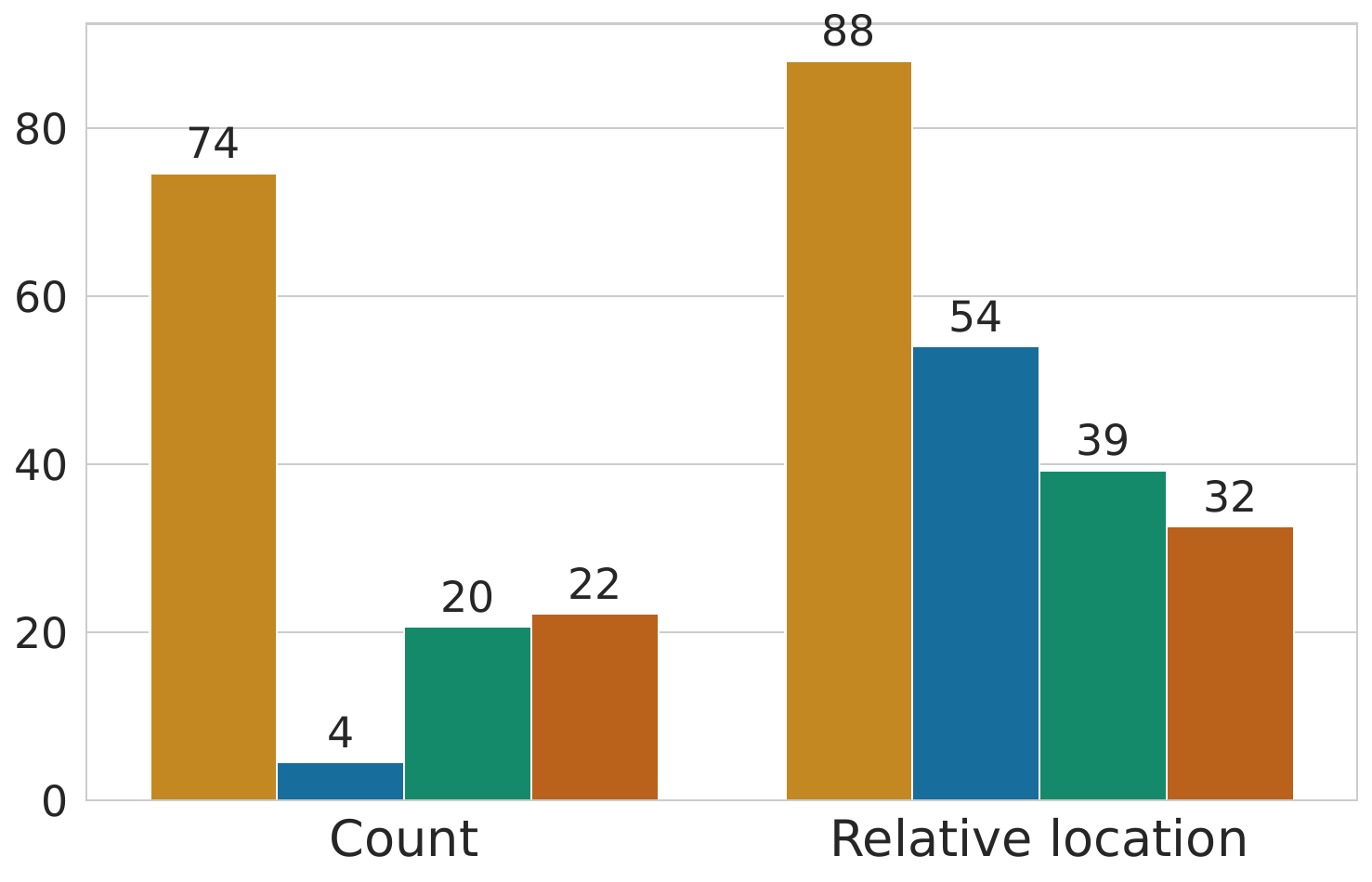}
        \caption{COCO\label{fig:coco}}
    \end{subfigure}
    \hfill
    \begin{subfigure}[b]{0.32\linewidth}
        \includegraphics[width=\linewidth]{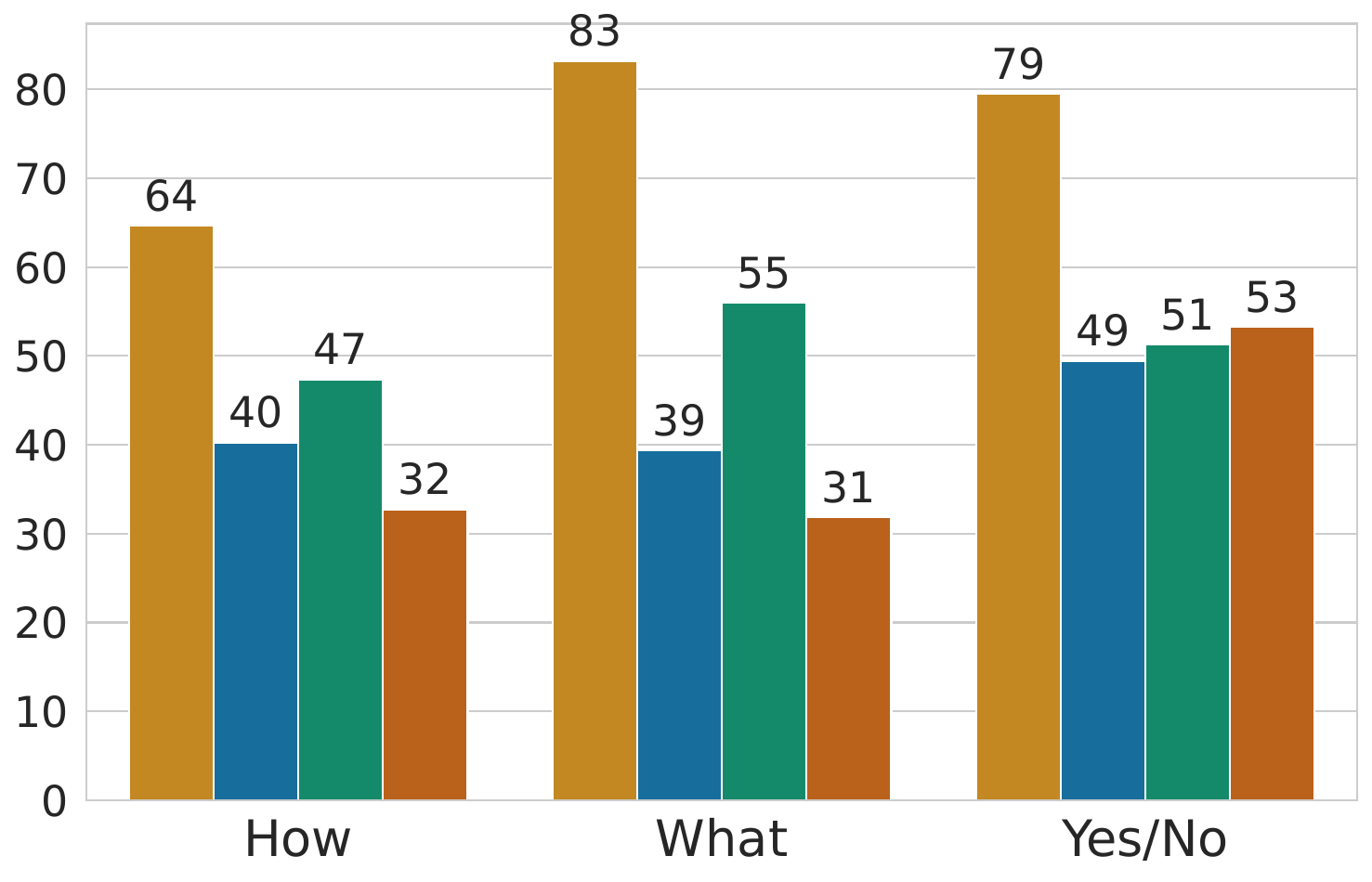}
        \caption{Q-Bench\label{fig:qbench}}
    \end{subfigure}
    
    \begin{subfigure}[b]{0.32\linewidth}
        \includegraphics[width=\linewidth]{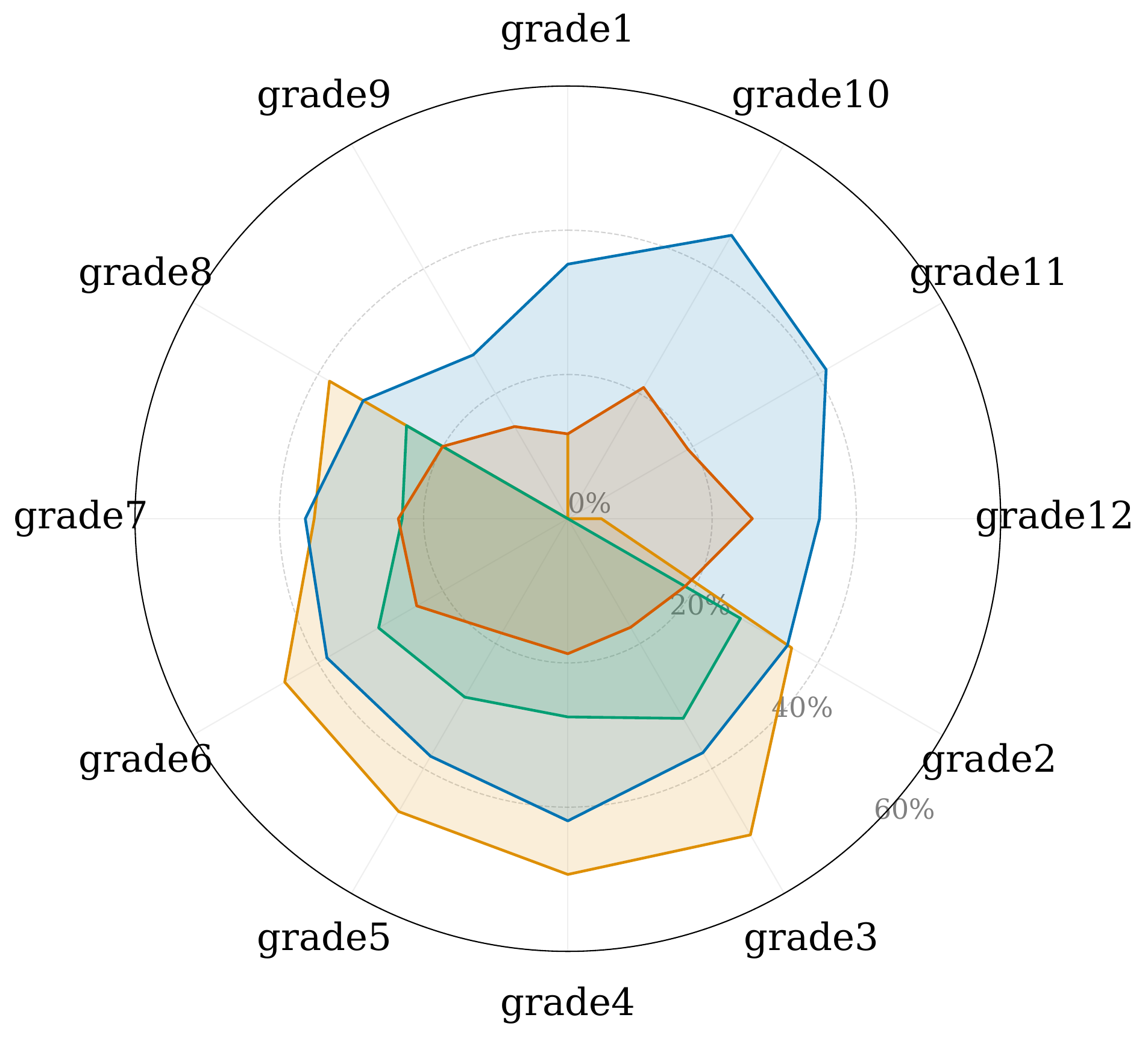}
        \caption{Science QA\label{fig:scienceqa}}
    \end{subfigure}
    \hfill
    \begin{subfigure}[b]{0.32\linewidth}
        \includegraphics[width=\linewidth]{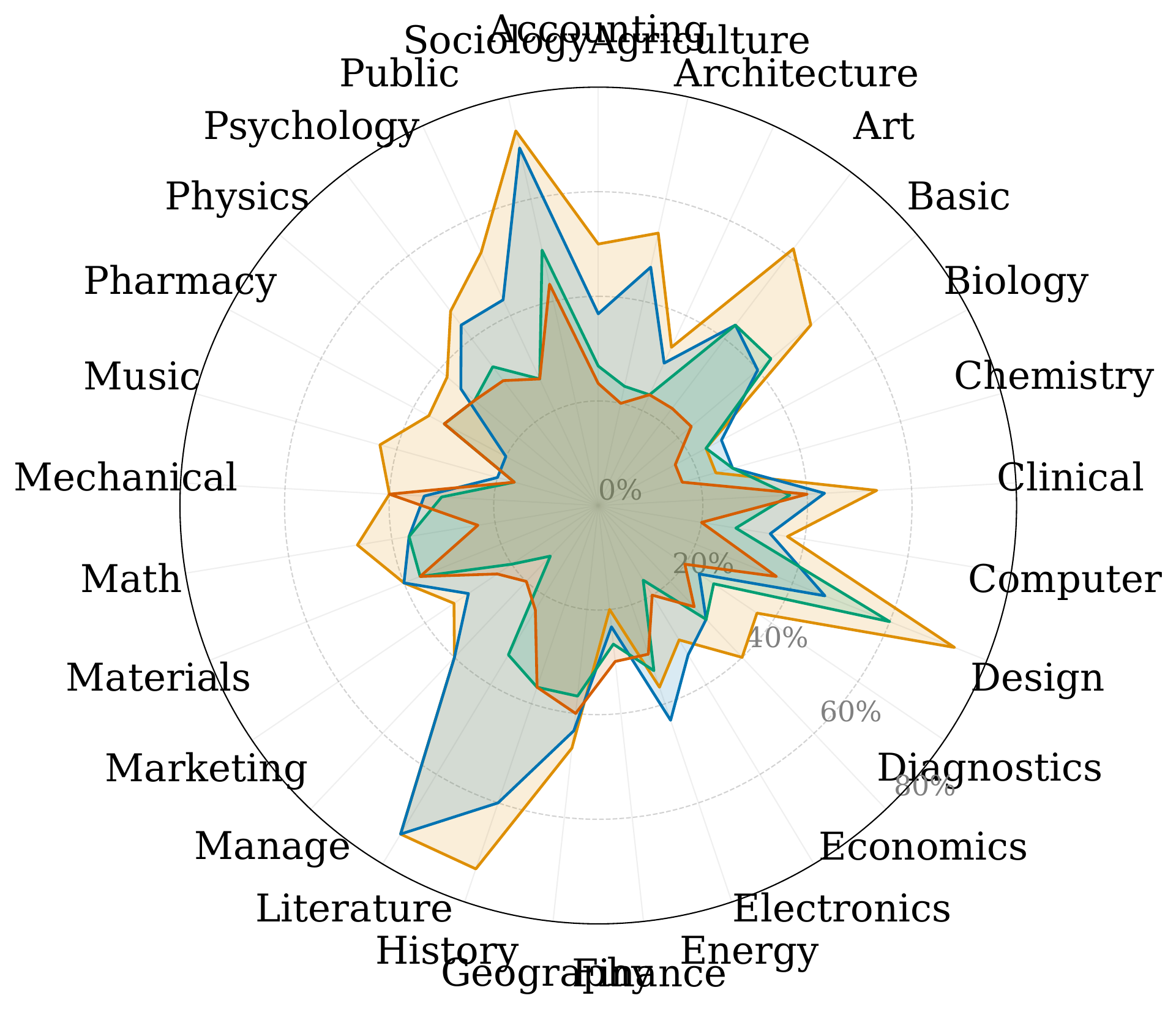}
        \caption{MMMU\label{fig:mmmu}}
    \end{subfigure}
    \hfill
    \begin{subfigure}[b]{0.32\linewidth}
        \includegraphics[width=\linewidth]{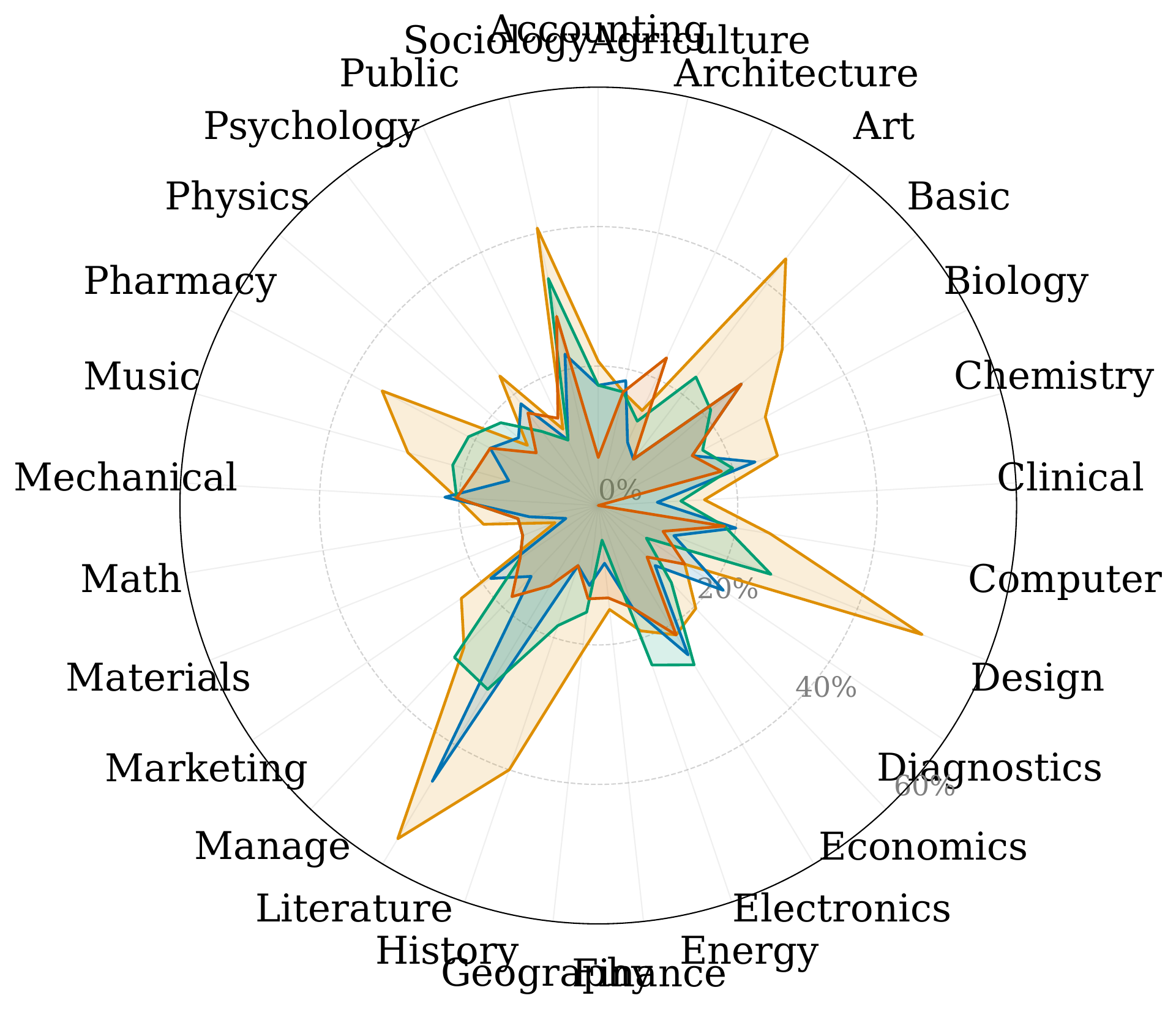}
        \caption{MMMUPro\label{fig:mmmupro}}
    \end{subfigure}
    
    \caption{{\bf Analysis of sub-categories across datasets showing dependency on individual modalities.} Although benchmarks may be designed for inter-modality reasoning, we show a strong dependence on text for categories such as relative location in ADE and COCO, or higher-grade questions in ScienceQA and multiple categories in MMMU and MMMUPro. This highlights how aggregate metrics can obscure that many instances may not require multi-modal reasoning. We show \textcolor{radaryellow}{standard accuracy} in \textcolor{radaryellow}{yellow} and contributions from \textcolor{radarblue}{text} in \textcolor{radarblue}{blue}, \textcolor{radargreen}{image} in \textcolor{radargreen}{green}, and \textcolor{radarorange}{random} in \textcolor{radarorange}{orange}.\label{fig:category_analysis}}
\end{figure*}

Aggregate performance metrics can be misleading, often obscuring strong intra-modality dependencies at the sub-category level. As shown in \Cref{fig:category_analysis}, the benchmarks that appeared to use inter-modality dependencies in \Cref{fig:multimodal_results} also contain intra-modality dependencies when evaluated at a granular level.

This discrepancy is evident across several datasets. In ADE~\citep{zhou2019semantic} and COCO~\citep{lin2014microsoft, tong2024cambrian}, while a text-only model's overall performance is only marginally above chance (see \Cref{fig:multimodal_results}), it achieves substantial accuracy on the relative location sub-category (\Cref{fig:ade,fig:coco}). This phenomenon is amplified in knowledge-intensive benchmarks. In ScienceQA~\citep{lu2022learn} (\Cref{fig:scienceqa}), text-only performance accounts for the majority of the accuracy of questions aimed at grades 10-12. Likewise, many academic subjects within the MMMU and MMMU Pro benchmarks~\citep{yue2024mmmu} (\Cref{fig:mmmu,fig:mmmupro}) contain many instances solvable with a question or an image, respectively, allowing unimodal models to succeed without question or visual information. Conversely, Q-Bench~\citep{wu2023qbench} (\Cref{fig:qbench}) exhibits the opposite pattern. Individual categories show a dependence on both image and text intra-modality dependencies, yet the aggregate metrics in \Cref{fig:multimodal_results} indicate a notable bias toward the image modality.

These findings are corroborated by our analysis of datasets such as MME~\citep{fu2023mme} and BLINK~\citep{fu2024blink} in \Cref{fig:appendix_category_analysis}. We demonstrate that the degree of modality dependence is often inconsistent within a single benchmark. This highlights the multi-dimensional nature of multi-modal datasets, intra- and inter-modality dependencies emerge and vary unpredictably across different sub-populations of the data. 

\section{Limitations and Future Work}\label{sec:future_work}

\begin{figure}[t!] 
    \centering
    \begin{subfigure}{\linewidth}
        \centering
        \includegraphics[width=\linewidth]{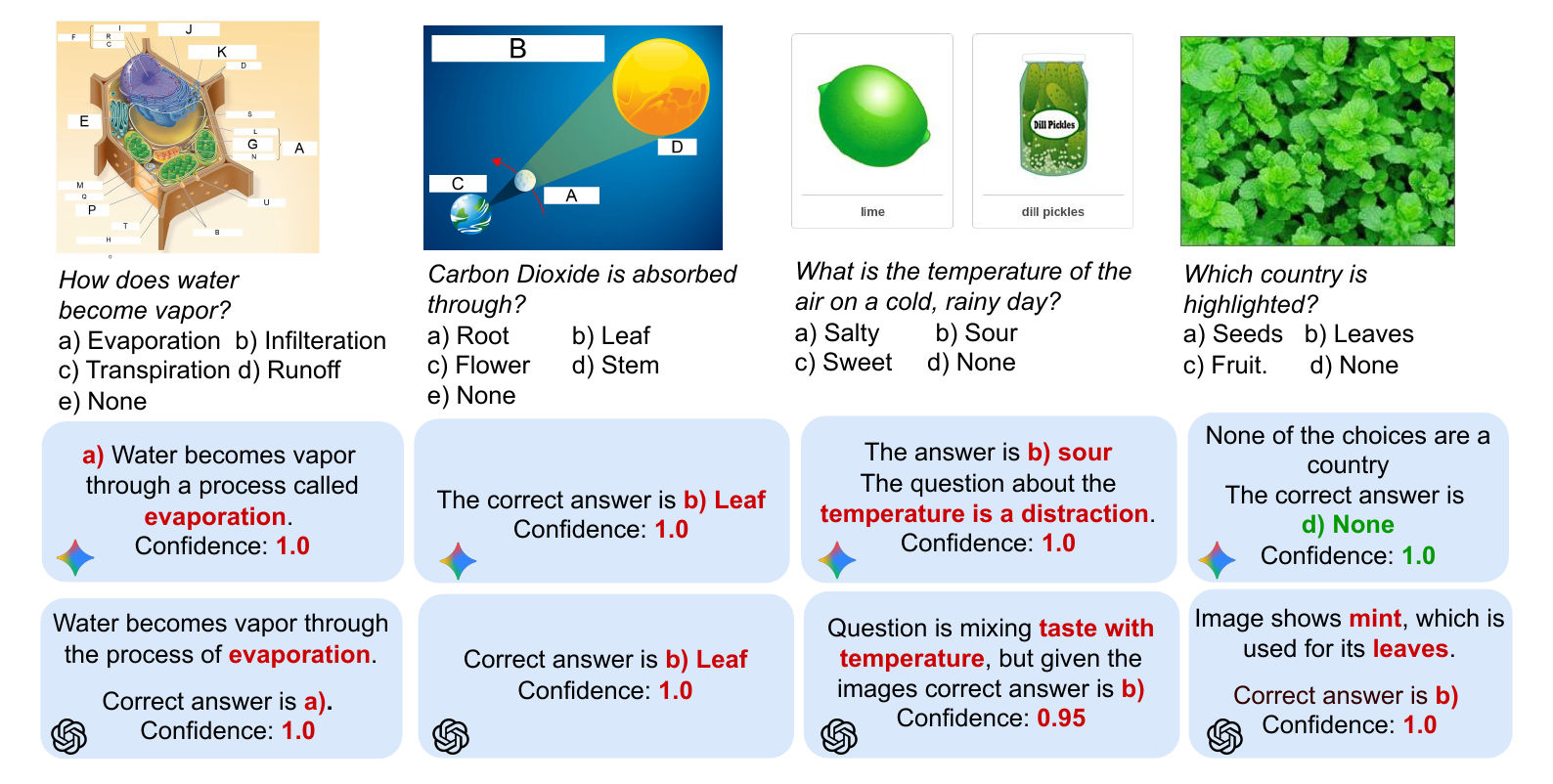}
        \vspace{-0.2in}
    \end{subfigure}
            \caption{{\bf MLLM failure modes in MCVQA.} Visualization from GPT-5 and Gemini 2.5 Pro showing failure modes in MCVQA, such as relying only on text for factual questions while ignoring the image, or conversely, choosing visually correlated answers while ignoring the question. In all cases, the models were prompted to select one choice and provide a confidence score between 0 and 1.\label{fig:evaluation_limitation}}
\end{figure}

Our analysis is constrained by the field's reliance on MCVQA benchmarks. This common practice often fails to test for true multi-modal understanding due to two prevalent failure modes (see \Cref{fig:evaluation_limitation}): text-based intra-modality dependencies, where models ignore the image for factual questions; and image-based intra-modality dependencies, where models select visually correlated answers while disregarding the actual question.

To more holistically evaluate multi-modal capabilities, we propose several crucial future directions. First, we should progress towards building benchmarks that focus on open-ended answer generation and evaluation~\citep{rei2020comet, balepur2025these}. Evaluating free-form responses presents significant challenges. The same meaning can be expressed in many ways, making automated evaluation difficult. This often requires human evaluation, which is slow and expensive. We believe progress in this direction is essential for measuring the necessary multi-modal capabilities.

{Second, both benchmarks and models must support the ability to abstain from answering when presented with ambiguous inputs~\citep{whitehead2022reliable, feng2024teaching, stengel2024lacie}.}
We conduct a preliminary experiment with OpenAI GPT-5 and Google Gemini 2.5 Pro, showing cases where the image or the question was irrelevant to the answer in \Cref{fig:evaluation_limitation}. 
Despite facilitating abstention by augmenting the instruction set with a ``None of the above'' option, this approach is largely insufficient to overcome the dependence on uni-modal dependencies for both the models. This highlights that models have a tendency to generate a plausible-sounding but incorrect response over acknowledging ambiguity or lack of information with confidence. Future work should prioritize methods to encourage meaningful abstention.

Lastly, we encourage future benchmarks and models to report not only aggregated performance but also modality-specific and random baselines to better measure progress. From a benchmark perspective, this helps in understanding how a new dataset compares with existing ones in the inherent biases. From a model perspective, it clarifies where performance gains come from and guides meaningful future improvements.
\section{Conclusion}

Our work critically dissects the intra- and inter-modality dependencies of MLLMs on 23 benchmarks. We show that no dataset is truly multi-modal, as each measures multiple dimensions of multi-modal learning with different strengths of intra- and inter-modality dependence. These strengths vary substantially both across benchmarks and across categories within the same benchmark.
We find that efforts to mitigate text-based dependencies have often introduced new image-based dependencies, perpetuating a cycle of superficial fixes. This suggests that meaningful progress cannot be achieved simply by developing more benchmarks or chasing leaderboard metrics. Instead, we must critically assess the existing evaluation methods. This includes moving beyond standard multiple-choice formats, incorporating scenarios where models should abstain when they are uncertain, and examining how a model arrives at an answer rather than only what answer it produces.

\section*{Acknowledgement}
This work was supported by the Institute of Information \& Communications Technology Planning \& Evaluation (IITP) with a grant funded by the Ministry of Science and ICT (MSIT) of the Republic of Korea in connection with the Global AI Frontier Lab International Collaborative Research, Samsung Advanced Institute of Technology (under the project Next Generation Deep Learning: From Pattern Recognition to AI), National Science Foundation (NSF) award No. 1922658, Center for Advanced Imaging Innovation and Research (CAI2R), National Center for Biomedical Imaging and Bioengineering operated by NYU Langone Health, and National Institute of Biomedical Imaging and Bioengineering through award number P41EB017183. The computational requirements for this work were supported by NYU IT High Performance Computing resources, services, and staff expertise and NYU
Langone High Performance Computing Core’s resources and personnel. This work was partly supported in part by the NYUAD Center for Interdisciplinary Data Science \& AI (CIDSAI), funded by Tamkeen under the NYUAD Research Institute Award CG016. This content is solely the
responsibility of the authors and does not represent the views of the funding agencies.

\bibliography{references}

@string{aaai = "Proceedings of the AAAI National Conference on Artificial Intelligence (AAAI)"}

@string{chi = "Proceedings of the ACM Conference on Human Factors in Computing Systems (CHI)"}

@string{cvpr = "Proceedings of the IEEE International Conference on Computer Vision and Pattern Recognition (CVPR)"}

@string{eccv = "Proceedings of the European Conference on Computer Vision (ECCV)"}

@string{emnlp = "Proceedings of the Conference on Empirical Methods in Natural Language Processing (EMNLP)"}

@string{iccv = "Proceedings of the International Conference on Computer Vision (ICCV)"}

@string{iclr = "Proceedings of the International Conference on Learning Representations (ICLR)"}

@string{icml = "Proceedings of the International Conference on Machine Learning (ICML)"}

@string{nips = "Advances in Neural Information Processing Systems (NeurIPS)"}

@string{computer = "IEEE Computer"}

@string{ijcv = "International Journal of Computer Vision"}

@string{nature = "Nature"}

@string{network = "Network: Computation in Neural Systems"}

@string{science = "Science"}

@inproceedings{singh2019towards,
  title={Towards vqa models that can read},
  author={Singh, Amanpreet and Natarajan, Vivek and Shah, Meet and Jiang, Yu and Chen, Xinlei and Batra, Dhruv and Parikh, Devi and Rohrbach, Marcus},
  booktitle=cvpr,
  year={2019}
}

@article{masry2022chartqa,
  title={Chartqa: A benchmark for question answering about charts with visual and logical reasoning},
  author={Masry, Ahmed and Long, Do Xuan and Tan, Jia Qing and Joty, Shafiq and Hoque, Enamul},
  journal={arXiv preprint arXiv:2203.10244},
  year={2022}
}

@inproceedings{kembhavi2016diagram,
  title={A diagram is worth a dozen images},
  author={Kembhavi, Aniruddha and Salvato, Mike and Kolve, Eric and Seo, Minjoon and Hajishirzi, Hannaneh and Farhadi, Ali},
  booktitle=eccv,
  year={2016},
}

@inproceedings{dietterich2000ensemble,
  title={Ensemble methods in machine learning},
  author={Dietterich, Thomas G},
  booktitle={International workshop on multiple classifier systems},
  year={2000},
}

@inproceedings{gurari2018vizwiz,
  title={Vizwiz grand challenge: Answering visual questions from blind people},
  author={Gurari, Danna and Li, Qing and Stangl, Abigale J and Guo, Anhong and Lin, Chi and Grauman, Kristen and Luo, Jiebo and Bigham, Jeffrey P},
  booktitle=cvpr,
  year={2018}
}

@article{tong2024cambrian,
  title={Cambrian-1: A fully open, vision-centric exploration of multimodal llms},
  author={Tong, Peter and Brown, Ellis and Wu, Penghao and Woo, Sanghyun and IYER, Adithya Jairam Vedagiri and Akula, Sai Charitha and Yang, Shusheng and Yang, Jihan and Middepogu, Manoj and Wang, Ziteng and others},
  journal=nips,
  year={2024}
}

@inproceedings{hudson2019gqa,
  title={Gqa: A new dataset for real-world visual reasoning and compositional question answering},
  author={Hudson, Drew A and Manning, Christopher D},
  booktitle=cvpr,
  year={2019}
}

@article{lu2022learn,
  title={Learn to explain: Multimodal reasoning via thought chains for science question answering},
  author={Lu, Pan and Mishra, Swaroop and Xia, Tanglin and Qiu, Liang and Chang, Kai-Wei and Zhu, Song-Chun and Tafjord, Oyvind and Clark, Peter and Kalyan, Ashwin},
  journal=nips,
  year={2022}
}

@article{bai2025qwen2,
  title={Qwen2. 5-vl technical report},
  author={Bai, Shuai and Chen, Keqin and Liu, Xuejing and Wang, Jialin and Ge, Wenbin and Song, Sibo and Dang, Kai and Wang, Peng and Wang, Shijie and Tang, Jun and others},
  journal={arXiv preprint arXiv:2502.13923},
  year={2025}
}

@article{suhr2019nlvr2,
  title={Nlvr2 visual bias analysis},
  author={Suhr, Alane and Artzi, Yoav},
  journal={arXiv preprint arXiv:1909.10411},
  year={2019}
}

@misc{liu2024llavanext,
    title={LLaVA-NeXT: Improved reasoning, OCR, and world knowledge},
    author={Liu, Haotian and Li, Chunyuan and Li, Yuheng and Li, Bo and Zhang, Yuanhan and Shen, Sheng and Lee, Yong Jae},
    month={January},
    year={2024}
}

@inproceedings{yue2024mmmu,
  title={Mmmu: A massive multi-discipline multimodal understanding and reasoning benchmark for expert agi},
  author={Yue, Xiang and Ni, Yuansheng and Zhang, Kai and Zheng, Tianyu and Liu, Ruoqi and Zhang, Ge and Stevens, Samuel and Jiang, Dongfu and Ren, Weiming and Sun, Yuxuan and others},
  booktitle=cvpr,
  year={2024}
}

@inproceedings{
madaan2024jointly,
title={Jointly Modeling Inter- \& Intra-Modality Dependencies for Multi-modal Learning},
author={Divyam Madaan and Taro Makino and Sumit Chopra and Kyunghyun Cho},
booktitle=nips,
year={2024},
}

@article{liu2024ocrbench,
  title={Ocrbench: on the hidden mystery of ocr in large multimodal models},
  author={Liu, Yuliang and Li, Zhang and Huang, Mingxin and Yang, Biao and Yu, Wenwen and Li, Chunyuan and Yin, Xu-Cheng and Liu, Cheng-Lin and Jin, Lianwen and Bai, Xiang},
  journal={Science China Information Sciences},
  year={2024},
}

@article{li2024survey,
  title={A survey on multimodal benchmarks: In the era of large ai models},
  author={Li, Lin and Chen, Guikun and Shi, Hanrong and Xiao, Jun and Chen, Long},
  journal={arXiv preprint arXiv:2409.18142},
  year={2024}
}

@inproceedings{kornblith2019similarity,
  title={Similarity of neural network representations revisited},
  author={Kornblith, Simon and Norouzi, Mohammad and Lee, Honglak and Hinton, Geoffrey},
  booktitle=icml,
  year={2019},
}

@article{tjandrasuwita2025understanding,
  title={Understanding the emergence of multimodal representation alignment},
  author={Tjandrasuwita, Megan and Ekbote, Chanakya and Ziyin, Liu and Liang, Paul Pu},
  journal={arXiv preprint arXiv:2502.16282},
  year={2025}
}

@article{gat2021perceptual,
  title={Perceptual score: What data modalities does your model perceive?},
  author={Gat, Itai and Schwartz, Idan and Schwing, Alex},
  journal=nips,
  year={2021}
}

@article{parcalabescu2022mm,
  title={Mm-shap: A performance-agnostic metric for measuring multimodal contributions in vision and language models \& tasks},
  author={Parcalabescu, Letitia and Frank, Anette},
  journal={arXiv preprint arXiv:2212.08158},
  year={2022}
}

@article{lu2023theory,
  title={A theory of multimodal learning},
  author={Lu, Zhou},
  journal=nips,
  year={2023}
}

@inproceedings{goyal2017making,
  title={Making the v in vqa matter: Elevating the role of image understanding in visual question answering},
  author={Goyal, Yash and Khot, Tejas and Summers-Stay, Douglas and Batra, Dhruv and Parikh, Devi},
  booktitle=cvpr,
  year={2017}
}

@article{fu2023mme,
  title={MME: A Comprehensive Evaluation Benchmark for Multimodal Large Language Models},
  author={Fu, Chaoyou and Chen, Peixian and Shen, Yunhang and Qin, Yulei and Zhang, Mengdan and Lin, Xu and Yang, Jinrui and Zheng, Xiawu and Li, Ke and Sun, Xing and others},
  journal={arXiv preprint arXiv:2306.13394},
  year={2023}
}

@inproceedings{li2023evaluating,
  title={Evaluating Object Hallucination in Large Vision-Language Models},
  author={Li, Yifan and Du, Yifan and Zhou, Kun and Wang, Jinpeng and Zhao, Wayne Xin and Wen, Ji-Rong},
  booktitle=emnlp,
  year={2023}
}

@misc{grokv2024,
  title={Grok-1.5 Vision Preview},
  author={xAI},
  year={2024},
  url={https://x.ai/blog/grok-1.5v}
}

@misc{commandA2025,
  title={Introducing Command A Vision: Multimodal AI built for business},
  author={Cohere},
  year={2025},
}

@inproceedings{lin2014microsoft,
  title={Microsoft coco: Common objects in context},
  author={Lin, Tsung-Yi and Maire, Michael and Belongie, Serge and Hays, James and Perona, Pietro and Ramanan, Deva and Doll{\'a}r, Piotr and Zitnick, C Lawrence},
  booktitle=eccv,
  year={2014},
}

@misc{vicuna2023,
    title = {Vicuna: An Open-Source Chatbot Impressing GPT-4 with 90\%* ChatGPT Quality},
    author = {Chiang, Wei-Lin and Li, Zhuohan and Lin, Zi and Sheng, Ying and Wu, Zhanghao and Zhang, Hao and Zheng, Lianmin and Zhuang, Siyuan and Zhuang, Yonghao and Gonzalez, Jo seph E. and Stoica, Ion and Xing, Eric P.},
    year = {2023}
}

@article{team2024gemini,
  title={Gemini 1.5: Unlocking multimodal understanding across millions of tokens of context},
  author={Team, Gemini and Georgiev, Petko and Lei, Ving Ian and Burnell, Ryan and Bai, Libin and Gulati, Anmol and Tanzer, Garrett and Vincent, Damien and Pan, Zhufeng and Wang, Shibo and others},
  journal={arXiv preprint arXiv:2403.05530},
  year={2024}
}

@article{comanici2025gemini,
  title={Gemini 2.5: Pushing the frontier with advanced reasoning, multimodality, long context, and next generation agentic capabilities},
  author={Comanici, Gheorghe and Bieber, Eric and Schaekermann, Mike and Pasupat, Ice and Sachdeva, Noveen and Dhillon, Inderjit and Blistein, Marcel and Ram, Ori and Zhang, Dan and Rosen, Evan and others},
  journal={arXiv preprint arXiv:2507.06261},
  year={2025}
}

@inproceedings{fu2024blink,
  title={Blink: Multimodal large language models can see but not perceive},
  author={Fu, Xingyu and Hu, Yushi and Li, Bangzheng and Feng, Yu and Wang, Haoyu and Lin, Xudong and Roth, Dan and Smith, Noah A and Ma, Wei-Chiu and Krishna, Ranjay},
  booktitle=eccv,
  year={2024},
}

@inproceedings{tong2024eyes,
  title={Eyes wide shut? exploring the visual shortcomings of multimodal llms},
  author={Tong, Shengbang and Liu, Zhuang and Zhai, Yuexiang and Ma, Yi and LeCun, Yann and Xie, Saining},
  booktitle=cvpr,
  year={2024}
}

@article{balepur2025these,
  title={Which of these best describes multiple choice evaluation with llms? a) forced b) flawed c) fixable d) all of the above},
  author={Balepur, Nishant and Rudinger, Rachel and Boyd-Graber, Jordan Lee},
  journal={arXiv preprint arXiv:2502.14127},
  year={2025}
}

@inproceedings{rei2020comet,
  title={COMET: A Neural Framework for MT Evaluation},
  author={Rei, Ricardo and Stewart, Craig and Farinha, Ana C and Lavie, Alon},
  booktitle=emnlp,
  year={2020}
}

@article{zhou2019semantic,
  title={Semantic understanding of scenes through the ade20k dataset},
  author={Zhou, Bolei and Zhao, Hang and Puig, Xavier and Xiao, Tete and Fidler, Sanja and Barriuso, Adela and Torralba, Antonio},
  journal=ijcv,
  year={2019},
}

@article{wu2023qbench,
  title={Q-bench: A benchmark for general-purpose foundation models on low-level vision},
  author={Wu, Haoning and Zhang, Zicheng and Zhang, Erli and Chen, Chaofeng and Liao, Liang and Wang, Annan and Li, Chunyi and Sun, Wenxiu and Yan, Qiong and Zhai, Guangtao and others},
  journal={arXiv preprint arXiv:2309.14181},
  year={2023}
}

@article{liang2023quantifying,
  title={Quantifying \& modeling multimodal interactions: An information decomposition framework},
  author={Liang, Paul Pu and Cheng, Yun and Fan, Xiang and Ling, Chun Kai and Nie, Suzanne and Chen, Richard and Deng, Zihao and Allen, Nicholas and Auerbach, Randy and Mahmood, Faisal and others},
  journal=nips,
  year={2023}
}

@article{li2021align,
  title={Align before fuse: Vision and language representation learning with momentum distillation},
  author={Li, Junnan and Selvaraju, Ramprasaath and Gotmare, Akhilesh and Joty, Shafiq and Xiong, Caiming and Hoi, Steven Chu Hong},
  journal=nips,
  year={2021}
}

@inproceedings{wu2022characterizing,
  title={Characterizing and overcoming the greedy nature of learning in multi-modal deep neural networks},
  author={Wu, Nan and Jastrzebski, Stanislaw and Cho, Kyunghyun and Geras, Krzysztof J},
  booktitle=icml,
  year={2022},
}

@article{zheng2023judging,
  title={Judging llm-as-a-judge with mt-bench and chatbot arena},
  author={Zheng, Lianmin and Chiang, Wei-Lin and Sheng, Ying and Zhuang, Siyuan and Wu, Zhanghao and Zhuang, Yonghao and Lin, Zi and Li, Zhuohan and Li, Dacheng and Xing, Eric and others},
  journal=nips,
  year={2023}
}

@article{liu2023visual,
  title={Visual instruction tuning},
  author={Liu, Haotian and Li, Chunyuan and Wu, Qingyang and Lee, Yong Jae},
  journal=nips,
  year={2023}
}

@article{young2024yi,
  title={Yi: Open foundation models by 01. ai},
  author={Young, Alex and Chen, Bei and Li, Chao and Huang, Chengen and Zhang, Ge and Zhang, Guanwei and Wang, Guoyin and Li, Heng and Zhu, Jiangcheng and Chen, Jianqun and others},
  journal={arXiv preprint arXiv:2403.04652},
  year={2024}
}

@inproceedings{antol2015vqa,
  title={Vqa: Visual question answering},
  author={Antol, Stanislaw and Agrawal, Aishwarya and Lu, Jiasen and Mitchell, Margaret and Batra, Dhruv and Zitnick, C Lawrence and Parikh, Devi},
  booktitle=iccv,
  year={2015}
}

@inproceedings{agrawal2018don,
  title={Don't just assume; look and answer: Overcoming priors for visual question answering},
  author={Agrawal, Aishwarya and Batra, Dhruv and Parikh, Devi and Kembhavi, Aniruddha},
  booktitle=cvpr,
  year={2018}
}

@inproceedings{liu2024mmbench,
  title={Mmbench: Is your multi-modal model an all-around player?},
  author={Liu, Yuan and Duan, Haodong and Zhang, Yuanhan and Li, Bo and Zhang, Songyang and Zhao, Wangbo and Yuan, Yike and Wang, Jiaqi and He, Conghui and Liu, Ziwei and others},
  booktitle=eccv,
  year={2024},
}

@article{li2023seed,
  title={Seed-bench: Benchmarking multimodal llms with generative comprehension},
  author={Li, Bohao and Wang, Rui and Wang, Guangzhi and Ge, Yuying and Ge, Yixiao and Shan, Ying},
  journal={arXiv preprint arXiv:2307.16125},
  year={2023}
}

@article{lu2023mathvista,
  title={Mathvista: Evaluating mathematical reasoning of foundation models in visual contexts},
  author={Lu, Pan and Bansal, Hritik and Xia, Tony and Liu, Jiacheng and Li, Chunyuan and Hajishirzi, Hannaneh and Cheng, Hao and Chang, Kai-Wei and Galley, Michel and Gao, Jianfeng},
  journal={arXiv preprint arXiv:2310.02255},
  year={2023}
}

@inproceedings{brazil2023omni3d,
  title={Omni3d: A large benchmark and model for 3d object detection in the wild},
  author={Brazil, Garrick and Kumar, Abhinav and Straub, Julian and Ravi, Nikhila and Johnson, Justin and Gkioxari, Georgia},
  booktitle=cvpr,
  year={2023}
}

@inproceedings{
wu2024qbench,
title={Q-Bench: A Benchmark for General-Purpose Foundation Models on Low-level Vision},
author={Haoning Wu and Zicheng Zhang and Erli Zhang and Chaofeng Chen and Liang Liao and Annan Wang and Chunyi Li and Wenxiu Sun and Qiong Yan and Guangtao Zhai and Weisi Lin},
booktitle=iclr,
year={2024},
}

@inproceedings{wu2024v,
  title={V$^*$: Guided visual search as a core mechanism in multimodal llms},
  author={Wu, Penghao and Xie, Saining},
  booktitle=cvpr,
  year={2024}
}

@article{chen2024we,
  title={Are we on the right way for evaluating large vision-language models?},
  author={Chen, Lin and Li, Jinsong and Dong, Xiaoyi and Zhang, Pan and Zang, Yuhang and Chen, Zehui and Duan, Haodong and Wang, Jiaqi and Qiao, Yu and Lin, Dahua and others},
  journal=nips,
  year={2024}
}

@inproceedings{dancette2021beyond,
  title={Beyond question-based biases: Assessing multimodal shortcut learning in visual question answering},
  author={Dancette, Corentin and Cadene, Remi and Teney, Damien and Cord, Matthieu},
  booktitle=iccv,
  year={2021}
}

@inproceedings{
zhang2024understanding,
title={Understanding Unimodal Bias in Multimodal Deep Linear Networks},
author={Yedi Zhang and Peter E. Latham and Andrew M Saxe},
booktitle=icml,
year={2024},
}

@inproceedings{huang2022modality,
  title={Modality competition: What makes joint training of multi-modal network fail in deep learning?(provably)},
  author={Huang, Yu and Lin, Junyang and Zhou, Chang and Yang, Hongxia and Huang, Longbo},
  booktitle=icml,
  year={2022},
}

@inproceedings{wang2020makes,
  title={What makes training multi-modal classification networks hard?},
  author={Wang, Weiyao and Tran, Du and Feiszli, Matt},
  booktitle=cvpr,
  year={2020}
}

@inproceedings{zhang2024multimodal,
  title={Multimodal representation learning by alternating unimodal adaptation},
  author={Zhang, Xiaohui and Yoon, Jaehong and Bansal, Mohit and Yao, Huaxiu},
  booktitle=cvpr,
  year={2024}
}

@inproceedings{si2022language,
  title={Language Prior Is Not the Only Shortcut: A Benchmark for Shortcut Learning in VQA},
  author={Si, Qingyi and Meng, Fandong and Zheng, Mingyu and Lin, Zheng and Liu, Yuanxin and Fu, Peng and Cao, Yanan and Wang, Weiping and Zhou, Jie},
  booktitle=emnlp,
  year={2022}
}

@inproceedings{zhai2023sigmoid,
  title={Sigmoid loss for language image pre-training},
  author={Zhai, Xiaohua and Mustafa, Basil and Kolesnikov, Alexander and Beyer, Lucas},
  booktitle=cvpr,
  year={2023}
}

@inproceedings{radford2021learning,
  title={Learning transferable visual models from natural language supervision},
  author={Radford, Alec and Kim, Jong Wook and Hallacy, Chris and Ramesh, Aditya and Goh, Gabriel and Agarwal, Sandhini and Sastry, Girish and Askell, Amanda and Mishkin, Pamela and Clark, Jack and others},
  booktitle=icml,
  year={2021},
}

@inproceedings{feng2024teaching,
  title={Teaching LLMs to Abstain across Languages via Multilingual Feedback},
  author={Feng, Shangbin and Shi, Weijia and Wang, Yike and Ding, Wenxuan and Ahia, Orevaoghene and Li, Shuyue Stella and Balachandran, Vidhisha and Sitaram, Sunayana and Tsvetkov, Yulia},
  booktitle=emnlp,
  year={2024}
}

@inproceedings{whitehead2022reliable,
  title={Reliable visual question answering: Abstain rather than answer incorrectly},
  author={Whitehead, Spencer and Petryk, Suzanne and Shakib, Vedaad and Gonzalez, Joseph and Darrell, Trevor and Rohrbach, Anna and Rohrbach, Marcus},
  booktitle=eccv,
  year={2022},
  organization={Springer}
}

@misc{
yue2025mmmupro,
title={{MMMU}-Pro: A More Robust  Multi-discipline Multimodal Understanding Benchmark},
author={Xiang Yue and Tianyu Zheng and Yuansheng Ni and Yubo Wang and Kai Zhang and Shengbang Tong and Yuxuan Sun and Botao Yu and Ge Zhang and Huan Sun and Yu Su and Wenhu Chen and Graham Neubig},
year={2025},
}

@article{stengel2024lacie,
  title={LACIE: Listener-aware finetuning for calibration in large language models},
  author={Stengel-Eskin, Elias and Hase, Peter and Bansal, Mohit},
  journal=nips,
  year={2024}
}

@inproceedings{liu2022convnet,
  title={A convnet for the 2020s},
  author={Liu, Zhuang and Mao, Hanzi and Wu, Chao-Yuan and Feichtenhofer, Christoph and Darrell, Trevor and Xie, Saining},
  booktitle={Proceedings of the IEEE/CVF conference on computer vision and pattern recognition},
  pages={11976--11986},
  year={2022}
}

@article{oquabdinov2,
  title={DINOv2: Learning Robust Visual Features without Supervision},
  author={Oquab, Maxime and Darcet, Timoth{\'e}e and Moutakanni, Th{\'e}o and Vo, Huy V and Szafraniec, Marc and Khalidov, Vasil and Fernandez, Pierre and HAZIZA, Daniel and Massa, Francisco and El-Nouby, Alaaeldin and others},
year=2024,
  journal={Transactions on Machine Learning Research}
}

@inproceedings{wenderoth2025measuring,
  title={Measuring cross-modal interactions in multimodal models},
  author={Wenderoth, Laura and Hemker, Konstantin and Simidjievski, Nikola and Jamnik, Mateja},
  booktitle=aaai,
  year={2025}
}

@article{hu2022shape,
  title={Shape: An unified approach to evaluate the contribution and cooperation of individual modalities},
  author={Hu, Pengbo and Li, Xingyu and Zhou, Yi},
  journal={arXiv preprint arXiv:2205.00302},
  year={2022}
}

@article{gu_illusion_2025,
  title={The Illusion of Readiness in Health AI},
  author={Gu, Yu and Fu, Jingjing and Liu, Xiaodong and Valanarasu, Jeya Maria Jose and Codella, Noel CF and Tan, Reuben and Liu, Qianchu and Jin, Ying and Zhang, Sheng and Wang, Jinyu and others},
  journal={arXiv preprint arXiv:2509.18234},
  year={2025}
}

@article{bai_qwen3-vl_2025,
  title={Qwen3-vl technical report},
  author={Bai, Shuai and Cai, Yuxuan and Chen, Ruizhe and Chen, Keqin and Chen, Xionghui and Cheng, Zesen and Deng, Lianghao and Ding, Wei and Gao, Chang and Ge, Chunjiang and others},
  journal={arXiv preprint arXiv:2511.21631},
  year={2025}
}
\bibliographystyle{abbrvnat}

\newpage
\renewcommand\thefigure{\thesection.\arabic{figure}}    
\renewcommand\thetable{\thesection.\arabic{table}}
\appendix

\section*{Appendix}
\paragraph{Organization} We provide the implementation details in \Cref{appendix:experimental_details} and additional results in \Cref{appendix:additional_results}.

\section{Experimental Details}\label{appendix:experimental_details}

\paragraph{Implementations.} We use the Cambrian-1~\citep{tong2024cambrian} open-sourced codebase for all the experiments. We use their publicly released models for evaluation. Datasets like AI2D~\citep{kembhavi2016diagram}, ChartQA~\citep{masry2022chartqa}, MMBench~\citep{liu2024mmbench}, MME~\citep{fu2023mme}, MMMU~\citep{yue2024mmmu}, POPE~\citep{li2023evaluating}, RealWorldQA~\citep{grokv2024}, SEED~\citep{li2023seed}, TextVQA~\citep{singh2019towards}, and VizWiz~\citep{gurari2018vizwiz} were sourced from LMMS-eval, while others such as ADE, Blink~\citep{fu2024blink}, COCO~\citep{lin2014microsoft}, GQA~\citep{hudson2019gqa}, MathVista~\citep{lu2023mathvista}, MMMUPro~\citep{yue2024mmmu}, MMStar~\citep{chen2024we}, MMVP~\citep{tong2024eyes}, OCRBench~\citep{liu2024ocrbench}, Omni3D~\citep{brazil2023omni3d, tong2024cambrian}, Q-Bench~\citep{wu2024qbench}, ScienceQA~\citep{lu2022learn}, and $V^* $Bench~\citep{wu2024v} were used from their respective sources.

\section{Additional Results}\label{appendix:additional_results}
\begin{figure*}[h!]
    \centering
    \begin{subfigure}[b]{0.49\linewidth}
        \includegraphics[width=\linewidth]{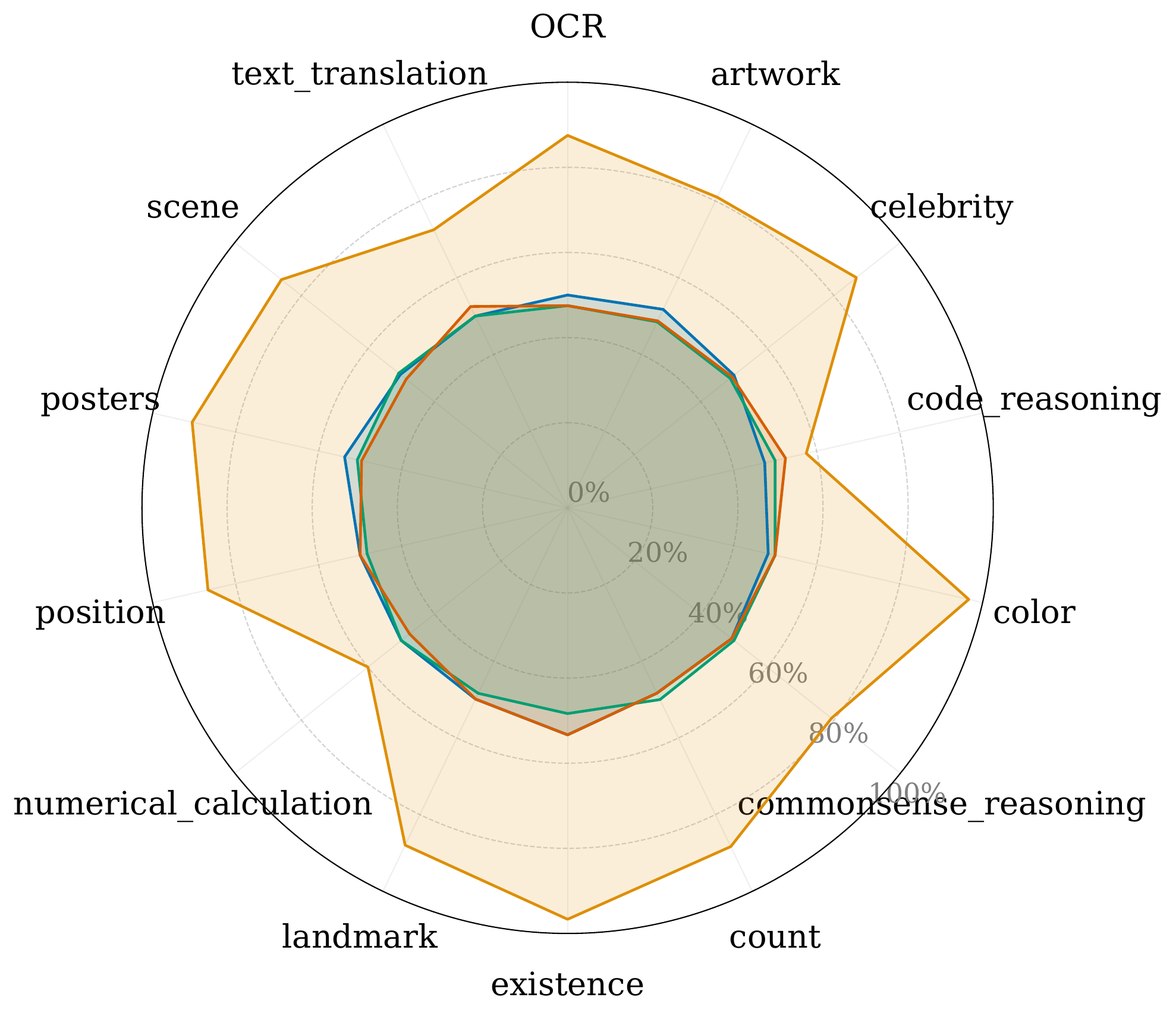}
        \caption{MME}
    \end{subfigure}
    % \hfill
    \begin{subfigure}[b]{0.49\linewidth}
        \includegraphics[width=\linewidth]{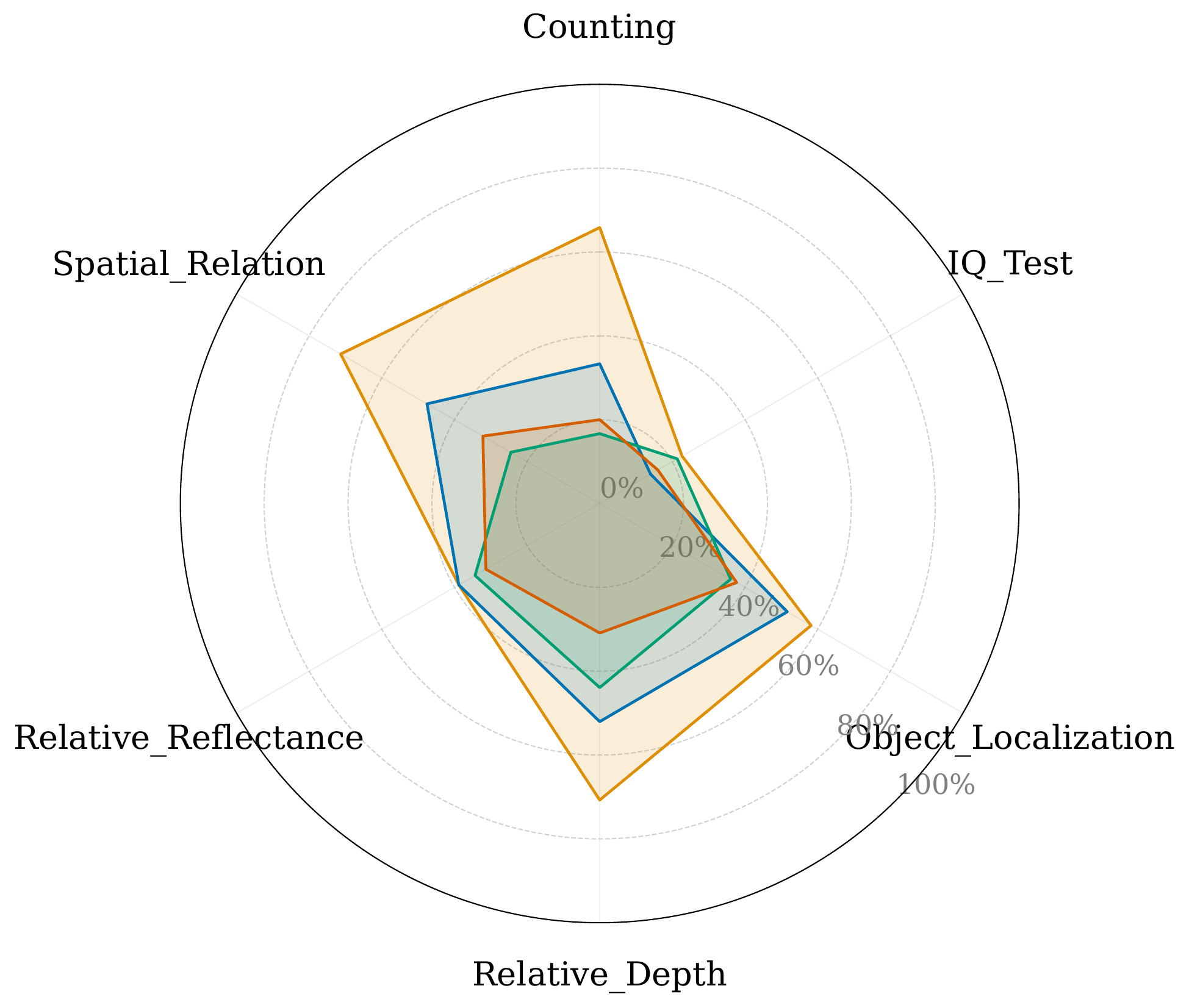}
        \caption{BLINK}
    \end{subfigure}
    \hfill
    \caption{Analysis of sub-categories for MME and BLINK dataset.\label{fig:appendix_category_analysis}}
\end{figure*}

\begin{figure*}[t]
    \centering
    \begin{subfigure}[b]{0.49\linewidth}
        \includegraphics[width=\linewidth]{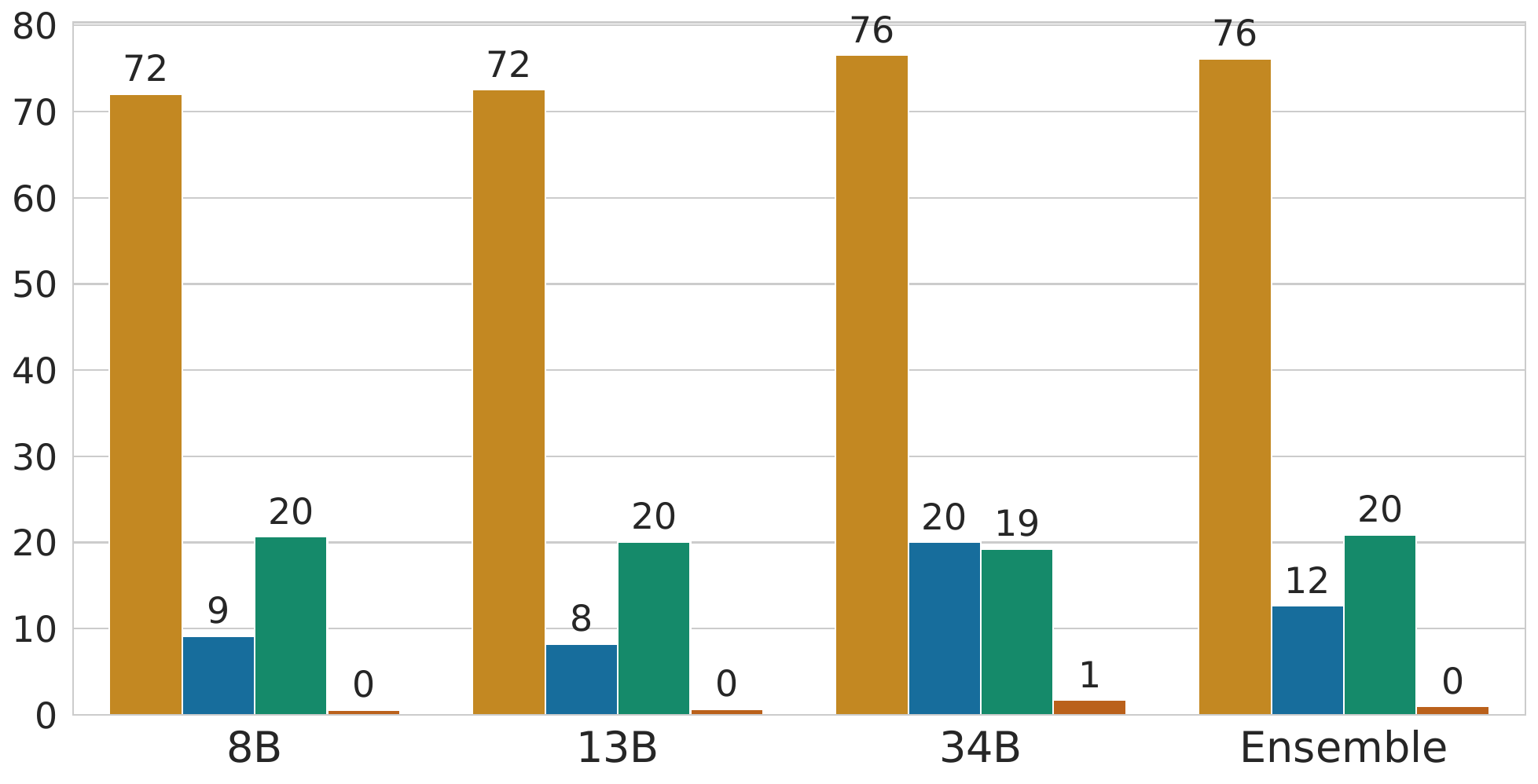}
        \caption{TextVQA}
    \end{subfigure}
    \begin{subfigure}[b]{0.49\linewidth}
        \includegraphics[width=\linewidth]{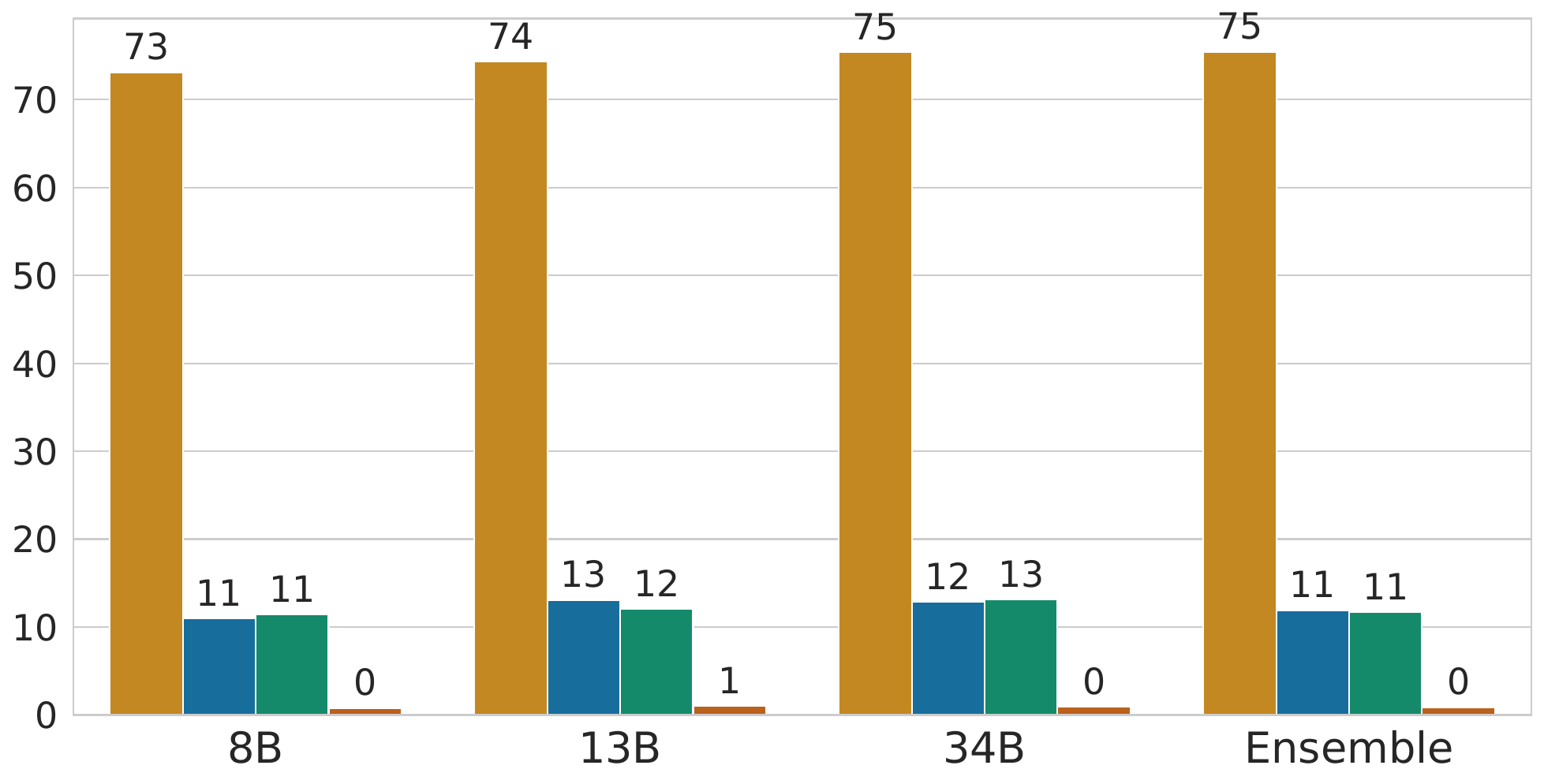}
        \caption{ChartQA}
    \end{subfigure}
    \hfill
    \begin{subfigure}[b]{0.49\linewidth}
        \includegraphics[width=\linewidth]{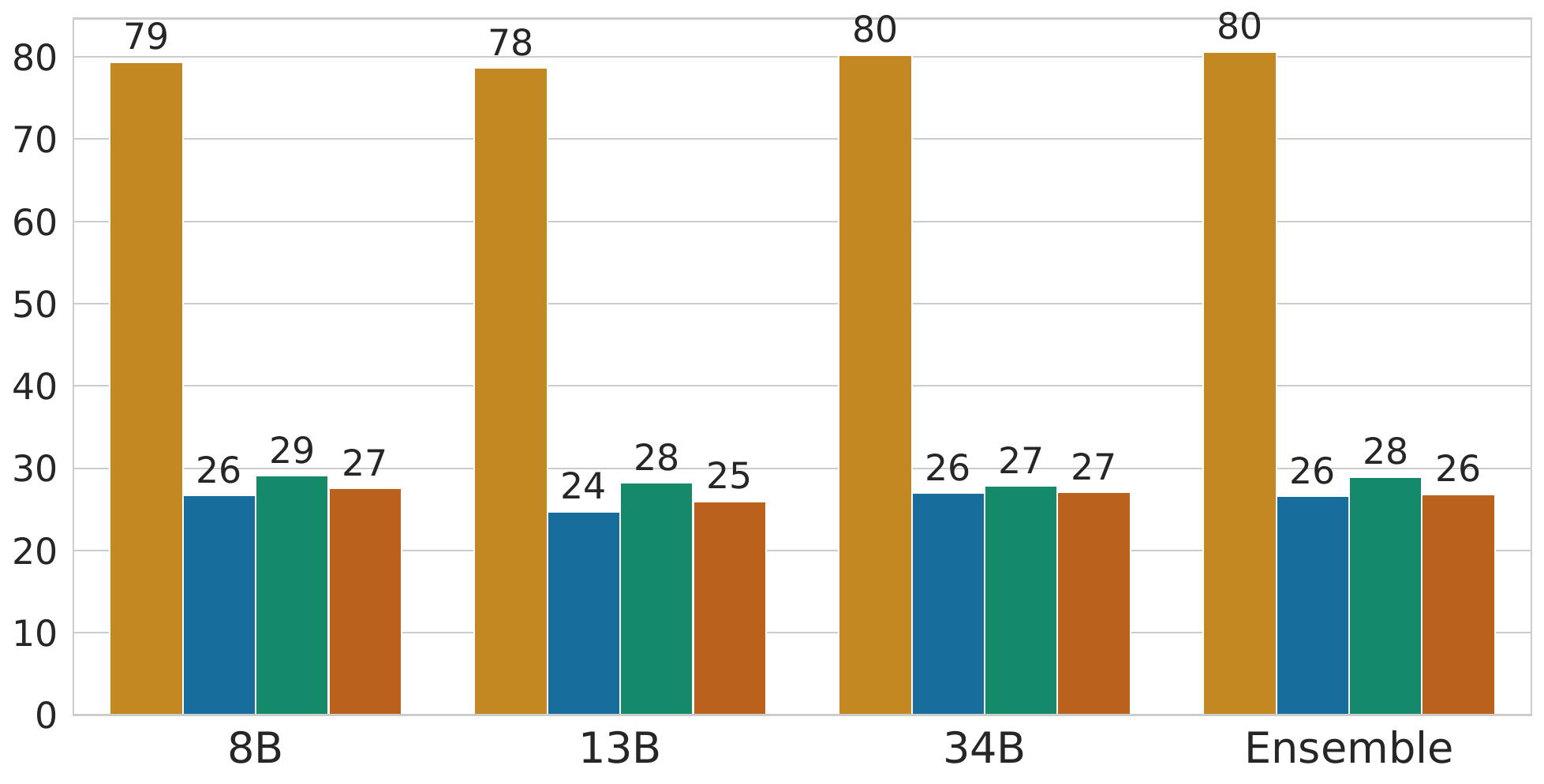}
        \caption{COCO}
    \end{subfigure}
        \hfill
    \begin{subfigure}[b]{0.49\linewidth}
        \includegraphics[width=\linewidth]{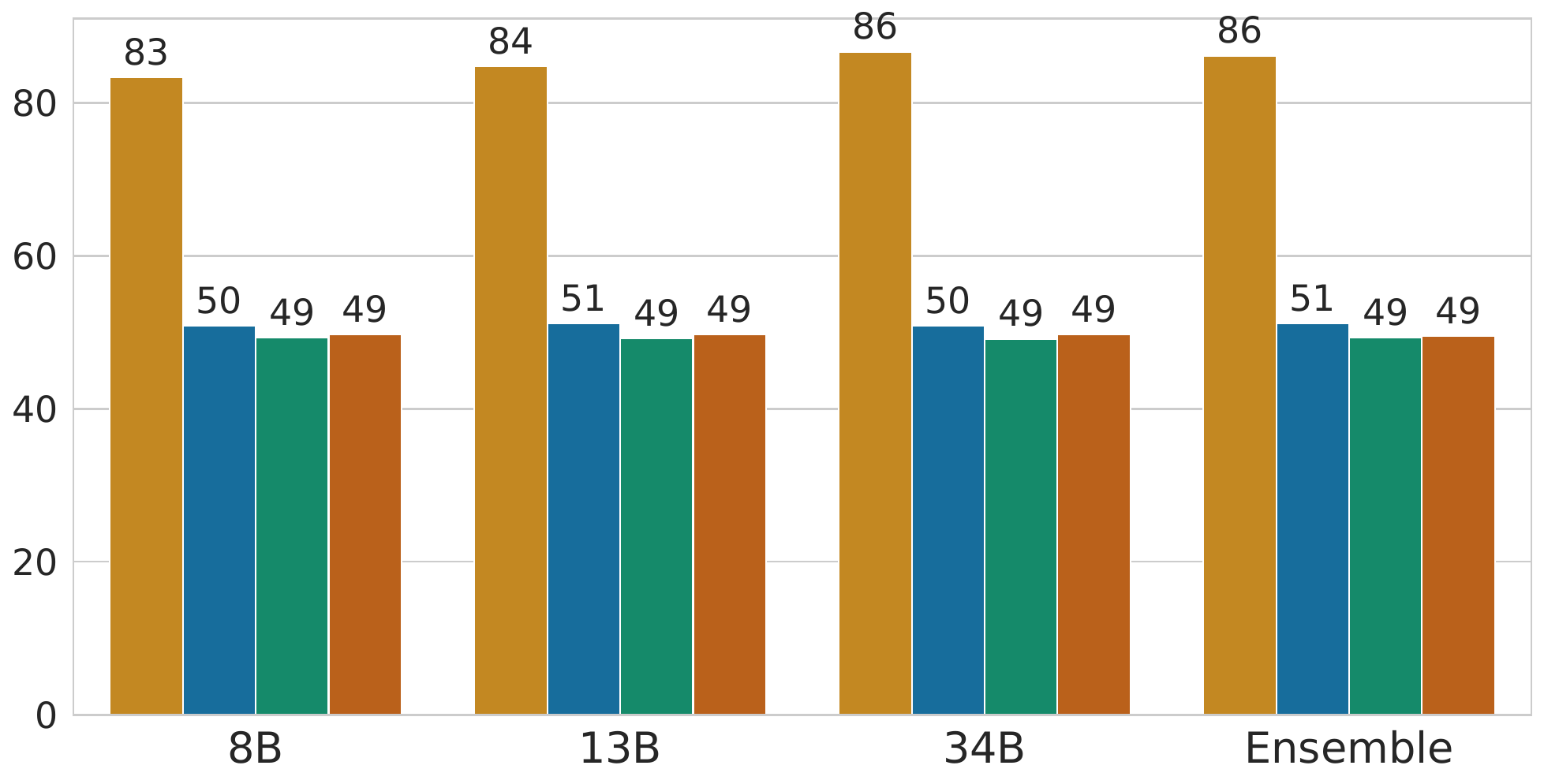}
        \caption{MME}
    \end{subfigure}
        \hfill
    \caption{{\bf Effect of model size on additional datasets.} Performance of various models (8B, 13B, 34B, and a majority-vote ensemble) with both image and text intra-modality dependencies (top) and primarily inter-modality dependency (bottom). The bars represent \textcolor{radaryellow}{standard accuracy} and contributions from \textcolor{radarblue}{text}, \textcolor{radargreen}{image}, and \textcolor{radarorange}{random} (bars are in the same order). \label{fig:appendix_model_size}}
\end{figure*}

\begin{figure*}[t]
    \centering
    \hfill
    \begin{subfigure}[b]{0.49\linewidth}
        \includegraphics[width=\linewidth]{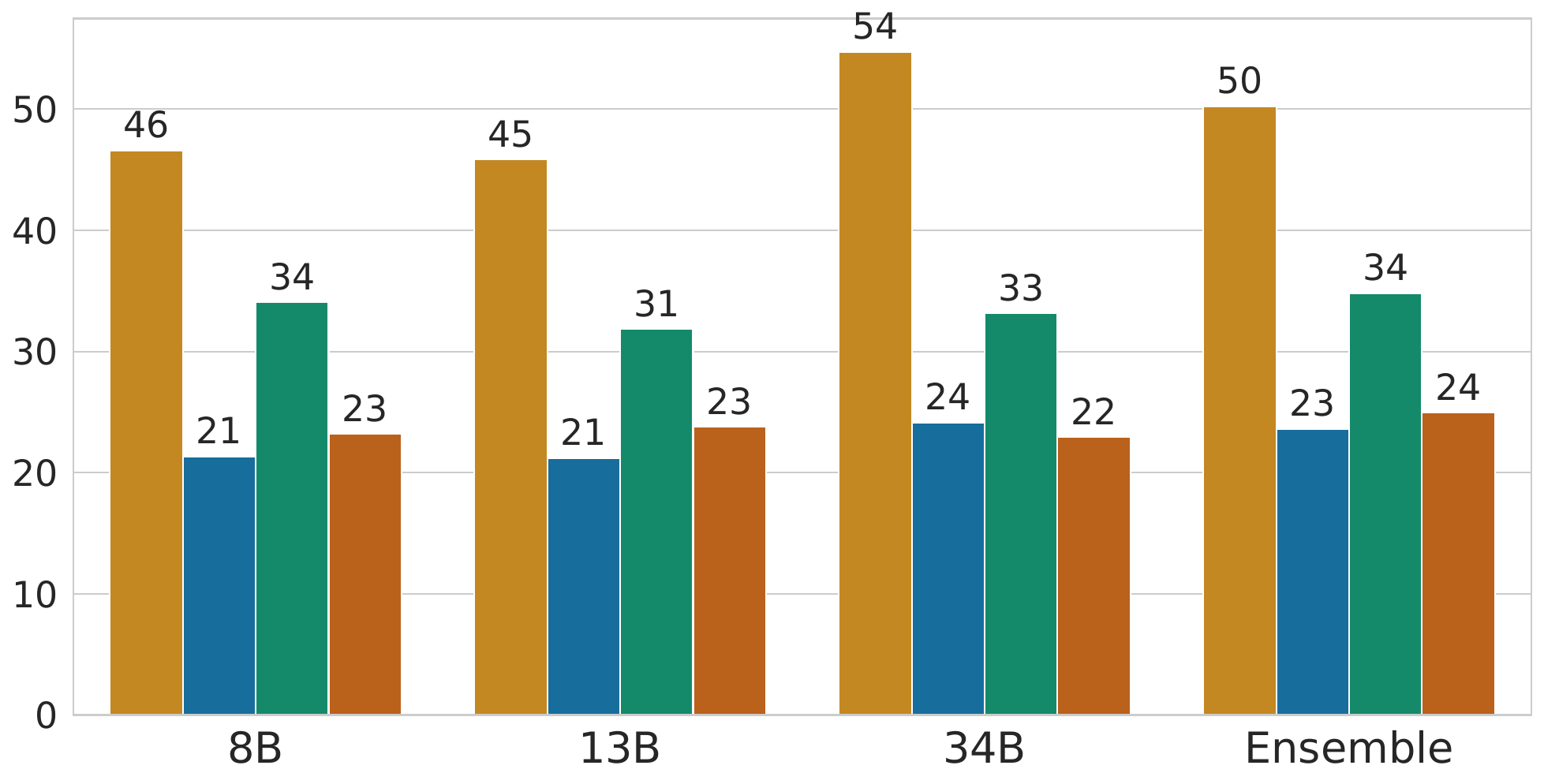}
        \caption{MMStar}
    \end{subfigure}
    \hfill
    \begin{subfigure}[b]{0.49\linewidth}
        \includegraphics[width=\linewidth]{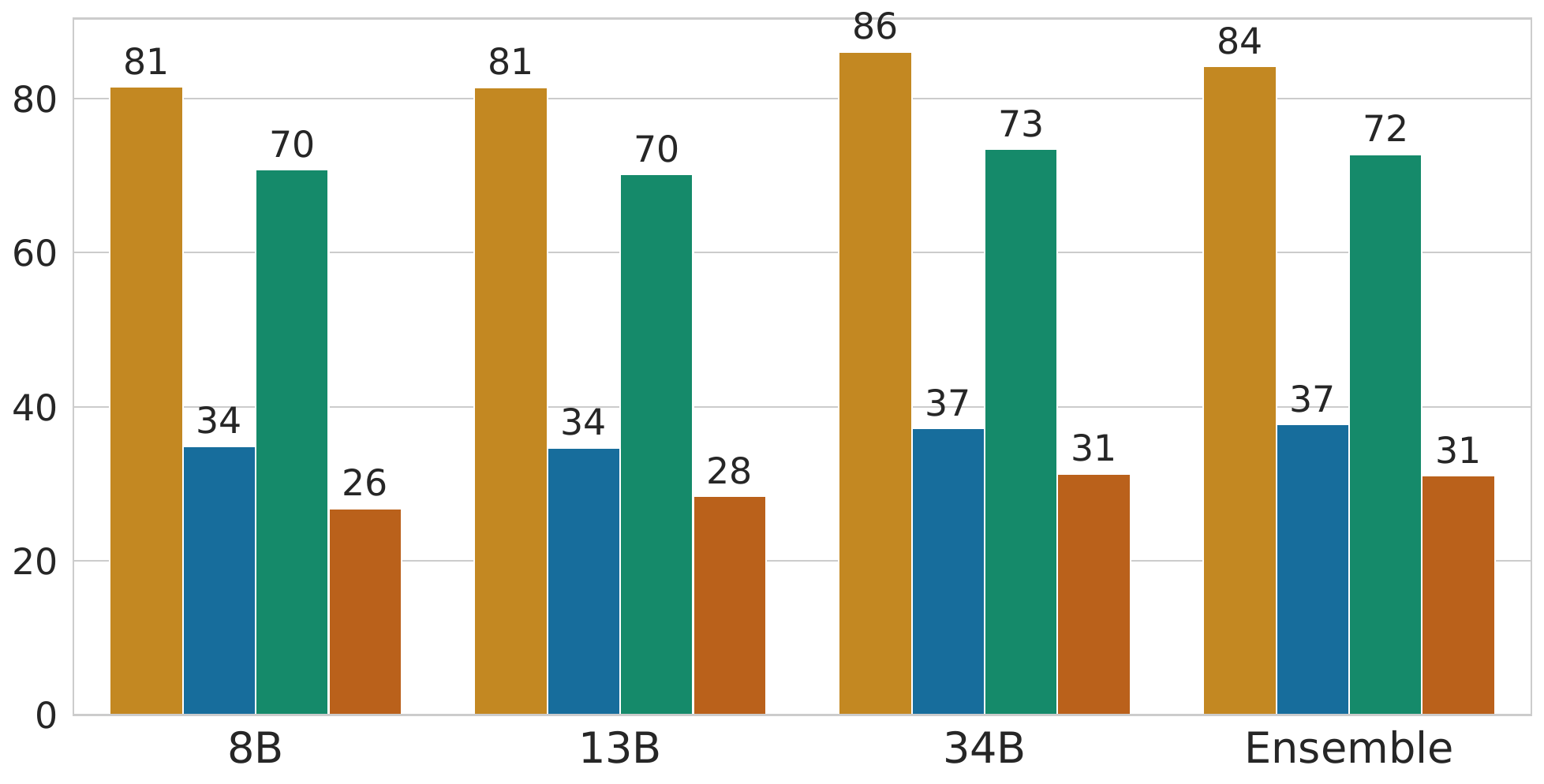}
        \caption{MMBench (en)}
    \end{subfigure}
    \hfill
    \begin{subfigure}[b]{0.49\linewidth}
        \includegraphics[width=\linewidth]{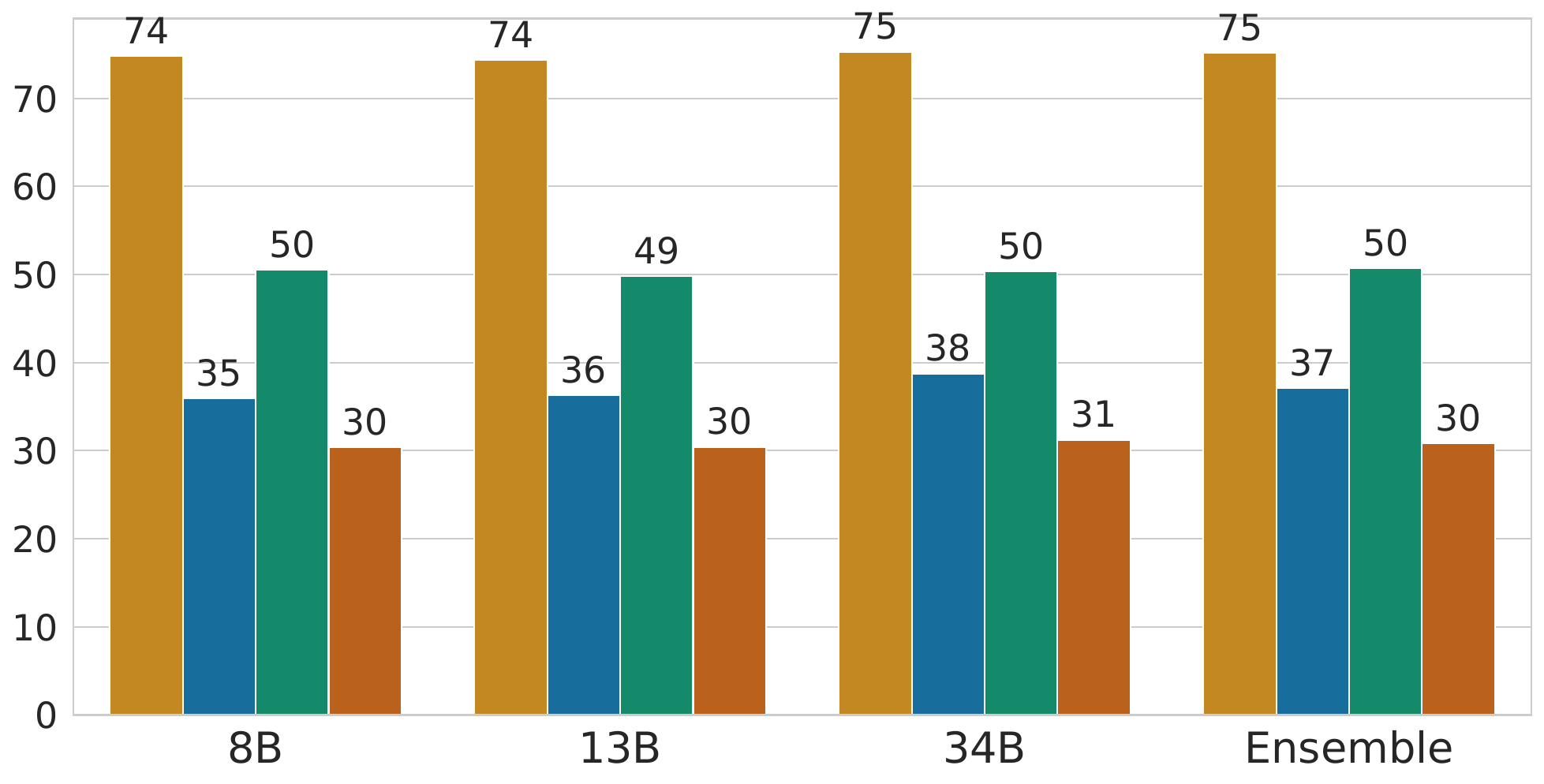}
        \caption{SEED-Bench}
    \end{subfigure}
        \begin{subfigure}[b]{0.49\linewidth}
        \includegraphics[width=\linewidth]{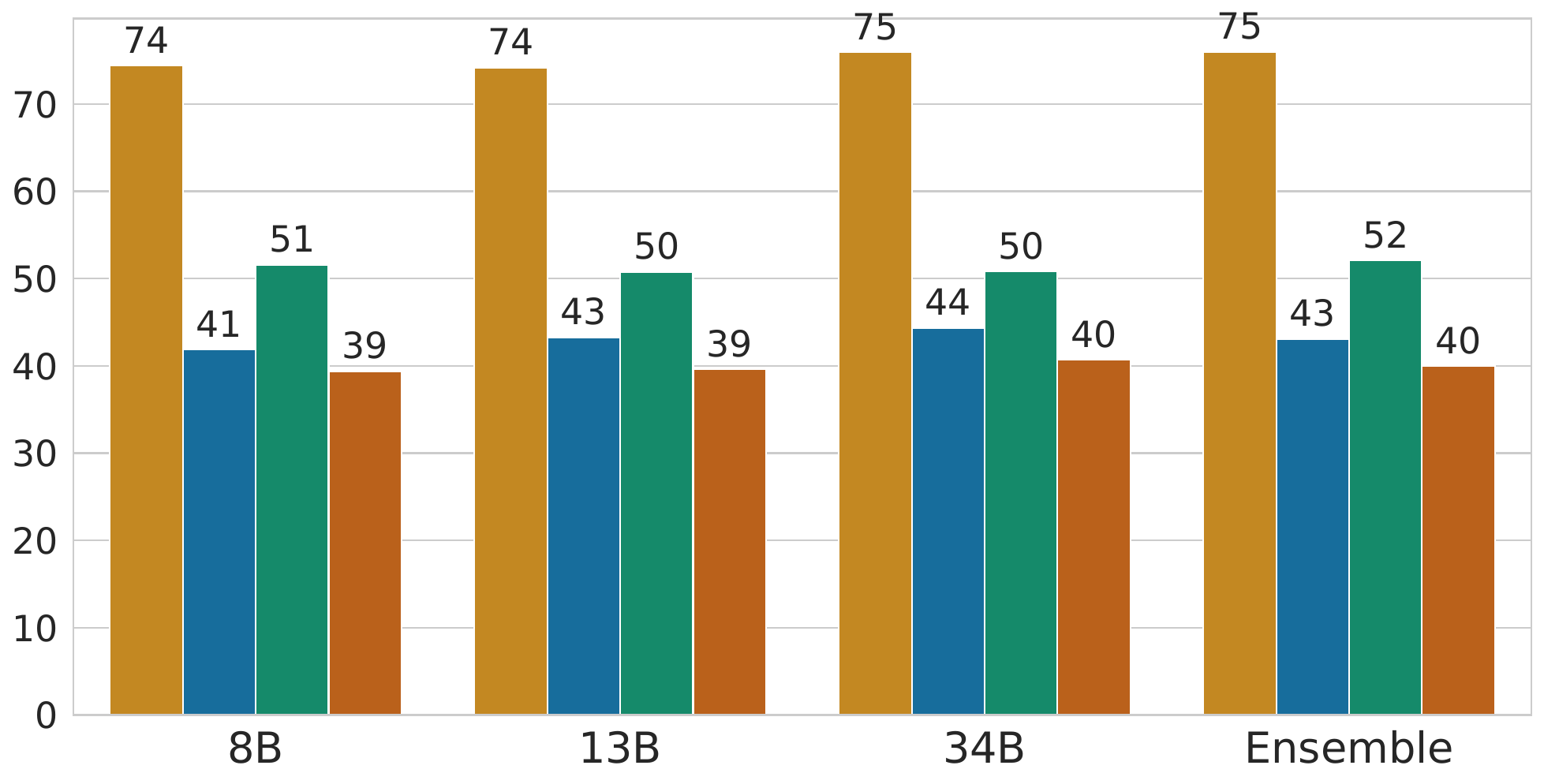}
        \caption{Q-Bench}
    \end{subfigure}
        \hfill
    \caption{{\bf Effect of model size on additional datasets with image intra-modality dependency.} Performance of various models (8B, 13B, 34B, and a majority-vote ensemble). The bars represent \textcolor{radaryellow}{standard accuracy} and contributions from \textcolor{radarblue}{text}, \textcolor{radargreen}{image}, and \textcolor{radarorange}{random} (bars are in the same order).\label{fig:appendix_image_model_size}}

\end{figure*}

\begin{figure*}[t]
    \centering
        \begin{subfigure}[b]{0.49\linewidth}
        \includegraphics[width=\linewidth]{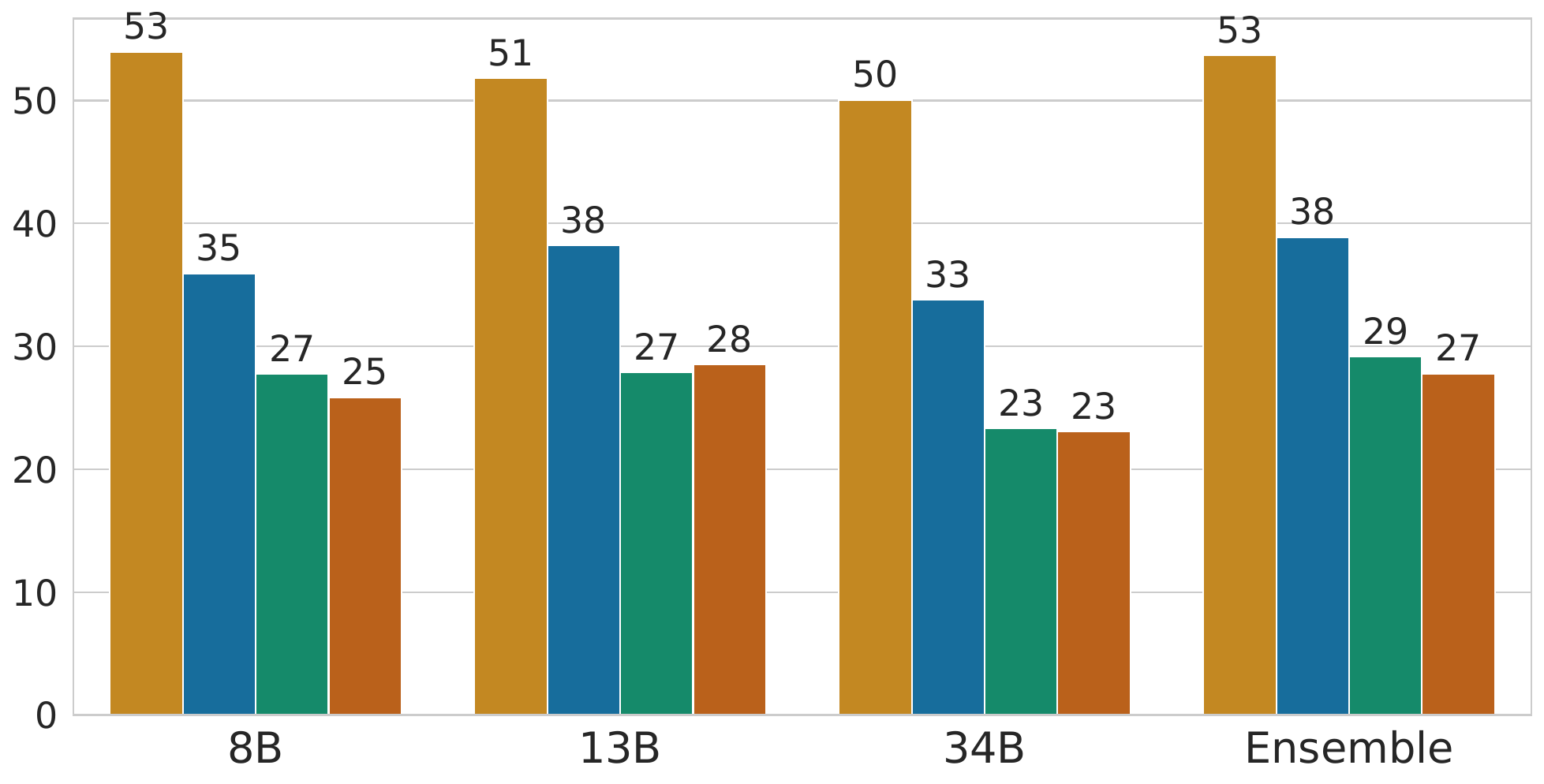}
        \caption{BLINK}
    \end{subfigure}
    \hfill
    \begin{subfigure}[b]{0.49\linewidth}
        \includegraphics[width=\linewidth]{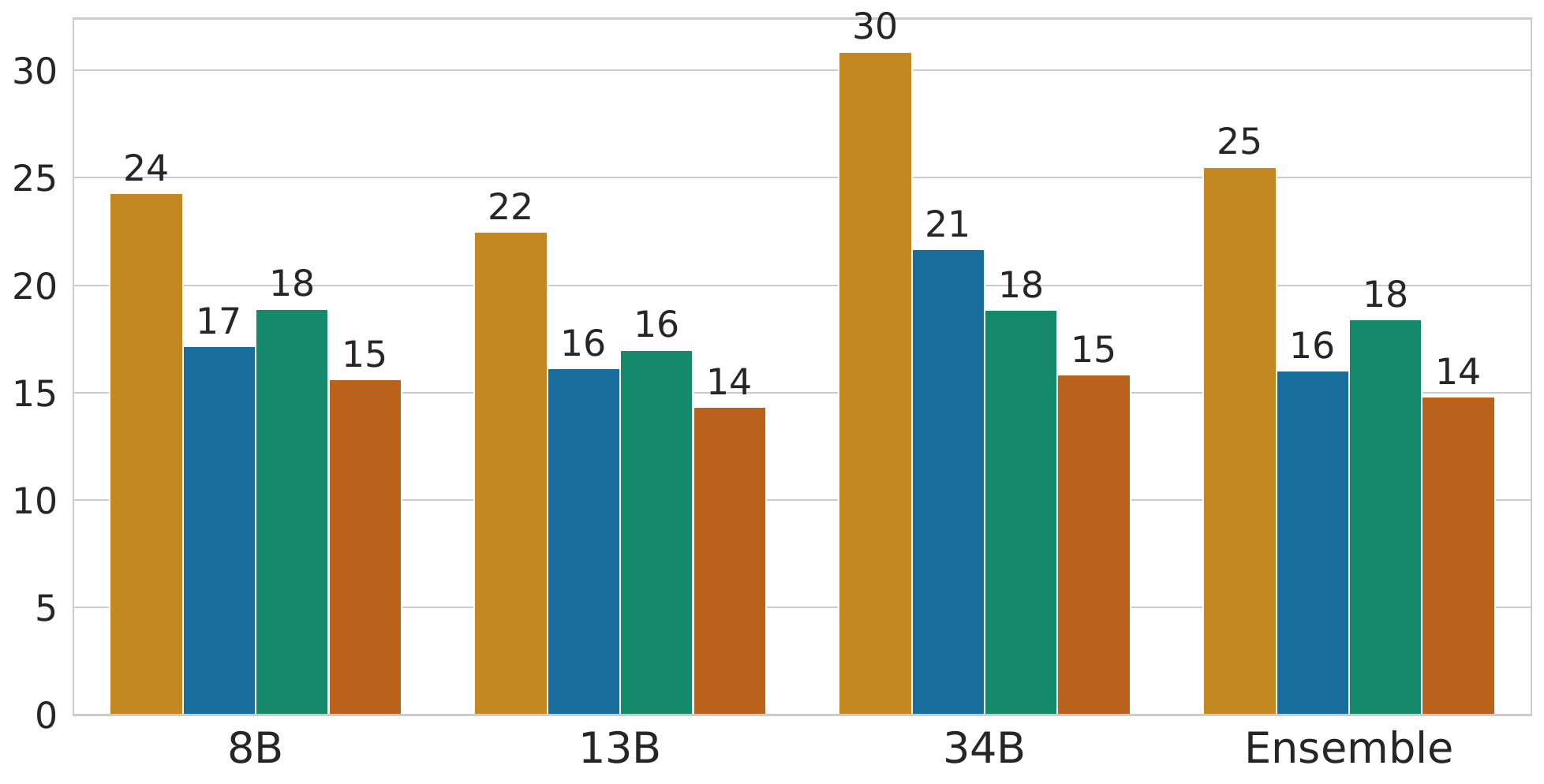}
        \caption{MMMU Pro}
    \end{subfigure}
        \hfill
    \begin{subfigure}[b]{0.49\linewidth}
        \includegraphics[width=\linewidth]{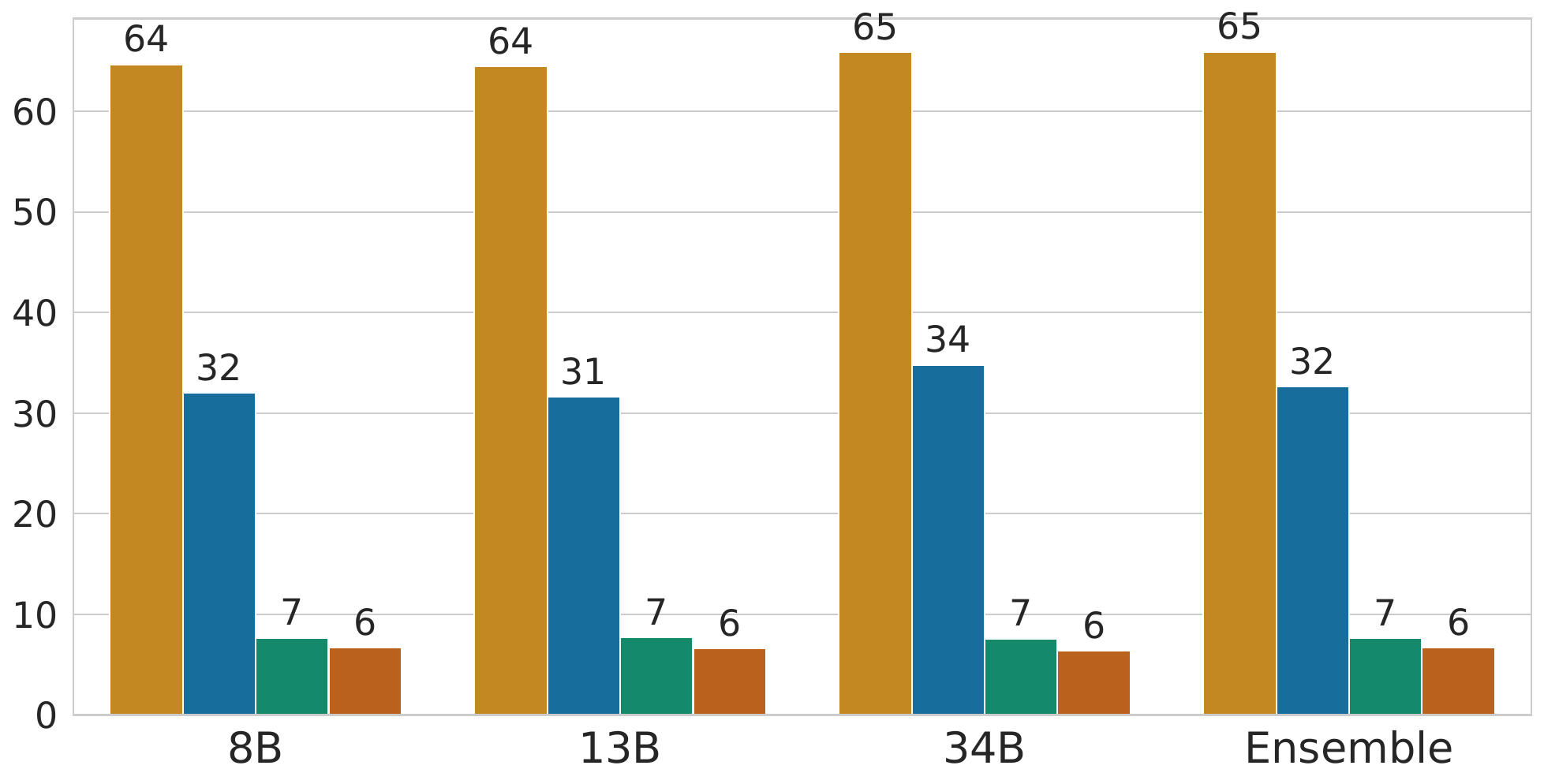}
        \caption{GQA}
    \end{subfigure}
    \hfill
        \hfill
    \begin{subfigure}[b]{0.49\linewidth}
        \includegraphics[width=\linewidth]{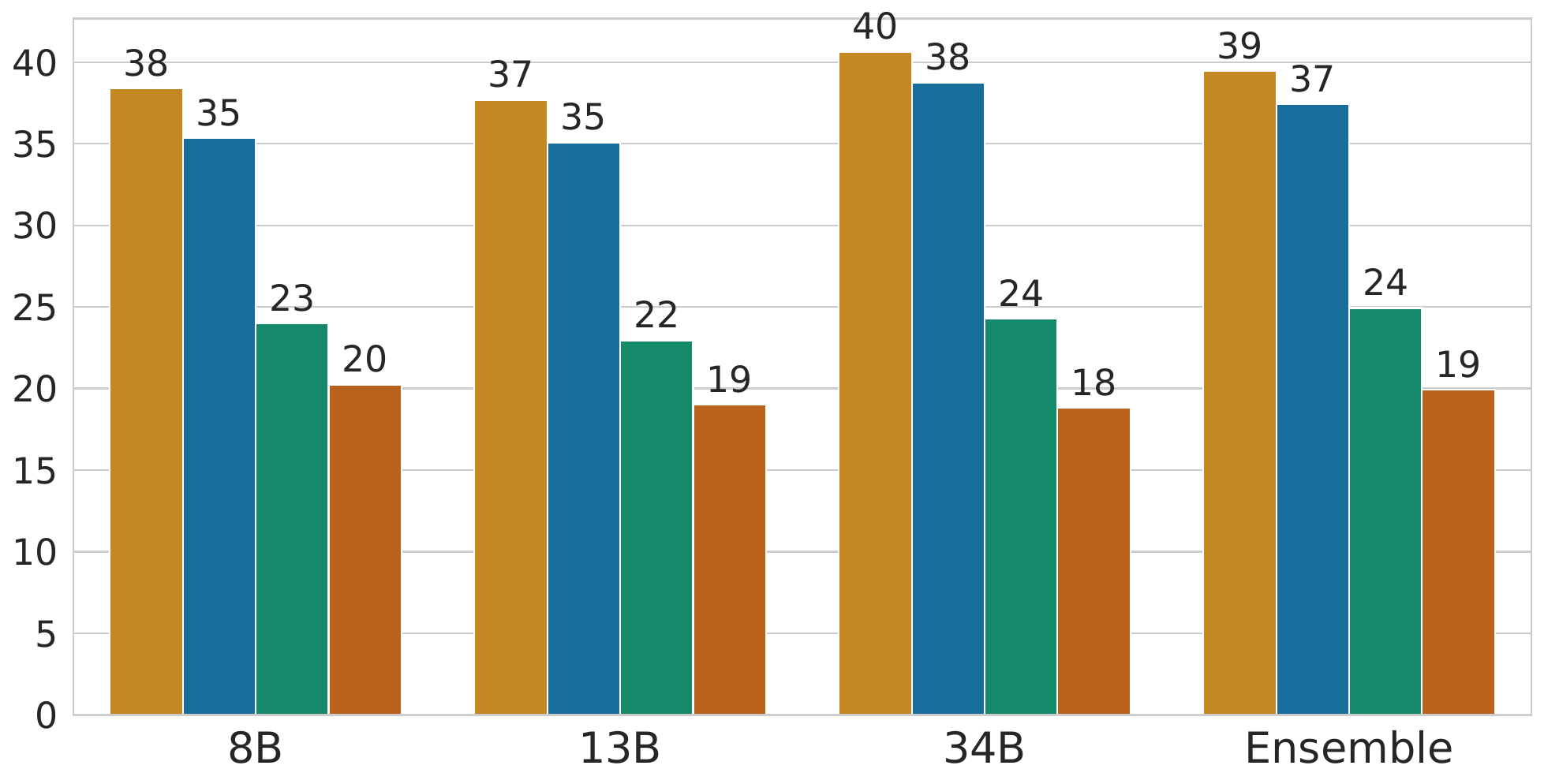}
        \caption{ScienceQA}
    \end{subfigure}
        \hfill
    \begin{subfigure}[b]{0.49\linewidth}
        \includegraphics[width=\linewidth]{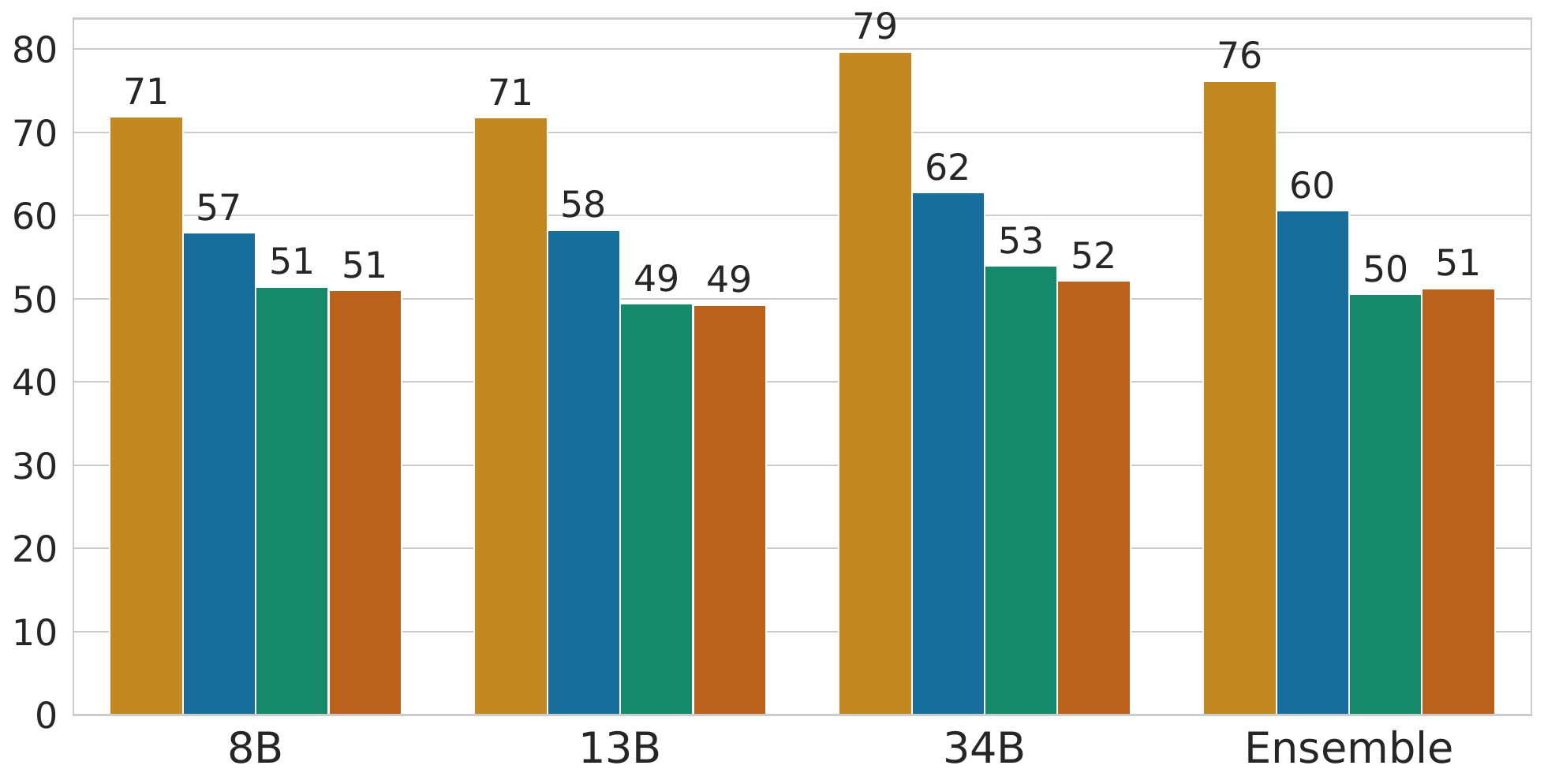}
        \caption{Omni}
    \end{subfigure}
            \hfill
    \begin{subfigure}[b]{0.49\linewidth}
        \includegraphics[width=\linewidth]{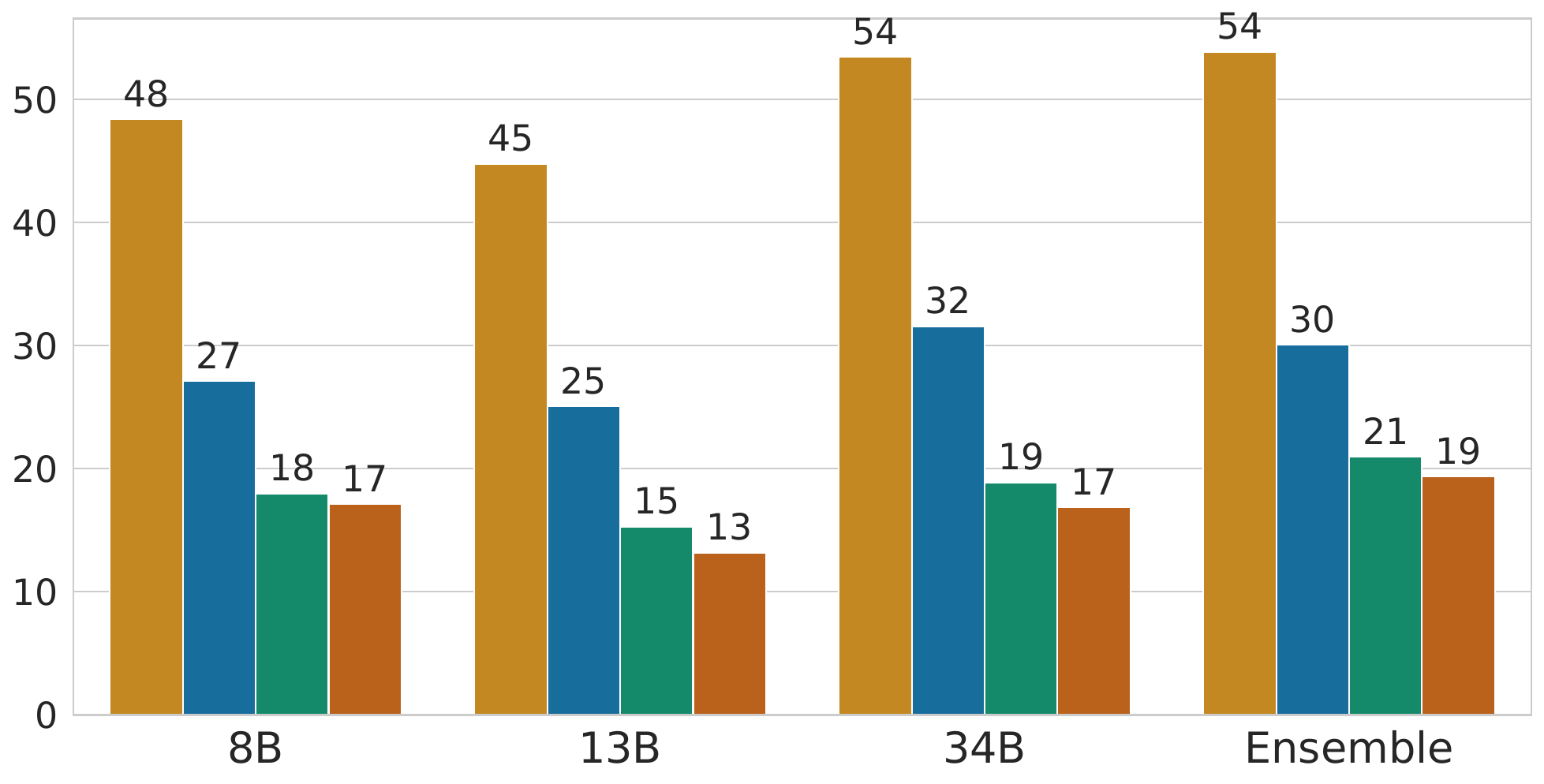}
        \caption{MathVista}
    \end{subfigure}
    \caption{{\bf Effect of model size on additional datasets with text intra-modality dependency.} Performance of various models (8B, 13B, 34B, and a majority-vote ensemble). The bars represent \textcolor{radaryellow}{standard accuracy} and contributions from \textcolor{radarblue}{text}, \textcolor{radargreen}{image}, and \textcolor{radarorange}{random} (bars are in the same order).\label{fig:appendix_text_model_size}}
\end{figure*}

\begin{figure*}[t]
    \centering
    \hfill
    \begin{subfigure}[b]{0.49\linewidth}
        \includegraphics[width=\linewidth]{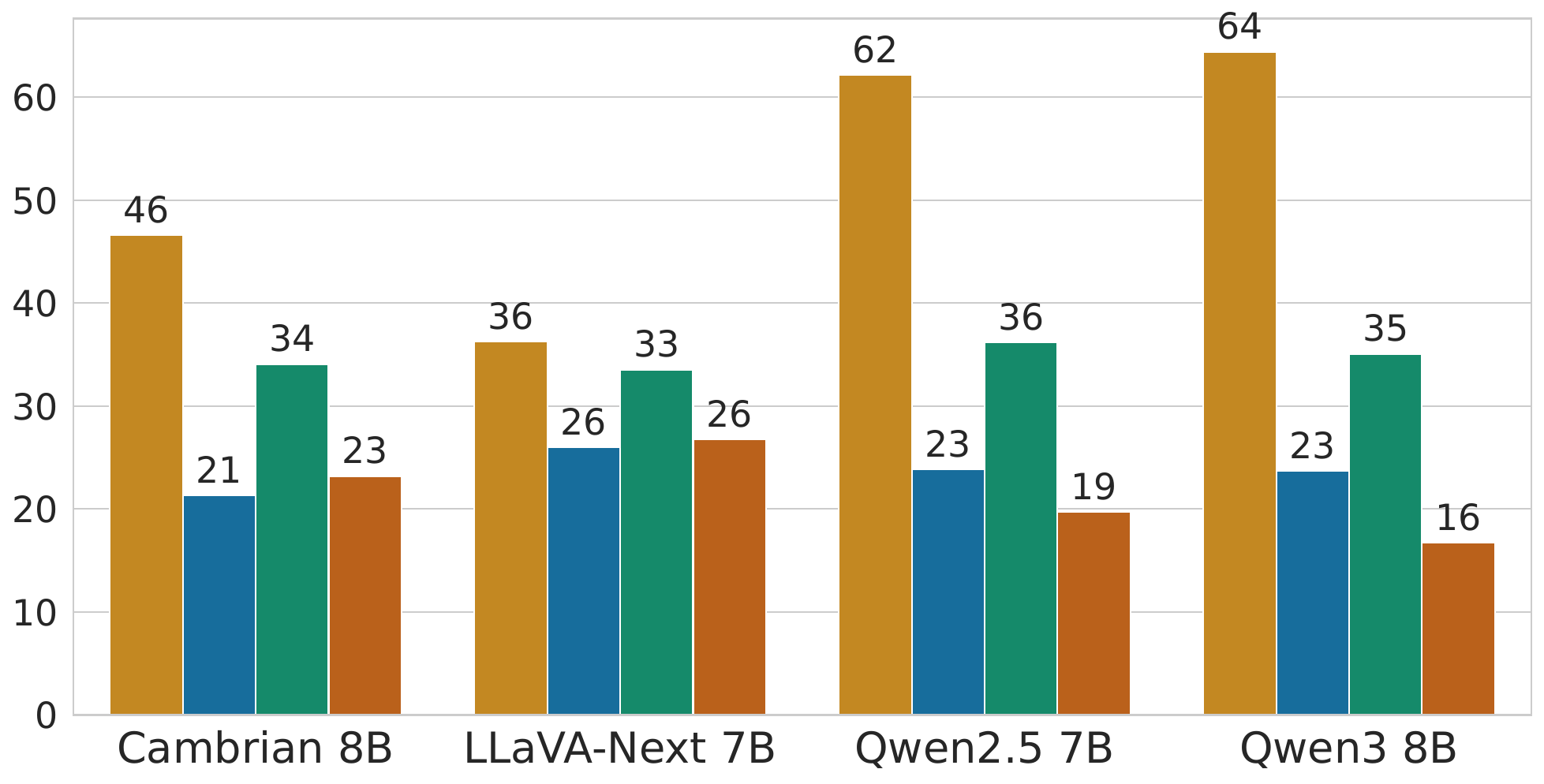}
        \caption{MMStar}
    \end{subfigure}
    \hfill
    \begin{subfigure}[b]{0.49\linewidth}
        \includegraphics[width=\linewidth]{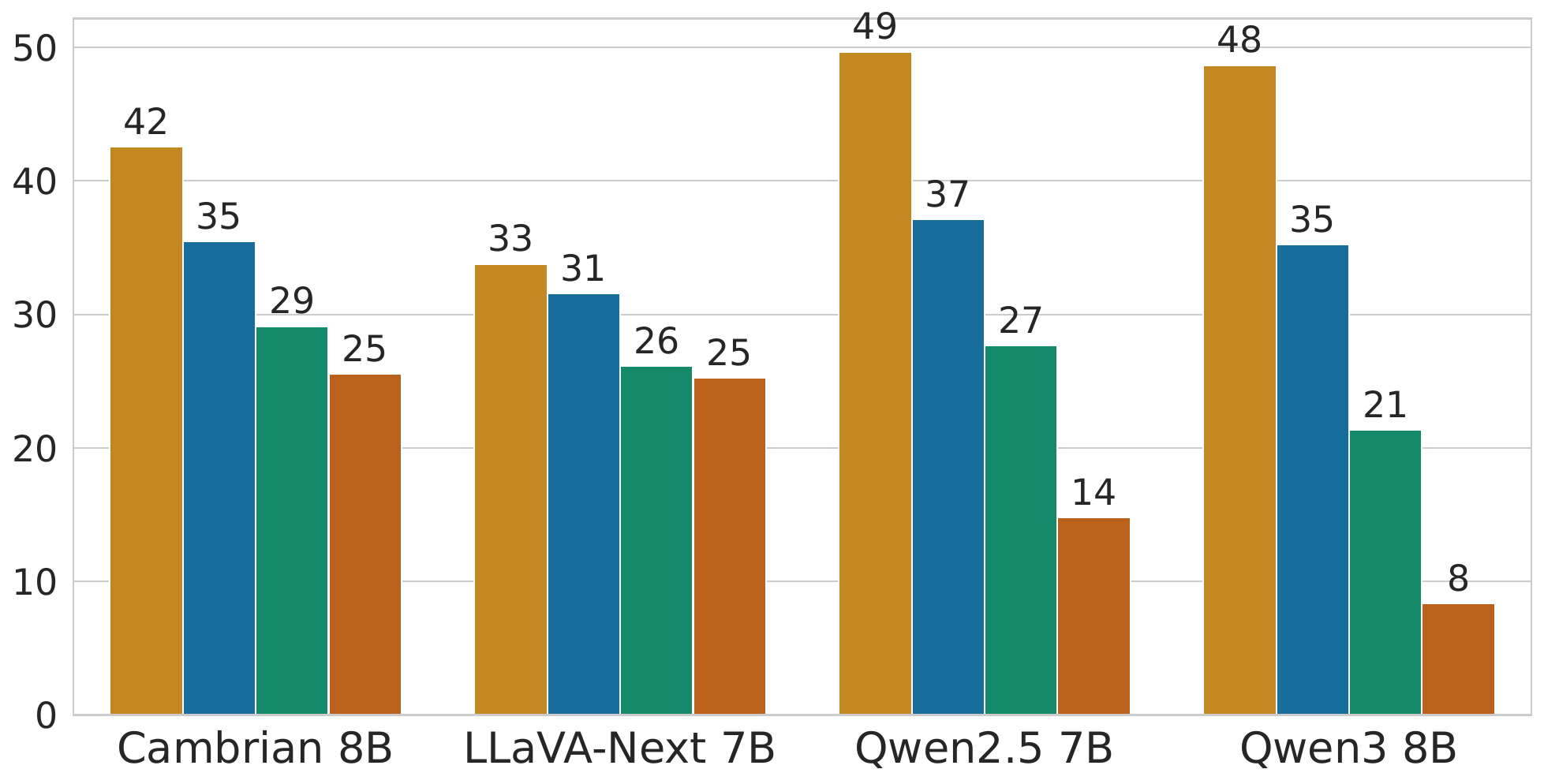}
        \caption{MMMU}
    \end{subfigure}
        \hfill
    \begin{subfigure}[b]{0.49\linewidth}
        \includegraphics[width=\linewidth]{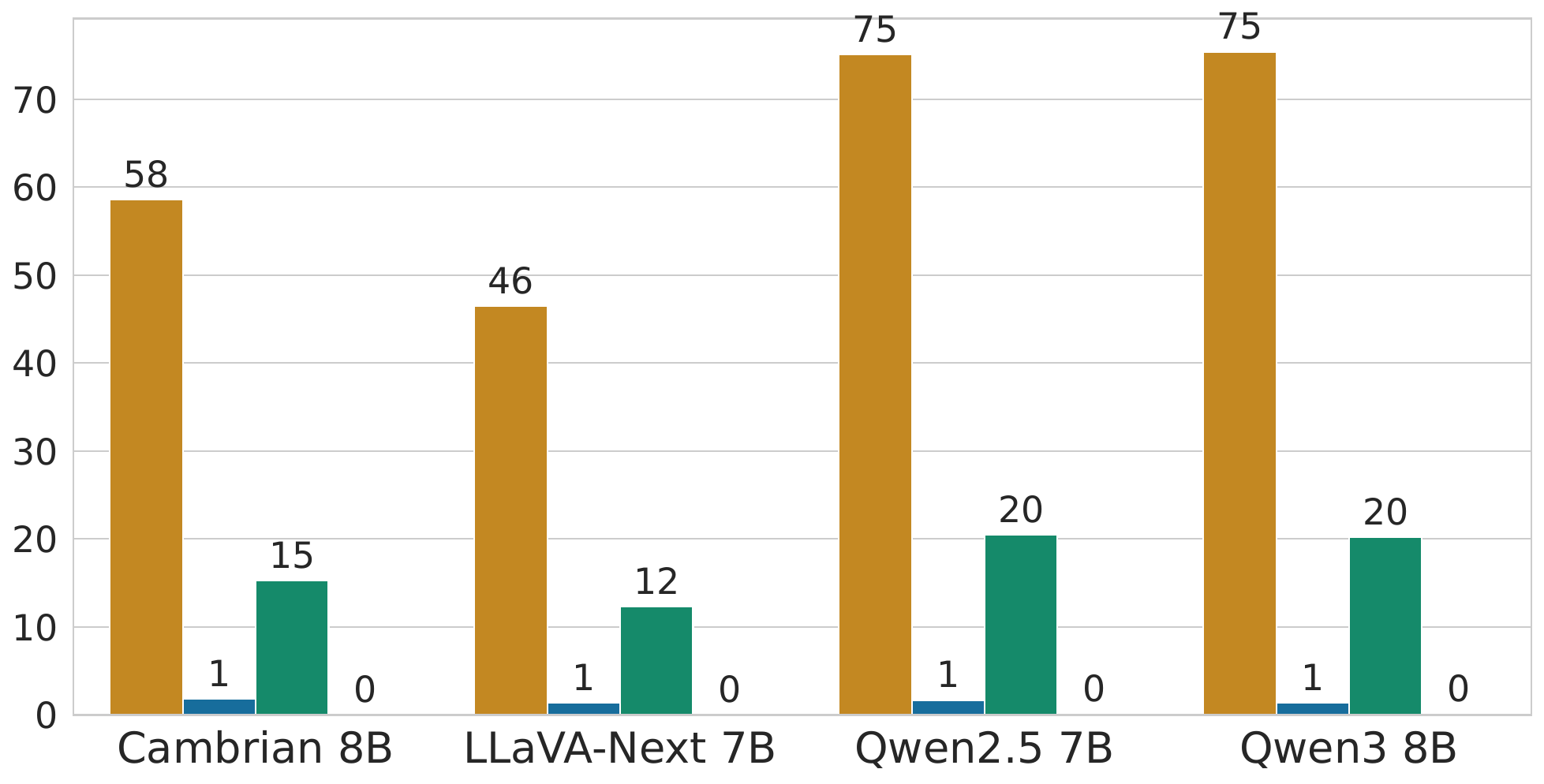}
        \caption{OCRBench}
    \end{subfigure}
    \hfill
    \begin{subfigure}[b]{0.49\linewidth}
        \includegraphics[width=\linewidth]{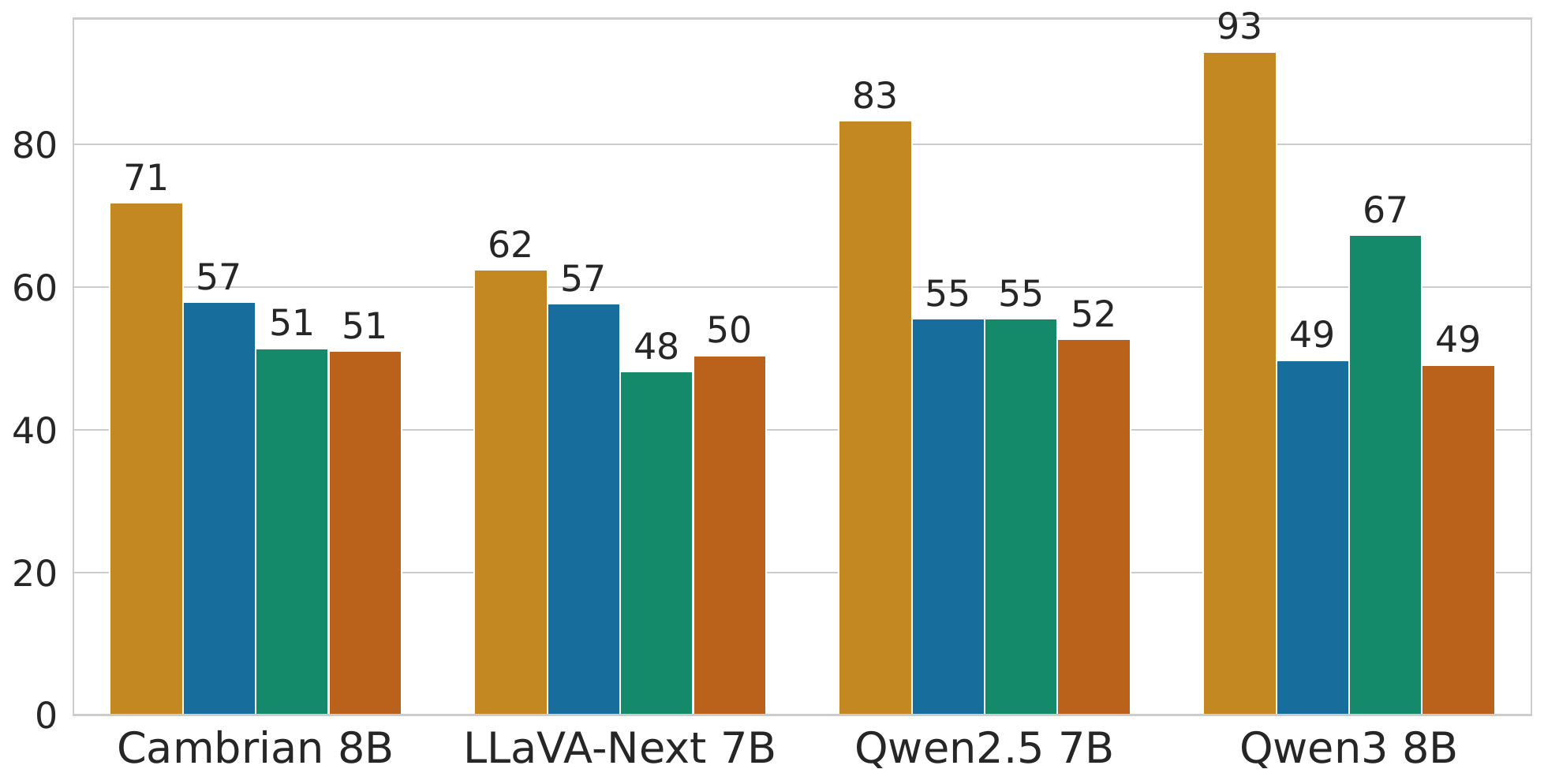}
        \caption{Omni}
    \end{subfigure}
        \hfill
    \begin{subfigure}[b]{0.49\linewidth}
        \includegraphics[width=\linewidth]{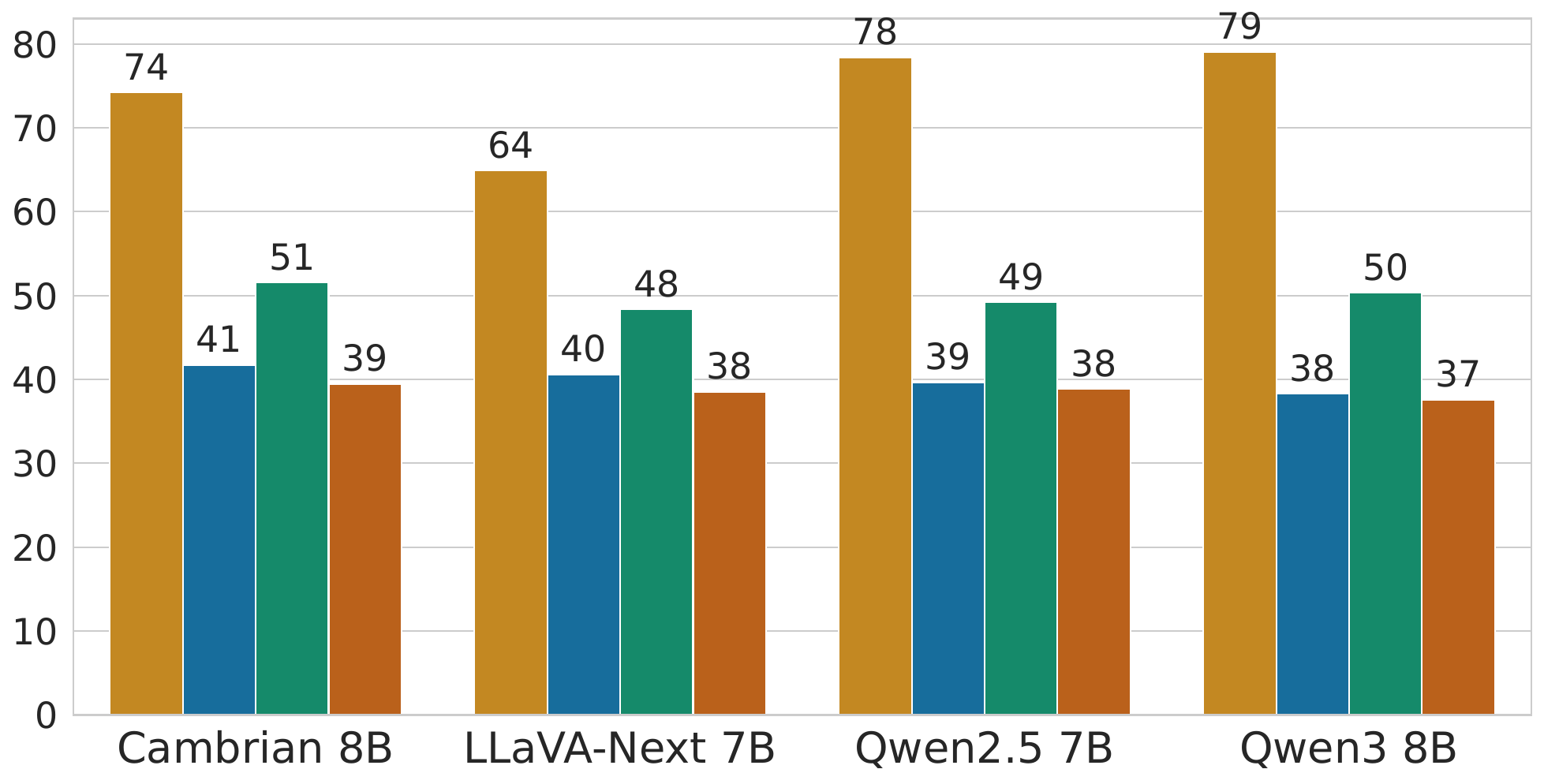}
        \caption{QBench}
    \end{subfigure}
    \hfill
    \begin{subfigure}[b]{0.49\linewidth}
        \includegraphics[width=\linewidth]{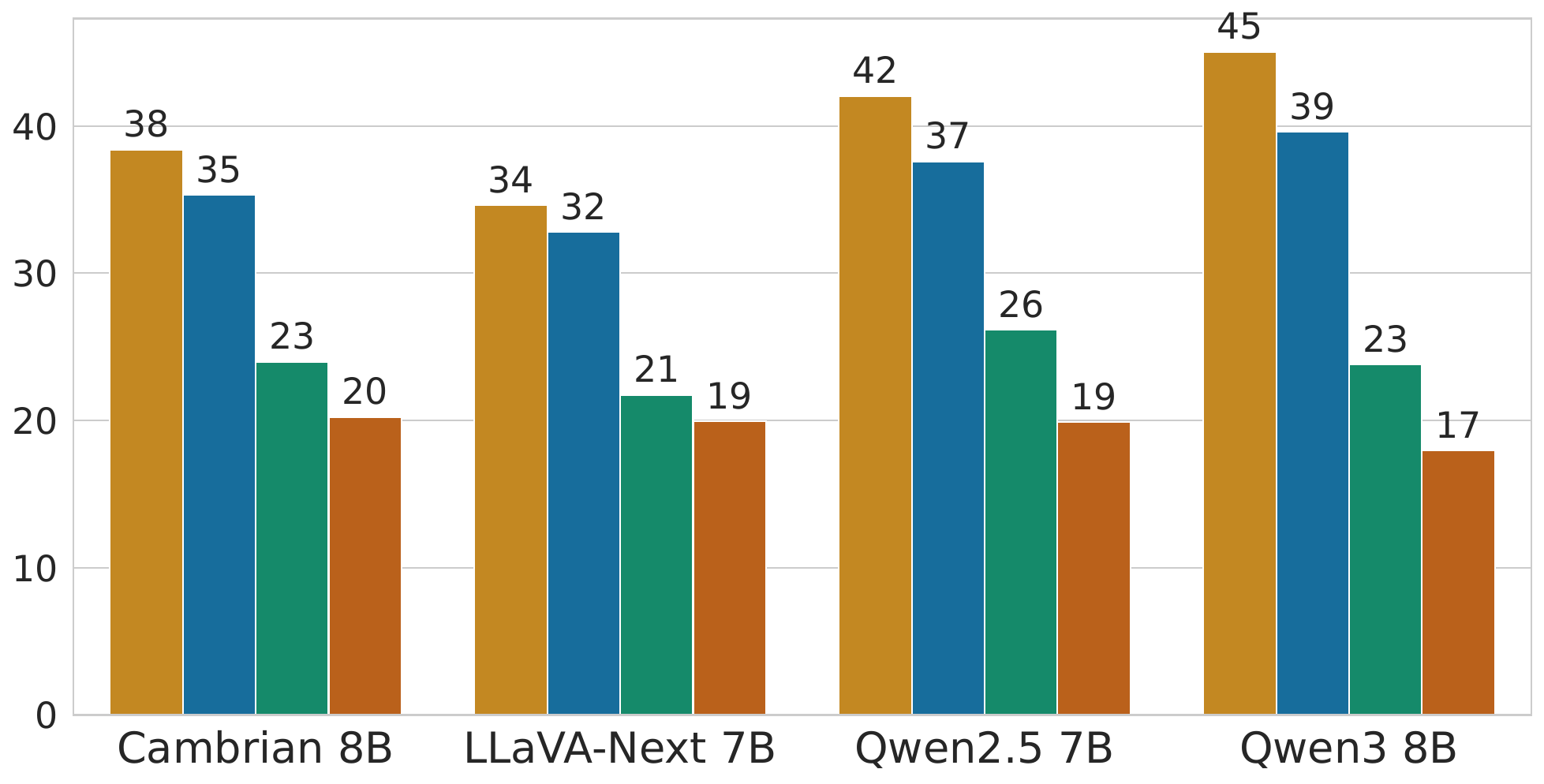}
        \caption{ScienceQA}
    \end{subfigure}
        \hfill
    \begin{subfigure}[b]{0.49\linewidth}
        \includegraphics[width=\linewidth]{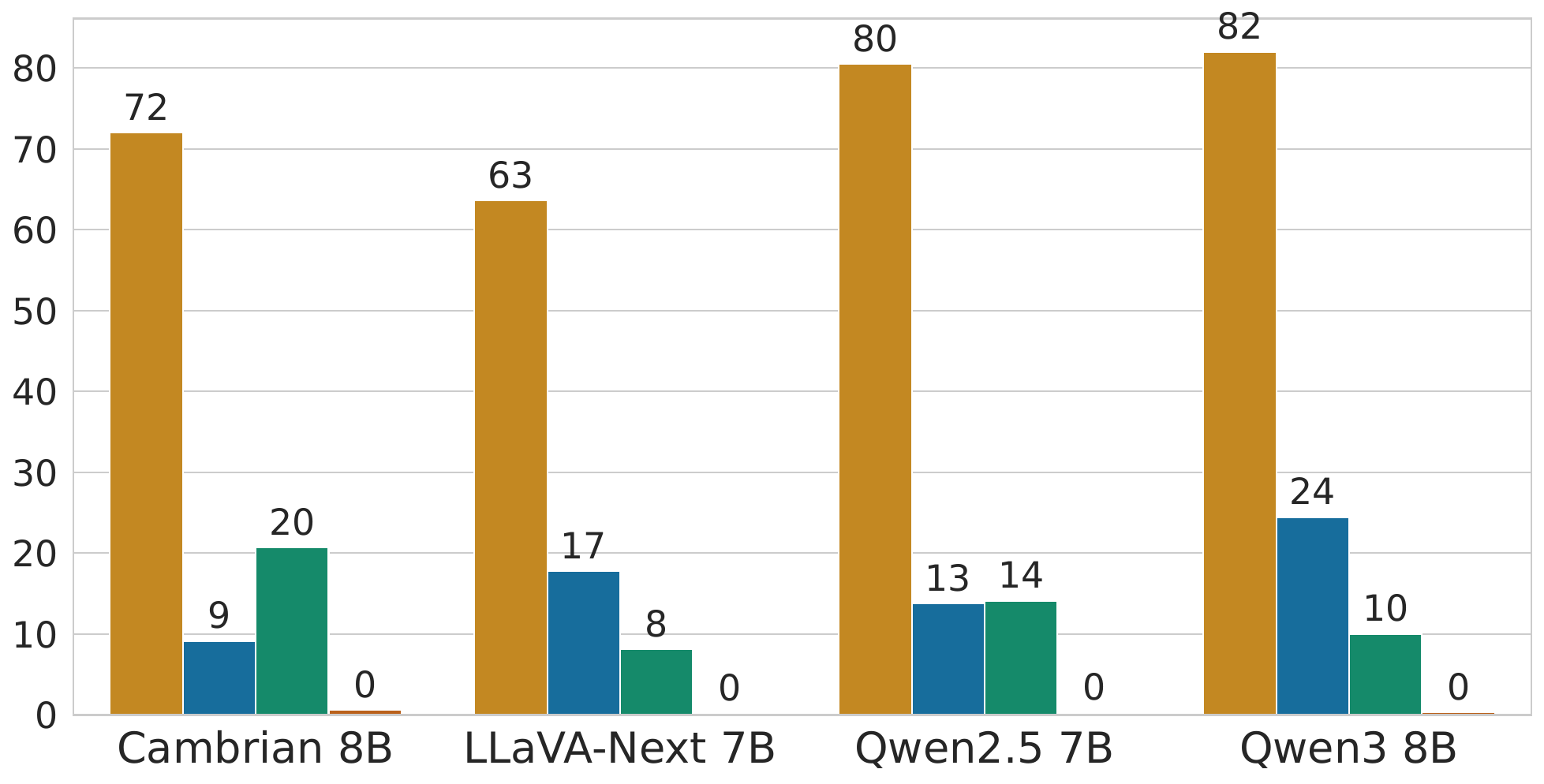}
        \caption{TextVQA}
    \end{subfigure}
    \hfill
    \begin{subfigure}[b]{0.49\linewidth}
        \includegraphics[width=\linewidth]{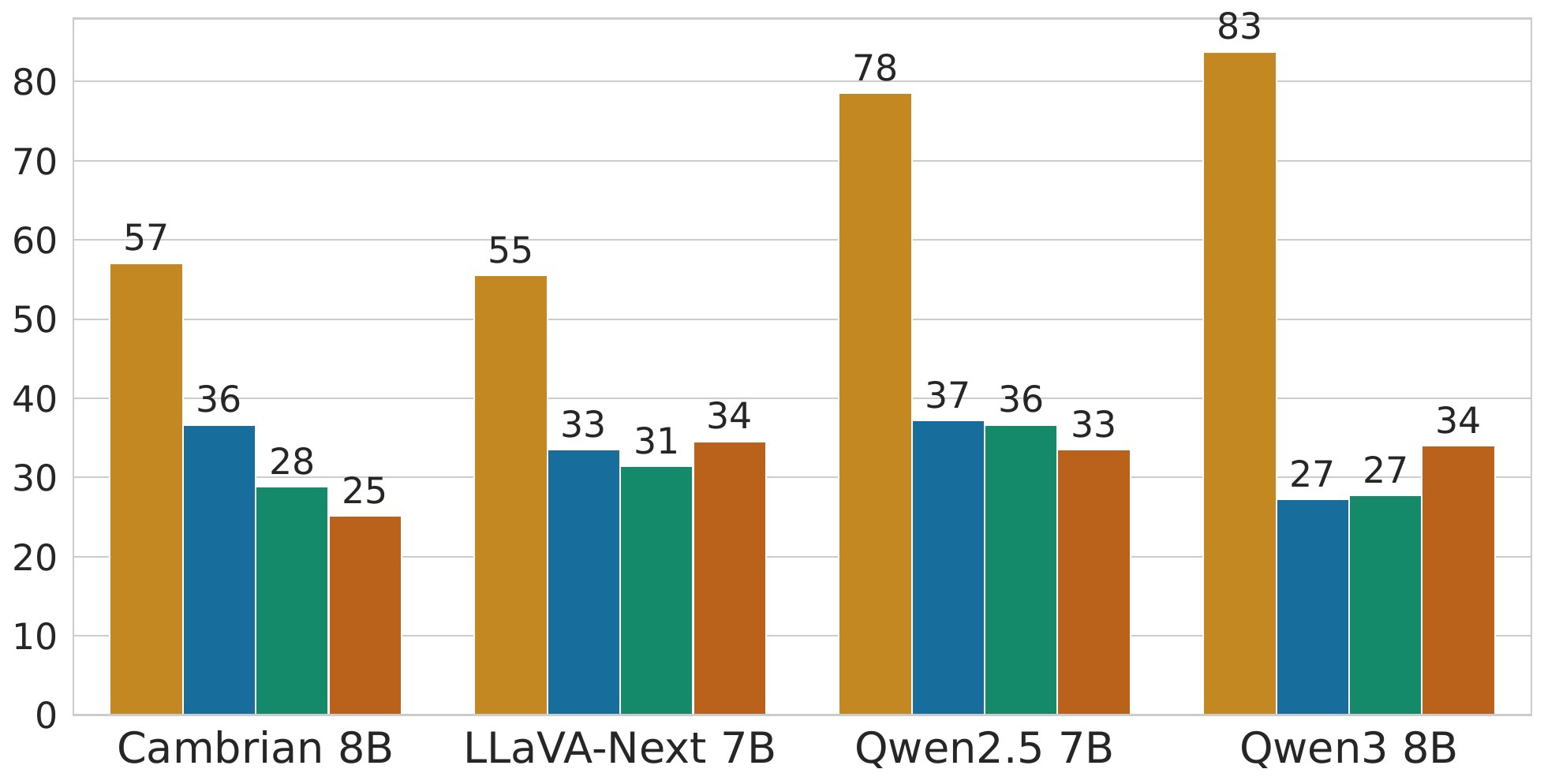}
        \caption{$V^*$ Bench}
    \end{subfigure}
    \caption{{\bf Effect of model types on additional datasets.} {Performance comparison between various models such as LLava-Next (May 2024), Cambrian-1 8b (June 2024), Qwen2.5-VL (April 2025) and Qwen3-VL (October 2025). The bars represent \textcolor{radaryellow}{standard accuracy} and attributed contributions from \textcolor{radarblue}{text}, \textcolor{radargreen}{image}, and \textcolor{radarorange}{random} (bars are in the same order).\label{fig:appendix_model_types}}}
\end{figure*}

The results on different categories for MME~\citep{fu2023mme} and BLINK~\citep{fu2024blink} datasets are shown in \Cref{fig:appendix_category_analysis}. We further show the effect of model size on additional datasets in \Cref{fig:appendix_model_size}, 
\Cref{fig:appendix_image_model_size}, \Cref{fig:appendix_text_model_size} and the  
effect of model types on additional datasets in \Cref{fig:appendix_model_types}.

\end{document}